\documentclass[preprint,12pt,number]{elsarticle}

\usepackage{amssymb}
\usepackage{amsmath}
\usepackage[utf8]{inputenc}

\usepackage{amsfonts}
\usepackage{array}
\usepackage[caption=false,font=normalsize,labelfont=sf,textfont=sf]{subfig}
\usepackage{textcomp}
\usepackage{stfloats}
\usepackage{url}
\usepackage{verbatim}
\usepackage{graphicx}
\usepackage{multirow,multicol}
\usepackage{adjustbox} %
\usepackage{fontawesome5}
\usepackage{utfsym}
\usepackage{xcolor}
\usepackage{arydshln}
\usepackage{fourier}

\usepackage{placeins}
\usepackage{geometry}

\makeatletter
\let\c@author\relax
\def\ps@pprintTitle{%
  \let\@oddhead\@empty
  \let\@evenhead\@empty
  \let\@oddfoot\@empty
  \let\@evenfoot\@oddfoot
}
\makeatother

\usepackage{paralist, tabularx}
\usepackage{changepage} %
\usepackage{moresize}

\usepackage[breaklinks,colorlinks,backref=page]{hyperref} % ,backref=page
\hypersetup{urlcolor=blue, colorlinks=true}

\renewcommand\cite[1]{\citep{#1}}

\usepackage[capitalize]{cleveref}
\crefname{section}{Sec.}{Secs.}
\Crefname{section}{Section}{Sections}
\Crefname{table}{Table}{Tables}
\crefname{table}{Tab.}{Tabs.}

\journal{International Journal of Applied Earth Observation and Geoinformation}

\begin{document}

\begin{frontmatter}

\title{Operational Change Detection for Geographical Information: Overview and Challenges}

\author[label2]{Nicolas Gonthier}
\affiliation[label2]{organization={Univ Gustave Eiffel, ENSG, IGN, LASTIG},%
            country={France}
            }

%% Abstract
\begin{abstract}
The rapid evolution of territories due to climate change and human impact necessitates prompt and effective updates to geospatial databases maintained by National Mapping Agencies. 
That is why change detection is a rapidly evolving and increasingly important field with crucial applications in environmental monitoring, urban expansion, and disaster management. In this review, we present a comprehensive overview of change detection methods and propose future research directions to guide the community toward robust, interpretable, and operational systems, extending beyond the alert stage.
This paper first outlines the fundamental definition of change, emphasizing its multifaceted nature, from temporal to semantic characterization. It categorizes automatic change detection methods into four main families: rule-based, statistical, machine learning, and simulation methods. The strengths, limitations, and applicability of every family are discussed in the context of various input data.
Then, key applications for National Mapping Agencies are identified, particularly the optimization of geospatial database updating, change-based phenomena, and dynamics monitoring.
Finally, the paper highlights the current challenges for leveraging change detection such as the variability of change definition, the missing of relevant large-scale datasets (for both training and evaluation), the diversity of input data, the unstudied no-change detection and the human in the loop integration.

\end{abstract}

\end{frontmatter}

%% main text
%%
%%%%%%%%%%%%%%%%%%%%%%%%%%%%%%%%%%%%%%%%%%%%%%%%%%%%%%%%%%%%%%%%%%%%%%%%%%%%%%%%%%
\section{Introduction}
\label{sec:intro}

Most of the regional and National Mapping Agencies (NMA) have been structured by geospatial databases (also called geodatabases) built upon time (administrative boundaries, geodetic control marks, land use, land cover, topographic data, etc.), for environmentally sensitive planning and monitoring \citep{longley_introduction_2005} and serving public policies \cite{groot_challenges_1999}.
The agencies set up a data collection and recording system based on field information, to keep track of diverse topics: major and minor public works \cite{niroshan_ml_2024} up to abrupt changes such as natural disasters \cite{guo_understanding_2010}  to reflect the changes of the territory on maps and databases \cite{olteanu-raimond_scale_2017}. The information collected can be changes that have already occurred or future ones.
Over time, the number of information sources has increased, with the addition of aerial and satellite monitoring, administrative online publications, and more recently, with the "New space" \cite{aglietti_current_2020}, online social networks \cite{heidemann_online_2012}, and volunteered contributions \cite{sui_crowdsourcing_2012, truong_role_2022}. 
In the research field, automatic change detection methods thrive by leveraging advances in computer science \cite{coppin_digital_2004,stilla_change_2023}, statistics \cite{truong_selective_2020}, machine learning \cite{shi_change_2020}, and remote sensing \cite{mall_change_2022,shafique_deep_2022}.

However, the NMA update pipeline has not undergone fundamental changes for the time being, despite the introduction of automatic Change Detection (CD) methods.  
Traditionally, CD \cite{knudsen_automated_2003} refers to all computational methods used to detect changes in a given area between two dates, but the task may move to the capacitive unit to analyze temporal sequences of data to determine the relevant evolution \cite{irvin_teochat_2025}. 
Change detection, such as land cover mapping \cite{mallet_current_2020}, must address operational constraints and go beyond the experimental setup. This requires methods that are versatile, interactive, scalable to large territories, and effective at high spatial and semantic resolutions. 
Operational services demand reliable information layers that are temporally and spatially consistent, ideally at a national scale, and independent of nomenclature. 
Ultimately, the objective is not only to generate alerts, but also to enable direct database updates, with filtering and human validation tailored to specific use cases, geodatabase specifications, and institutional needs. 
Another objective is to develop feedback loops between GIS updating and remote sensing planning, as you can see in the Overview figure.

\begin{figure*}[ht]
    \centering
    \includegraphics[width=1\linewidth]{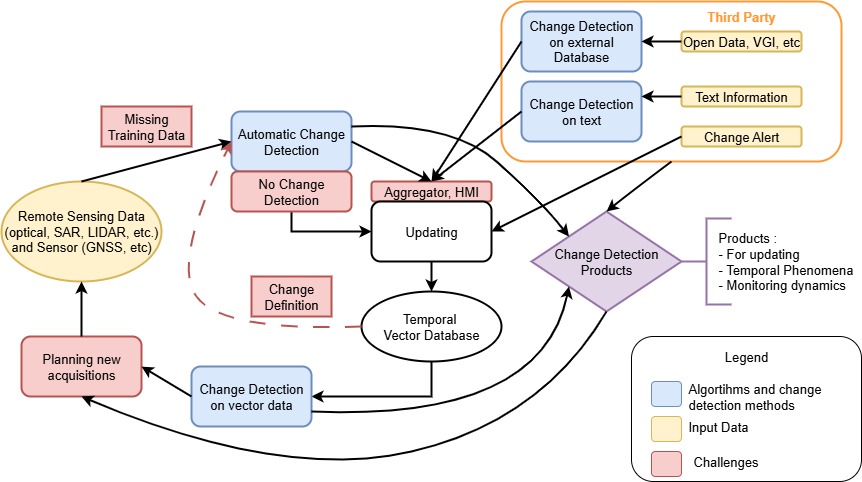}
    \caption{\textbf{Overview of the current challenges in the change detection integration.} Change Detection can be used for database updating, change phenomena or dynamics monitoring but it still faces several challenges especially to close the loop between acquisitions and updating vector geodatabase.}
    \label{fig:graphical_abstract}
\end{figure*}

An overview of the subject seen by NMA can be found in \cref{fig:graphical_abstract}.

In this paper, we discuss the role of change detection applied to large-scale operational geodatabases. Change detection and the main methods are first defined in \cref{sec:Def_DCHAN}. Then, the key applications of change detection are presented in \cref{sec:Key_applications}, followed by a discussion of the current challenges in \cref{sec:challenges_DCHAN}. 

%%%%%%%%%%%%%%%%%%%%%%%%%%%%%%%%%%%%%%%%%%%%%%%%%%%%%%%%%%%%%%%%%%%%%%%%%%%%%%%%%%

\section{Definitions}
\label{sec:Def_DCHAN}

\subsection{Definition of change}
\label{sec:facets}

The definition of a change depends on the intended application. 
This concept is deeply intertwined with the adopted representation of the world, or \emph{ground truth}, as defined by database specifications in a geographic information system (GIS) \cite{pickles_ground_1995}. The interpretation of what constitutes a change depends on the objects of 
interest \cite{kadri-dahmani_updating_2001}. %
Thus, there is no such thing as a generic change and only spatio-temporal events \cite{yu_spatiotemporal_2020}. A change can be characterized by six aspects, developed in \ref{annex:facets}.
The first five facets of change are defined by \cite{zhu_remote_2022} in the context of remote sensing data but can be applied to any input data. The first two aspects involve identifying when and where a significant difference occurs between two observations of the same element. The third aspect of change pertains to the semantization of this change, ranging from land cover \cite{meyer_changes_1994} to physical alterations \cite{falconnier_modelling_2020,cui_spatiotemporal_2019}, while the fourth aspect encompasses the quantification of change (magnitude, speed, etc.) \cite{petit_quantifying_2001,kennedy_trajectorybased_2007}. The "why" aspect addresses the causes of change (climatic variability, natural or anthropogenic disturbances, etc.). 
The sixth and newly introduced aspect is the \textbf{significance} of observed changes is crucial for mapmakers and policymakers \cite{groot_challenges_1999,kadri-dahmani_updating_2001}. 
Not all detected changes require updates to geodatabases or represent a relevant phenomenon, as their significance can vary based on user perspectives \cite{pickles_ground_1995}. 

\begin{figure*}[ht!]
    \centering
    \includegraphics[width=1\linewidth]{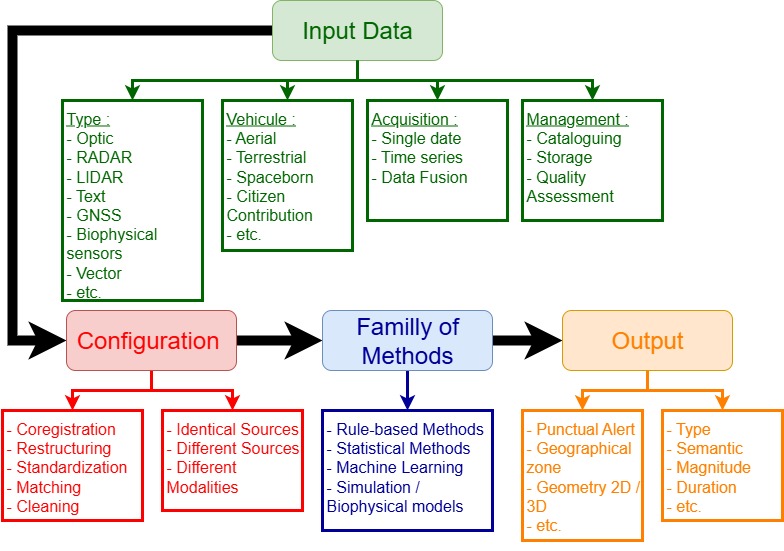}
    \caption{Full Pipeline of data for the change detection task.}
    \label{fig:full_pipeline}
\end{figure*}

In \cref{fig:full_pipeline}, one can see the full overview of the change detection pipeline. The four main elements are the input data (\cref{sec:inputdata}), the configuration\footnote{Configuration refers to selecting input data, often involving two or multiple images and the data comparisons from similar or distinct sources, etc.} (\cref{annex:Review_methods}), the method used (\cref{sec:FourFamilies_Overview}), and the output format (\ref{annex:output_level}). 
Experimental mapping \cite{mallet_current_2020} focuses on developing and assessing new methods, while operational mapping focuses on upscaling processes and delivering reliable products within a predefined time schedule \cite{chen_global_2015}. That's why
all the pre-processing steps (co-registration, normalization, pairing, etc.) are crucial in an operational setup, whereas most of the time they are omitted from the experimental one.

\subsection{A wide variety of data that is difficult to manage}
\label{sec:inputdata}

The number of remote sensing data sources has exploded in terms of quantity, data characteristics (optical \cite{tian_largescale_2022}, multispectral \cite{daudt_urban_2018}, hyperspectral \cite{amieva_deeplearningbased_2023}, RADAR \cite{hornacek_potential_2012}, LIDAR \cite{degelis_benchmarking_2021}, 3D data \cite{qin_3d_2016}), acquisition vectors (airborne, spaceborne, UAV \cite{cracknell_development_2018} or terrestrial), and accessibility.
Some sensors have a very high temporal revisit ability, allowing us to work with time series rather than a single image \cite{guyet_long_2016,toker_dynamicearthnet_2022,dagobert_detection_2022}.

This temporal depth allows NMAs to move closer to near-real-time updates, although it also increases the volume of data to process and archive. 
Second, the integration of crowd-sourced and volunteered data (e.g. GNSS traces, Volunteered geographic information (VGI), social media contributions) has gained momentum, providing complementary perspectives to authoritative data and uncovering features often absent from traditional workflows \cite{ivanovic_filteringbased_2019,olteanu-raimond_scale_2017}.
However, their heterogeneity and uncertain quality still challenge operational adoption. 
Finally, the emergence of textual and multimodal data fusion, fueled by advances in natural language processing and foundation models, represents a disruptive opportunity to capture administrative or planned changes before they manifest in imagery \cite{adrot_using_2022}.

These advances imply a transition from periodic and cyclical updating of geodatabases to continuous and dynamic revision workflows, where multiple heterogeneous streams must be reconciled. Operationally, this demands scalable processing chains, standardized data ingestion frameworks, and robust human-in-the-loop validation to ensure that the richness of new input data translates into consistent and high quality geospatial knowledge.

\subsection{The	four main families of methods of automatic change detection}
\label{sec:FourFamilies_Overview}

The diversity of all the sources mentioned above means that there is no single automatic CD method, but rather a multitude. Even for a given source, various approaches might exist, depending on the needs and stakes involved. 
Different data sources do not provide the same geometric of semantic information \cite{zhu_review_2018}.
Independent of the data source, we can distinguish four main families of CD methods as in \cite{yu_spatiotemporal_2020}\footnote{Initially, the four families have been defined to cluster event extraction methods.}: 
\begin{enumerate}
\setlength\itemsep{-0.5em}
\item Rule-based methods ;
\item Statistical methods ;
\item Machine learning methods ;
\item Simulation / biophysical models.
\end{enumerate}
It should be noted that the boundaries between the different families of methods are porous, and approaches can combine components from different families. More details can be found in \ref{annex:Review_methods}.

\subsubsection{Rule-based methods}

Rule-based methods are based on extracting explicit features and applying predefined thresholds or logical operators to detect changes \cite{jin_comprehensive_2013}. They remain attractive for operational pipelines due to their simplicity, interpretability, and low computational cost. Classical approaches based on pixel differences, ratios, or canonical correlations \cite{howarth_procedures_1981,nielsen_multivariate_1998} are still used for remote sensing data, whereas alerts based on keywords are used for textual data.
Object-based and semantic rules \cite{nebiker_building_2014} now operating at the level of buildings or roads, ease integration with vector geodatabases. Rule-based logic is also being hybridized with uncertainty modeling (e.g., fuzzy thresholds) and applied to deep-learned embeddings \cite{saha_unsupervised_2019}, where simple rules can exploit richer semantic features. However, these methods remain limited by their reliance on handcrafted heuristics, which can reduce transferability across sensors, regions, and applications. Their performance depends strongly on data quality and requires extensive pre-processing \cite{du_radiometric_2002}.% and adaptive thresholds.

\subsubsection{Statistical methods}

Statistical methods remain attractive for operational CD due to computational efficiency and explicit uncertainty handling. Beyond classical summaries and tests (means/SD \cite{knoefel_germanys_2021}, advanced statistics \cite{derksen_geometry_2020}), recent evolutions focus on: (i) robust, distribution-free change-point detection with online/streaming capability (extending sliding-window schemes \cite{truong_selective_2020}) to handle dense time series and irregular sampling; (ii) season–trend decomposition with adaptive thresholds, improving resilience to phenology and acquisition variability over multi-year archives (building on parametric fits \cite{verbesselt_realtime_2012}); (iii) spatiotemporal modeling (neighborhood/graph constraints) to suppress isolated false alarms while preserving sharp boundaries; (iv) multiple-testing control (e.g., FDR) and a-contrario formulations \cite{tailanian_contrario_2023,dagobert_detection_2022} to keep large-area false positives bounded; (v) uncertainty aware fusion (fuzzy logic \cite{yang_automatic_2018,wu_postclassification_2017}, Dempster–Shafer \cite{vallet_homological_2013}) that aggregates heterogeneous sensors; and (vi) statistical testing in learned feature spaces. 
Maximum likelihood estimation \cite{bhattacharya_maximum_1987}, or hidden Markov models are widely applied to time series for failure/outlier detection \cite{truong_selective_2020,amiri-simkooei_offset_2018,gazeaux_detecting_2013,quarello_gnssseg_2022}.
These methods facilitate incremental geodatabase updates and human-in-the-loop triage through ranked, reproducible alerts, and explainable decisions. However, they remain sensitive to distribution shifts (new sensors or new regions) and require careful calibration, while rare or subtle changes still pose challenges for operational scalability.

\subsubsection{Machine Learning}

Supervised deep learning has emerged as the most promising solution for CD, currently setting the state of the art for raster-based approaches \cite{shafique_deep_2022,peng_deep_2025}, especially in the semantic case \cite{daudt_multitask_2019,toker_dynamicearthnet_2022}. 
Recent advances in deep learning have been adapted to the case of supervised change detection, such as Transformers \cite{zhao_adapting_2023,zhu_changevit_2024,chen_remote_2022} and Mamba \cite{chen_changemamba_2024,paranjape_mambabased_2024,zheng_changemask_2022} but without bringing very significant gains \cite{corley_change_2024}.
Models designed to operate directly on vector geodatabases pave the way for operational updating \cite{bernhard_mapformer_2023,chen_objformer_2024}. 
Label-efficient learning could be the next main game changer in CD: self-supervised pretraining and representation learning noticeably reduce the reliance on scarce change labels and improve transfer across sensors/regions \cite{mall_changeaware_2023,akiva_selfsupervised_2022,mendieta_geospatial_2023}, while weakly/semi-supervised schemes leverage coarse or partial supervision at scale \cite{daudt_weakly_2021,wang_stcrnet_2023}.
Together, these developments foster scalable pipelines that pretrain once, fine-tune locally, and generate map-ready edits with uncertainty cues.
However, deep learning models are often overconfident \cite{hechtlinger_cautious_2018}, lacking reliable uncertainty estimation \cite{li_overcoming_2025}, which can restrict their application in detecting extremely rare changes. Moreover, semantic CD still lags behind binary detection despite recent advances \cite{toker_dynamicearthnet_2022,chen_changemamba_2024} even in the experimental setup.

\subsubsection{Simulation and Biophysical model}

Numerical simulation makes it possible to work on spatio-temporal scales that differ from simple field observations, to predict future state,  to understand the causes of change or to test scenarios.
Indeed, these simulations are valuable for providing early warnings and improving disaster mitigation, such as in flash flood \cite{hapuarachchi_review_2011}, earthquake \cite{hoshiba_numerical_2015}, or dust storm warning systems \cite{yu_3d_2017}, but also model social \cite{pan_global_2019} or ecological processes \cite{gomez-dans_location_2022,falconnier_modelling_2020}. 
 These methods are well-suited for predicting land evolution based on physical models, such as soil erosion \cite{guo_modelling_2019}, landslides \cite{vanbeek_regional_2004}.  
Multi-agent simulations \cite{parker_multiagent_2003} or cellular automata can also simulate complex system behaviors, such as land cover evolution or contamination detection \cite{shafiee_agentbased_2013}. 
Note that both simulation and machine learning methods can be used for predicting changes \cite{metzger_urban_2023}.
Numerical simulations are a promising tool for generating synthetic data to evaluate CD methods \cite{champion_2d_2010} or to train machine learning models \cite{song_syntheworld_2023,benidir_change_2025,varghese_viewdelta_2025}.
Although powerful, simulations are computationally intensive with respect to spatial and temporal resolution and must regularly assimilate real-world observations to remain accurate \cite{hoshiba_numerical_2015,dee_bias_2005}.

\subsection{Output levels}

CD methods can have different outputs, depending on the input data, the method, and the main goal. Outputs can be point alerts (precisely geolocated or not), object geometry (2D or 3D) and additional information such as the type of change, its semantics, amplitude or duration.
In the case of two input data at T1 and T2, we can define six output levels, illustrated in \cref{fig:output}, but it can be extended to other types of data. The first level is binary alert when we only get information about the whole area. When the change is located, one obtains a change map. The simplest version is a binary map, but it can contain mono-class  or similarity information; a semantic map indicates the trajectory of change, whereas a panoptic map segments each object individually.
The objective is to maximize integration into the NMA production process by advancing to the most detailed output level.
Moreover, it is important to note that the elements of each source or output can be fuzzy in contrast to what is represented in \cref{fig:output} such as information coming from text. 
When updating geographic databases, the location of the change and its semantics are the two most important elements. They are needed to filter alerts according to their semantics (buildings, hydrography, etc.) to respond to a particular user.
Other needs will be interested in the amplitude or duration of a change, as in the case of monitoring the health of a forest \cite{stevens-rumann_considering_2022}.
In addition, output changes can be forecast based on future projects or trends \cite{voiron-canicio_forecasting_2012} or predictive models \cite{metzger_urban_2023}.

%%%%%%%%%%%%%%%%%%%%%%%%%%%%%%%%%%%%%%%%%%%%%%%%%%%%%%%%%%%%%%%%%%%%%%%%%%%%%%%%%%
\section{Key applications}
\label{sec:Key_applications}

\subsection{Optimizing updating solutions}

Reducing both the time and cost of updating geospatial databases is important for NMA to meet the growing demand for more diverse and timely data. Minimizing the lag between the modification of an object in the field and its integration into the geodatabase requires the development of successful CD tools that deliver significant gains in production efficiency. 
Although data quality is often associated with resolution or/and accuracy, timeliness is becoming increasingly critical (in land mapping \cite{1spatial_little_2025}, agricultural monitoring \cite{vandervelden_governance_2025}, and disaster response \cite{jayawardene_role_2021}). 
Various public policies, such as those related to the Common Agricultural Policy, infrastructure, coastal erosion, and forest monitoring, are moving towards continuous updating, with promising results already emerging from pilot projects \cite{concito_impact_2023}.
In response to this need, strategies for leveraging change detection tools may include transitioning towards a dynamic database that is continuously updated, sometimes with a full catch-up at a given frequency.
Data production will have to shift to be entirely driven by change information, with differential treatment depending on the topics. 
Thus, the CD must be able to reduce overdetection, to detect multiple types of objects, and to exploit multiple remote data sources to improve updating frequency and potentially semantic accuracy. 
It should be noted that this requires effective management of the interdependencies between the different phases of the pipeline and strategic choices at each stage (\cref{fig:full_pipeline}).

\subsection{Temporal Phenomena}

Some change-based phenomena can only be observed through temporal analysis. These phenomena may be related to a deviation from normal behavior (drift or rupture).
For example, deviations from expected patterns can signal declining forest health \cite{kennedy_detecting_2010}, while height time series disturbances can indicate land movements \cite{li_timeseries_2019}. 
Beyond simply updating geospatial databases, change detection is instrumental in mapping thematic issues such as climate change and artificialization of land \cite{foodandfaoagricultureorganization_status_2015,hallegatte_mapping_2016}. The capacity of CD method to detect spatio-temporal phenomena must be developed to address problems that cannot be resolved through field surveys and visual analysis due to prohibitive cost or time constraints.
This involves the development of efficient time series analysis and visualization methods that can handle the current challenges of CD (definition of change, multimodality, and interaction loop).  
Combining remote sensing with social data \cite{bichen_multimodal_2025} is another interesting axis of research to study social temporal phenomenon.

\subsection{Monitoring dynamics}

In the context of climate change, providing reliable indicators on the evolution of specific zones is the key to monitoring environmental and territorial dynamics. These indicators can highlight areas of significant change, guiding new targeted acquisitions and focused investigations for updating geospatial databases and reaching a fully integrated and looped pipeline (see Overview figure). 
This approach not only improves the efficiency of monitoring efforts, but also ensures that interventions are timely and data-driven. This requires being able to distinguish what can be seen by a particular sensor and what cannot.
The LaVerDi \cite{knoefel_germanys_2021} and the European Ground Motion \cite{crameri_european_2023} services can be seen as remote sensing evolutivity scores, but they are semantically limited.
Moreover, equivalent vector-based services are still missing. 
The interaction between detection methods at different spatial and semantic resolution scales is completely absent from the literature.
Another topic that deserves to be addressed is the ethics of monitoring.

%%%%%%%%%%%%%%%%%%%%%%%%%%%%%%%%%%%%%%%%%%%%%%%%%%%%%%%%%%%%%%%%%%%%%%%%%%%%%%%%%%
\section{Current challenges for operational change detection}
\label{sec:challenges_DCHAN}

Change detection has been studied extensively since the 90s \cite{singh_review_1989,mouat_remote_1993,bhattacharya_maximum_1987}, but with few significant operational impacts. 
Indeed, operational services \cite{mallet_current_2020}  based on CD are needed for accuracy and scalability to produce reliable information layers, temporally and spatially homogeneous over a large territory. This require being efficient, scalable, versatile (to handle diverse input data sources and even add new data to the production chain easily) and interactive (to allow easy modification and validation to guarantee the quality standards).   
Modern CD has to respect operational constraints and go beyond the experimental setup. 
The main issues about change detection can be decomposed into: (i) the difficulty of change definition, (ii) the lack of large-scale datasets, (iii) multimodality, (iv) the "no-change detection" concept, which has been neglected so far and finally  (v) the human-in-the loop paradigm.

\subsection{Difficulty of defining relevant change}
\label{sec:difficulty_chan_def}

The first challenge in CD is defining what a change is for the users; thus, it is a specification issue \cite{pickles_ground_1995}. As suggested \cite{carlotto_detecting_1989}, the ability to detect significant changes in the imagery is directly associated with the target application.
A change may be visible in one source but not in another. An event can be a change for a given user but not for another \cite{kadri-dahmani_updating_2001}.
Current methods tend to generate false positives due to misalignment to the geodatabase specifications or to detect errors within the sources. To reduce these false alarms, it is necessary to improve the reliability of sources, but also to work on quality improvement and standardization of geographic databases. 
This involves clear definitions of relevant or meaningful changes \cite{bruzzone_conceptual_2010} and ensuring consistency in specifications throughout the geodatabase. 
Enhancing interoperability between databases can help share change information and transfer methods across production processes and institutions.
Thus, more works are needed on consistency maintenance \cite{kadri-dahmani_consistent_2008}, spatio-temporal ontologies \cite{kauppinen_modeling_2007,bernard_tsn_2017} and semantics but also on temporal database \cite{kulkarni_temporal_2012,rainunandal_spatiotemporal_2013} for better sharing of change information and to embrace the time dimension of the updating of database \cite{peuquet_time_2005}.

\subsection{Missing of  large-scale  datasets}

Establishing an effective evaluation framework to test new methods used in an operational mapping process is crucial but missing for now. 
Such a benchmark, created following the specifications of various thematic layers produced by different NMAs, is needed to enable the evaluation of methods from academia or private companies, at large scale.
The research questions that arise beyond the simple but costly creation of these evaluation data are how to find a good trade-off between a specific object and all the layers of interest to update, how to evaluate the complete production process and not just the automatic model, and how to manage interoperability between evaluation benchmarks.
Moreover, defining relevant metrics for an operational setting is challenging and still an open question despite previous works \cite{toker_dynamicearthnet_2022}. Those benchmarking frameworks may include A/B testing \cite{young_improving_2014} for addressing the utility of methods by human operators.
In the case of remote sensing, there are no high-quality large-scale annotated datasets for training models for multi-label CD at very high spatial resolution and national scale \cite{luangluewut_problem_2024}, although supervised models are by far the most effective \cite{shafique_deep_2022,peng_deep_2025}. 
Most of the datasets are automatically annotated \cite{daudt_weakly_2021}, spanning a small area \cite{tian_hiucd_2020} or mapping only one type of change: most of the time, building \cite{shen_s2looking_2021,ji_fully_2019,chen_spatialtemporal_2020,li_overcoming_2025} or disaster \cite{gupta_creating_2019}. They are mainly based on mid-resolution imagery such as Sentinel-2 \cite{daudt_urban_2018}.
There is also a lack of generalization of machine learning models to new temporal or spatial domains \cite{tian_hiucd_2020,toker_dynamicearthnet_2022,vincent_satellite_2024} and sensor changes \cite{ferraris_detection_2018}. 
The significance of the dataset lies not just in its scale, but in its quality as well \cite{roscher_better_2024}. The only solution should be to create a multiscale pan-European or even global dataset of change detection with a common nomenclature or at least datasets limited to a regional extent but with interoperable nomenclature.

About the foundation models \cite{jakubik_foundation_2023}, CD must not be considered as a downstream task but as a main task due to its temporal particularity. Thus, the multitemporal data must be included in the pretraining of those models with some prior knowledge about change \cite{mall_changeaware_2023}.

\subsection{Complex and Untamed Data for Operational Change Detection}

\subsubsection{Very High Resolutions}
The difficulty of the CD task is increasing exponentially with the spatial and semantic resolutions, but high spatial and semantic resolutions are needed for operational updating process.
Although image-based CD is widely studied, the case of very high resolution images is still very challenging compared to mid-resolution one\footnote{We consider MR : Medium Resolution (300 m > spatial resolution > 30 m). HR : High Resolution (30 m > spatial resolution > 1 m). VHR: Very High Resolution (spatial resolution < 1 m).}. 
The primary difficulty with this data type is the expense of acquisition and labeling. Conditions of acquisition have a higher impact on the resulting images, leading to an important tilted view \cite{pang_detecting_2023}, shadow, misalignment \cite{feng_finegrained_2015}, and high intraclass variability: vegetation during different seasons, etc. 
In addition to spatial resolution, high semantic resolution (higher semantic classes) is also challenging because it leads to a higher interclass similarity but is necessary to gain time on geodatabase updating. 
Occlusion is still an open problem for CD. Some objects can also be undercover, leading to ambiguity (for instance, a river or a path under tree cover). Leveraging oriented images could be a promising path forward. 
Note that for the 3D data in an operational setting, the challenges are more about being able to deal with acquisitions at different resolutions.

\subsubsection{Data Fusion and multimodality}
\label{sec:fusion}

Multimodal or multisensor approaches \cite{pohl_multisensor_1998} can improve land mapping tasks \cite{sumbul_bigearthnetmm_2021,garioud_flair_2023a}, yet research on multimodal CD is scarce.
The multimodal data fusion methods provide a more robust CD thanks to the complementarity between sources, while the comparison methods enable faster and more complete monitoring, thanks to the higher frequency monitoring \cite{zhu_review_2018}, making the latter more important.
In both cases, the methods need to be able to model the complementarity and differences between sources to represent the fact that some objects may be seen in one source but not in the other.
Recent experiments on the multimodal topic mainly address optical and SAR comparison but most on the time on limited spatial extent or binary detection. The seminal work of \cite{kawamura_automatic_1971} describes the potential of this task over a multimodal collection of datasets for CD. 
These methods have also benefited from deep learning \cite{zhan_crossdomain_2024,liu_deep_2018a,wu_commonality_2022}, but performance is still far from what is achieved when comparing the same modalities.
Sensor or source dissimilarities introduce additional issues in the context of operational CD that are not addressed by most classical methods. This topic could synergize with the sensor-agnostic topic \cite{lowman_seefar_2024} for some subsets of remote sensing data, due to significant barriers to use a new sensor in an operational setting.
This topic requires a wide range of skills and access to a wide range of data for the limited-resource NMA. This is why it is necessary to create versatile and adaptable methods to facilitate access at low cost to the diverse data.
The subject of multimodal CD is developed in \ref{sec:diff_sensors}.

\subsubsection{Text data : a major but overlooked source of data}

One of the main types of data used for updating official geodatabases is textual data: it remains the main source of information produced by administrations or companies. Textual data includes administrative documents at local or national levels, institutional and private publications (websites, municipal minutes, local press, etc.), as well as citizen online publications. These data may be in standardized text form such as official documents, but they are mainly free text, leading to a significant access challenge: various websites, paywalls, formats, etc. 
Furthermore, text analysis has recently exploded with increasingly effective methods such as GPT \cite{openai_gpt4_2024} or LLaMA \cite{touvron_llama_2023}, even for official documents \cite{gesnouin_llamandement_2024}; however, its application to CD remains underexplored. 
One can imagine an operational tool that multimodally combines text, image \cite{farah_mcatnat_2024}, and vector geodatabase even if for now the Vision Language Models (VLM)s only use raster input data \cite{zhang_earthgpt_2024,zhan_skyeyegpt_2024,irvin_teochat_2025,elgendy_geollava_2025,xuan_dynamicvl_2025}.
There are two main challenges for the text data. The first is to be able to automatically detect a past or future change in a given document with respect to the current state of the database, along with its specifications. 
The second is to be able to scrape and analyze all the possible documents in an area of interest and to decide if there is a change or not.
Text-based CD is challenging due to language ambiguity, local specificity, the variety of text data (clean pdf, raw text, outputs from OCR and HTR pipelines, HTML pages, etc.), and the need for cross-referencing sources \cite{dong_crossdocument_2025}.
VLMs are an interesting solution to manage language ambiguities and facilitate information access and discoverability.

\subsubsection{Integrating Citizen-Contributed Data}

VGI offers strong potential for enriching and accelerating geodatabase updates, particularly where official acquisition cycles are too slow \cite{canada_depth_2024}, but concerns about positional accuracy, contributor motivation, data sustainability,  legal implications \cite{olteanu-raimond_scale_2017}, lack of standardization and heterogeneous platforms \cite{guilcher_analysis_2022}, continue to limit its systematic adoption by NMA and more implication by the citizen.
GNSS traces collected during daily mobility or sports activities can provide valuable signals to identify missing or updated paths, but they require robust filtering and matching techniques to separate true changes from noise, detours, or context-related anomalies \cite{ivanovic_filteringbased_2019,ivanovic_potential_2020}. Similarly, the reconciliation of crowdsourced or in-situ vector data with authoritative datasets raises significant challenges in terms of schema alignment, uncertainty handling, and scalability \cite{fan_polygonbased_2016,naumann_manymany_2024,xavier_survey_2016}.
These issues underscore the need to develop advanced anomaly detection, data matching, and quality assessment frameworks, as well as solutions for data accessibility and for maintaining contributor engagement. 

\subsection{The unstudied no-change detection}

The dual task of change detection is "no-change" detection. It consists in identifying areas where landscape elements or objects of interest remain stable. It allows us to focus time and resources only on regions that require attention, and thus reduces the cost and duration needed to maintain large-scale geodatabases. However, changes in operational settings correspond to a very limited portion of areas of interest (e.g., 5\% per year for mapping of land cover at the national scale, in France). 
Although underexplored in the literature, no-change detection is typically inferred from change detection methods \cite{olteanu-raimond_use_2020}, and no-change zones are areas where there are no alerts above a given threshold. Current methods and commercial tools  \cite{airbus_direct_2024,maxar_persistent_2024} often lead to overdetection to guarantee greater exhaustiveness, limiting their effectiveness in identifying no-change zones. Enhancing CD methods can indirectly improve no-change detection, but focusing on the question of "no-change" can allow us to view the problem from a new angle and thus truly bring new methods for this specific operational challenge.
It is theoretically possible to detect "no-change" by modeling the distribution of "all" stable situations but no established method currently exists. Statistical approaches, such as the a-contrario \cite{dagobert_detection_2022,robin_acontrario_2010} or out-of-distribution \cite{bellier_detecting_2024} methods offer a potential starting point, but these methods produce too many false positives and do not generalize to other geographical areas for now.

\subsection{Human in the loop}
\label{sec:human_in_loop}

\subsubsection{Active Learning : a promising methodological path}

Active learning \cite{settles_active_2009,demir_updating_2013} is a promising way to a cost-effective production, particularly in resource-intensive tasks such as per-pixel annotations for CD \cite{ruzicka_deep_2020,deschamps_reinforcementbased_2022,wang_deep_2024} by merging both the constitution of the training dataset and the production of the final product.
Major digital companies already use active learning to manage large datasets \cite{kirillov_segment_2023}, but it is understudied in the case of detecting changes to update maps. It is a challenging topic because many active learning strategies rely on metrics such as entropy \cite{ruzicka_deep_2020,deschamps_reinforcementbased_2022,wang_deep_2024} or low confidence prediction, while uncertainty prediction is difficult to detect for change due to the different types of change (see \cref{sec:facets}) and changes between sources. The imbalance and scarcity of change data, domain shift and the large-scale spatio-temporal shift in the data increase the difficulty of the task, indeed, it is not easy to find an efficient sampling function.
Designing an effective human-in-the-loop system for active learning in change detection requires careful consideration.
How to present the samples efficiently to the experts? 
What are the best type of feedback (binary change/no-change, specific change type, boundary refinement, etc.)? Here again, the main goal is to provide active learning method that can be industrialized and used in operational setup.

\subsubsection{Need for better interaction between humans and machines}

Combining the efforts of various individuals working within the same territory is essential for effective CD and GIS updates. 
First, the effort in working on the method that directly compares a new acquisition and the current state of the geodatabase is key \cite{braun_schema_2004,gressin_updating_2014,niroshan_ml_2024}. 
In addition, by providing confidence scores \cite{saha_confidence_2024}, by developing a tool to aggregate various alerts and information sources, the accuracy and timeliness of detecting changes can be improved. Efficient visualization tools are important to allow stakeholders to understand phenomena and update the database with ease \cite{olteanu-raimond_use_2020}.  
Better organization and better methods for sharing information are crucial to maximize the benefits of collaboration between teams \cite{wang_deep_2024}. The complete overview of the change detection pipeline can be seen in \cref{fig:full_pipeline}.
In the long term, VLM \cite{liu_changeagent_2024,tsujimoto_temporal_2024,elgendy_geollava_2025,xuan_dynamicvl_2025} could be a way to tackle the interaction problem if one day one can train such a model on enough data to learn a robust and aligned representation of remote sensing data, geographical information in text and vector format. This kind of model could handle the different definitions of change mentioned earlier (\cref{sec:difficulty_chan_def}) and the large variability of EO.

\subsection{Critical opinions on methods}

In \cref{tab:main_challange_dchan}, we provide critical opinions on the different families of methods defined in \cref{sec:FourFamilies_Overview}. It is a concise overview of the different methods with respect to the input data. 
Some of the families of methods are too simple to handle such data. 
A more complete scientific review is available in the \ref{annex:Review_methods}, with a presentation of the methods according to the different types of input data used. 
The penultimate column provides a ranking of the input data based on the complexity of using them in an operational setting, whereas the last column highlights the main challenge per data input.
Another important point is the need for an efficient and rapid method because operational services must be reliable for product delivery within a predefined schedule \cite{chen_global_2015} with a reasonable engineering team.

\newgeometry{top=1em, bottom=1em,left=3em,right=3em}     % use whatever margins you want 
% Color 
\newcommand{\high}{\textcolor{red}{High~}}
\newcommand{\med}{\textcolor{orange}{Medium~}}
\newcommand{\low}{\textcolor{green}{Low~}}

\begingroup
\renewcommand{\arraystretch}{1.5} %
\begin{table*}
    \centering
    \resizebox{\textwidth}{!}{%
 \begin{scriptsize}
    \begin{tabular}{|p{0.12\textwidth}|p{0.08\textwidth}||p{0.125\textwidth}|p{0.125\textwidth}|p{0.125\textwidth}|p{0.15\textwidth}||p{0.125\textwidth}|p{0.125\textwidth}|} \hline
            \multirow{2}{0.125\textwidth}{Data} &  \multirow{2}{0.08\textwidth}{Resolution} & \multicolumn{4}{|c||}{Methods} & Complexity & Operational \\ 
            \cline{3-6}
            &  & Ruled based & Statistical & Machine Learning & Simulation & & Challenges \\ \hline
            \multirow{3}{0.15\textwidth}{Optical Image} & VHR & \multirow{2}{0.15\textwidth}{\noway~ Too simple} & \multirow{3}{0.15\textwidth}{\faLaptop~ Efficient for anomaly detection}  & \multirow{3}{0.15\textwidth}{\faLock~ Promising for semantic but \\ lack of data} & \multirow{3}{0.15\textwidth}{\faLock~ Way to create synthetic data} & \high & Spatial and Semantic Accuracy - Scalability \\ \cline{2-2} \cline{7-8}
            & HR & & & & & \med &  Spatial and Semantic Accuracy   \\ \cline{2-3} \cline{5-5} \cline{7-8}
            & MR & \faLockOpen~  Efficient for simple land cover & &  \faLockOpen~  Efficient for simple land cover  & &  \low & Heterogeneoous Comparison \\ \cline{2-6} \cline{7-8}
            & Street View  & {\noway~ Too simple} & {\noway~ Too simple} & \faGlasses ~  Under-studied &  \faGlasses ~  Under-studied & \high & Images in extreme conditions \\ \hline
            \multirow{4}{0.15\textwidth}{SAR} & VHR & \multirow{4}{0.15\textwidth}{\noway~ Too simple} & \faLaptop~ Promishing for small displacements &\multirow{4}{0.15\textwidth}{\faGlasses ~  Under-studied } & \multirow{4}{0.15\textwidth}{\faGlasses ~  Under-studied } & \high & Lack of efficient ML models - Acquisition Cost  \\ \cline{2-2} \cline{4-4} \cline{7-8}
            & HR & &  \faLockOpen~ Efficient for simple land cover and provide good updateness & & & \med &  Reducing Noise to Signal Ratio - Comparing images with different angles  \\ \hline
           Optical Time Series & MR &  \faLockOpen~  Efficient for simple land cover & \faLaptop~ Efficient for anomaly Detection  & \faLock~  Promising for reducing false positive &  \faLock~ Way to create synthetic data & \low & No change Detection \\ \hline
           SAR Time Series & MR &  \faLockOpen~  Efficient when instrumented scene & \faLaptop~ Effective for generic change & \faGlasses ~  Under-studied  & \faLock ~ Promishing for application related to soil characterics and displacement  & \med & Interpretability of the data   \\ \hline
           GNSS Time Series & & \noway ~  Too simple & \faLock ~  Promising & \faWeightHanging ~  Methods too costly &  {\faGlasses ~  Under-studied } & \med & Reducing Noise to Signal Ratio   \\ \hline
           Point Cloud & VHR & \faLockOpen~  Efficient for geometric change & \faLaptop~  Useful for uncertainty information & \faLock~ Promising for semantic change &  {\faGlasses ~  Under-studied} & \med & Scalability and mixing with semantic clues \\ \hline
          DSM & & \faLaptop~  Efficient for geometric change & \faLaptop ~  Useful for uncertainty information  & \faLock~  Promising for semantic change & \faLock ~  Promising for complex scene & \low & Semantisation \\ \hline
         Text & & \faLockOpen ~  Efficient for simple tasks  and provide good updateness & \noway~ No relevant & \faLock ~   Promising for document analysis & \faGlasses~ Under-studied to create synthetic data & \med & Scalability and Interaction Human Machine \\ \hline
           Vector Data & & \faLaptop~ Efficient for geometric change - Strong scalability & - &  \faWeightHanging~ Methods too costly & - & \med & Versatility and transferability on diverse thematic layer - Integration of data with various quality \\  \hline
          Heterogeneous Sensor Comparison  & & {\noway~ Too simple} & {\noway~ Too simple} & \faLock ~ Still hard & \faLock~ Promishing for sensor simulation  & \high & Versatility  \\ \hline
           Data Fusion &  & {\noway~ Too simple} & {\noway~ Too simple} & \faLock~ Promishing to have agnostic pipeline & \faGlasses~  Under-studied for physically based models & \med & Simplicity and integration  \\
    \hline
    \end{tabular}
    \end{scriptsize}
    \vspace{10mm}
  }
\caption{Main challenge in operational Change Detection depending on the type of data and family of methods. \faLock~ indicates a key but complex topic that is not solved yet, \faLockOpen~ indicates an almost solved problem in experimental setting, \faLaptop~ is about the need of transferring experimental solutions at the operational level, \faWeightHanging~ a costly kind of method (data hungry or computationally), \noway~ refers to a dead end method for this kind of data  and  \faGlasses~ indicates the need for more research investigations. The penultimate column provides information about the complexity of the input data. The last column mentions the main challenge of his input data to be used in an operational setting.}
    \label{tab:main_challange_dchan}
\end{table*}
\endgroup

\restoregeometry

%%%%%%%%%%%%%%%%%%%%%%%%%%%%%%%%%%%%%%%%%%%%%%%%%%%%%%%%%%%%%%%%%%%%%%%%%%%%%%%%%%
\section{Conclusion}

This review begins by establishing a core definition of change, emphasizing its multifaceted nature, categorizing automatic change detection techniques into four main families and providing for each family's strengths, weaknesses, and relevant contexts for various input data.
Key applications of CD for NMA are the optimization of geospatial database updating, the monitoring of change-based phenomena, and tracking dynamics. 
The challenges in leveraging CD for operational uses are numerous.
These include the inherent variability in defining "change" for different users, the scarcity of large-scale training or evaluation datasets, the diversity of input data, the under-explored area of "no-change" detection, the crucial aspect of human-in-the-loop integration, and the strict operational constraints that demand high reliability and scalability.
This discussion hopes to foster ongoing innovation in change detection techniques to effectively address the evolving requirements of GIS for NMA.

\FloatBarrier
\small{
\begingroup
\setlength{\bibsep}{0pt plus 0.3ex}
\bibliography{JAG_Review}

\begin{thebibliography}{246}
\providecommand{\natexlab}[1]{#1}
\providecommand{\url}[1]{\texttt{#1}}
\expandafter\ifx\csname urlstyle\endcsname\relax
  \providecommand{\doi}[1]{doi: #1}\else
  \providecommand{\doi}{doi: \begingroup \urlstyle{rm}\Url}\fi

\bibitem[{1Spatial}(2025)]{1spatial_little_2025}
{1Spatial}.
\newblock The {{Little Book}} of {{National Mapping}}, 2025.

\bibitem[Abbaspour et~al.(2021)Abbaspour, Chehreghan, and Chamani]{abbaspour_multiscale_2021}
R.~A. Abbaspour, A.~Chehreghan, and M.~Chamani.
\newblock Multi-scale polygons matching using a new geographic context descriptor.
\newblock \emph{Appl Geomat}, 13\penalty0 (4):\penalty0 885--899, Dec. 2021.
\newblock ISSN 1866-928X.
\newblock \doi{10.1007/s12518-021-00396-x}.

\bibitem[Abdelhady et~al.(2022)Abdelhady, Troy, Habib, and Manish]{abdelhady_simple_2022}
H.~U. Abdelhady, C.~D. Troy, A.~Habib, and R.~Manish.
\newblock A {{Simple}}, {{Fully Automated Shoreline Detection Algorithm}} for {{High-Resolution Multi-Spectral Imagery}}.
\newblock \emph{Remote Sensing}, 14\penalty0 (3):\penalty0 557, Jan. 2022.
\newblock ISSN 2072-4292.
\newblock \doi{10.3390/rs14030557}.

\bibitem[Adrot et~al.(2022)Adrot, Auclair, Coche, Fertier, Gracianne, and Montarnal]{adrot_using_2022}
A.~Adrot, S.~Auclair, J.~Coche, A.~Fertier, C.~Gracianne, and A.~Montarnal.
\newblock Using {{Social Media Data}} in {{Emergency Management}}: {{A Proposal}} for a {{Socio-Technical Framework}} and a {{Systematic Literature Review}}.
\newblock In \emph{{{ISCRAM}}}, 2022.

\bibitem[Aglietti(2020)]{aglietti_current_2020}
G.~S. Aglietti.
\newblock Current {{Challenges}} and {{Opportunities}} for {{Space Technologies}}.
\newblock \emph{Front. Space Technol.}, 1, June 2020.
\newblock ISSN 2673-5075.
\newblock \doi{10.3389/frspt.2020.00001}.

\bibitem[{Airbus}(2024)]{airbus_direct_2024}
{Airbus}.
\newblock Direct infrastructure {{Analytics}} access from across the industry.
\newblock https://intelligence.airbus.com/imagery/analytics/infrastructure-change-detection/, 2024.

\bibitem[Akiva et~al.(2022)Akiva, Purri, and Leotta]{akiva_selfsupervised_2022}
P.~Akiva, M.~Purri, and M.~Leotta.
\newblock Self-{{Supervised Material}} and {{Texture Representation Learning}} for {{Remote Sensing Tasks}}.
\newblock In \emph{{{CVPR}}}. arXiv, 2022.
\newblock \doi{10.48550/arXiv.2112.01715}.

\bibitem[Amieva et~al.(2023)Amieva, Austoni, Brovelli, Ansalone, Naylor, Serva, and Saux]{amieva_deeplearningbased_2023}
J.~F. Amieva, A.~Austoni, M.~A. Brovelli, L.~Ansalone, P.~Naylor, F.~Serva, and B.~L. Saux.
\newblock Deep-{{Learning-based Change Detection}} with {{Spaceborne Hyperspectral PRISMA}} data, Oct. 2023.

\bibitem[{Amiri-Simkooei} et~al.(2018){Amiri-Simkooei}, {Hosseini-Asl}, Asgari, and {Zangeneh-Nejad}]{amiri-simkooei_offset_2018}
A.~R. {Amiri-Simkooei}, M.~{Hosseini-Asl}, J.~Asgari, and F.~{Zangeneh-Nejad}.
\newblock Offset detection in {{GPS}} position time series using multivariate analysis.
\newblock \emph{GPS Solut}, 23\penalty0 (1):\penalty0 13, Nov. 2018.
\newblock ISSN 1521-1886.
\newblock \doi{10.1007/s10291-018-0805-z}.

\bibitem[Bagley(1922)]{bagley_concerning_1922}
J.~W. Bagley.
\newblock Concerning {{Aerial Photographic Mapping}}: {{A Review}}.
\newblock \emph{Geographical Review}, 12\penalty0 (4):\penalty0 628--635, 1922.
\newblock ISSN 0016-7428.
\newblock \doi{10.2307/208595}.

\bibitem[Baldwin(1947)]{baldwin_use_1947}
M.~Baldwin.
\newblock The {{Use Of Aerial Photographs In Soil Mapping}}.
\newblock \emph{Symposium of information relative to vises of aerial photographs by geologists: Washington, DC, Photogramm. Eng}, 13\penalty0 (4):\penalty0 532--536, 1947.

\bibitem[Balsak and San(2023)]{balsak_evaluation_2023}
A.~Balsak and B.~T. San.
\newblock Evaluation of the effect of spatial and temporal resolutions for digital change detection: Case of forest fire.
\newblock \emph{Nat Hazards}, 119\penalty0 (3):\penalty0 1799--1818, Dec. 2023.
\newblock ISSN 1573-0840.
\newblock \doi{10.1007/s11069-023-06199-0}.

\bibitem[Bandara and Patel(2022)]{bandara_transformerbased_2022}
W.~G.~C. Bandara and V.~M. Patel.
\newblock A {{Transformer-Based Siamese Network}} for {{Change Detection}}, Sept. 2022.

\bibitem[Barthelet et~al.(2011)Barthelet, Mercier, and Denise]{barthelet_building_2011}
E.~Barthelet, G.~Mercier, and L.~Denise.
\newblock Building change detection in a couple of optical and {{SAR}} high resolution images.
\newblock In \emph{2011 {{IEEE International Geoscience}} and {{Remote Sensing Symposium}}}, pages 2393--2396, July 2011.
\newblock \doi{10.1109/IGARSS.2011.6049692}.

\bibitem[Bellier and Audebert(2024)]{bellier_detecting_2024}
G.~L. Bellier and N.~Audebert.
\newblock Detecting {{Out-Of-Distribution Earth Observation Images}} with {{Diffusion Models}}.
\newblock In \emph{{{EARTHVISION}} 2024 {{IEEE}}/{{CVF CVPR Workshop}}. {{Large Scale Computer Vision}} for {{Remote Sensing Imagery}}}, 2024.

\bibitem[Benidir et~al.(2025)Benidir, Gonthier, and Mallet]{benidir_change_2025}
Y.~Benidir, N.~Gonthier, and C.~Mallet.
\newblock The {{Change You Want To Detect}}: {{Semantic Change Detection In Earth Observation With Hybrid Data Generation}}.
\newblock In \emph{Proceedings of the {{IEEE}}/{{CVF Conference}} on {{Computer Vision}} and {{Pattern Recognition}} ({{CVPR}})}, Nashville, TN, USA, 2025.

\bibitem[Bernard et~al.(2017)Bernard, {Villanova-Oliver}, Gensel, and Dao]{bernard_tsn_2017}
C.~Bernard, M.~{Villanova-Oliver}, J.~Gensel, and H.~Dao.
\newblock {{TSN}} et {{TSN-change}} : Ontologies pour repr{\'e}senter l'{\'e}volution des d{\'e}coupages territoriaux statistiques.
\newblock In \emph{{{SAGEO}} 2017 - {{Spatial Analysis}} and {{GEOmatics}}}, Rouen, France, Nov. 2017. GdR Magis CNRS.

\bibitem[Bernhard et~al.(2023)Bernhard, Strau{\ss}, and Schubert]{bernhard_mapformer_2023}
M.~Bernhard, N.~Strau{\ss}, and M.~Schubert.
\newblock {{MapFormer}}: {{Boosting Change Detection}} by {{Using Pre-change Information}}.
\newblock In \emph{Proceedings of the {{IEEE}}/{{CVF International Conference}} on {{Computer Vision}} ({{ICCV}})}, pages 16837--16846, Sept. 2023.

\bibitem[Bertin et~al.(2017)Bertin, Collilieux, Lebarbier, and Meza]{bertin_semiparametric_2017}
K.~Bertin, X.~Collilieux, E.~Lebarbier, and C.~Meza.
\newblock Semi-parametric segmentation of multiple series using a {{DP-Lasso}} strategy.
\newblock \emph{Journal of Statistical Computation and Simulation}, 87\penalty0 (6):\penalty0 1255--1268, Apr. 2017.
\newblock ISSN 0094-9655.
\newblock \doi{10.1080/00949655.2016.1260726}.

\bibitem[Bhattacharya(1987)]{bhattacharya_maximum_1987}
P.~K. Bhattacharya.
\newblock Maximum likelihood estimation of a change-point in the distribution of independent random variables: {{General}} multiparameter case.
\newblock \emph{Journal of Multivariate Analysis}, 23\penalty0 (2):\penalty0 183--208, Dec. 1987.
\newblock ISSN 0047-259X.
\newblock \doi{10.1016/0047-259X(87)90152-7}.

\bibitem[Bichen et~al.(2025)Bichen, Chen, Chen, Xiang, Hu, Hu, and Tu]{bichen_multimodal_2025}
F.~Bichen, X.~Chen, H.~Chen, Z.~Xiang, B.~Hu, Z.~Hu, and W.~Tu.
\newblock Multimodal {{Fusion Framework}} for {{Urban}} functional zone change detection using {{Remote Sensing}} and {{Social Sensing Data}}.
\newblock \emph{The International Archives of the Photogrammetry, Remote Sensing and Spatial Information Sciences}, XLVIII-G-2025:\penalty0 453--458, July 2025.
\newblock \doi{10.5194/isprs-archives-XLVIII-G-2025-453-2025}.

\bibitem[Bowman et~al.(2011)Bowman, Balch, Artaxo, Bond, Cochrane, D'Antonio, DeFries, Johnston, Keeley, Krawchuk, Kull, Mack, Moritz, Pyne, Roos, Scott, Sodhi, and Swetnam]{bowman_human_2011}
D.~M. J.~S. Bowman, J.~Balch, P.~Artaxo, W.~J. Bond, M.~A. Cochrane, C.~M. D'Antonio, R.~DeFries, F.~H. Johnston, J.~E. Keeley, M.~A. Krawchuk, C.~A. Kull, M.~Mack, M.~A. Moritz, S.~Pyne, C.~I. Roos, A.~C. Scott, N.~S. Sodhi, and T.~W. Swetnam.
\newblock The human dimension of fire regimes on {{Earth}}.
\newblock \emph{Journal of Biogeography}, 38\penalty0 (12):\penalty0 2223--2236, 2011.
\newblock ISSN 1365-2699.
\newblock \doi{10.1111/j.1365-2699.2011.02595.x}.

\bibitem[Braun(2004)]{braun_schema_2004}
A.~Braun.
\newblock From the {{Schema Matching}} to the {{Integration}} of {{Updating Information Into User Geographic Databases}}.
\newblock In \emph{Proc. 12th {{Int}}. {{Conf}}. on {{Geoinformatics-Geospatial Information Research}}: {{Bridging}} the {{Pacific}} and {{Atlantic}}}, University of G{\"a}vle, Sweden, 2004.

\bibitem[Bruzzone and Bovolo(2010)]{bruzzone_conceptual_2010}
L.~Bruzzone and F.~Bovolo.
\newblock A conceptual framework for change detection in very high resolution remote sensing images.
\newblock In \emph{2010 {{IEEE International Geoscience}} and {{Remote Sensing Symposium}}}, pages 2555--2558, July 2010.
\newblock \doi{10.1109/IGARSS.2010.5653930}.

\bibitem[Bruzzone and Prieto(2000)]{bruzzone_automatic_2000}
L.~Bruzzone and D.~Prieto.
\newblock Automatic analysis of the difference image for unsupervised change detection.
\newblock \emph{IEEE Trans. Geosci. Remote Sensing}, 38\penalty0 (3):\penalty0 1171--1182, May 2000.
\newblock ISSN 01962892.
\newblock \doi{10.1109/36.843009}.

\bibitem[{Canada} and {Portugal}(2024)]{canada_depth_2024}
{Canada} and {Portugal}.
\newblock In-depth review of timeliness, frequency and granularity of official statistics.
\newblock In \emph{72nd Plenary Session of the {{Conference}} of {{European Statisticians}}}, UNECE, 2024.

\bibitem[Carlotto(1989)]{carlotto_detecting_1989}
M.~J. Carlotto.
\newblock Detecting {{Man-Made Changes In Imagery}}.
\newblock In \emph{Intelligent {{Robots}} and {{Computer Vision VII}}}, volume 1002, pages 38--45. SPIE, Mar. 1989.
\newblock \doi{10.1117/12.960257}.

\bibitem[Champion(2011)]{champion_detection_2011}
N.~Champion.
\newblock \emph{D{\'e}tection de Changement {{2D}} {\`a} Partir d'imagerie Satellitaire : {{Application}} {\`a} La Mise {\`a} Jour Des Bases de Donn{\'e}es G{\'e}ographiques}.
\newblock These de doctorat, Paris 5, Jan. 2011.

\bibitem[Champion et~al.(2010)Champion, Boldo, {Pierrot-Deseilligny}, and Stamon]{champion_2d_2010}
N.~Champion, D.~Boldo, M.~{Pierrot-Deseilligny}, and G.~Stamon.
\newblock {{2D}} building change detection from high resolution satellite imagery: {{A}} two-step hierarchical method based on {{3D}} invariant primitives.
\newblock \emph{Pattern Recognition Letters}, 31\penalty0 (10):\penalty0 1138--1147, July 2010.
\newblock ISSN 0167-8655.
\newblock \doi{10.1016/j.patrec.2009.10.012}.

\bibitem[Chehata et~al.(2014)Chehata, Orny, Boukir, Guyon, and Wigneron]{chehata_objectbased_2014}
N.~Chehata, C.~Orny, S.~Boukir, D.~Guyon, and J.~Wigneron.
\newblock Object-based change detection in wind storm-damaged forest using high-resolution multispectral images.
\newblock \emph{International Journal of Remote Sensing}, 35\penalty0 (13):\penalty0 4758--4777, July 2014.
\newblock ISSN 0143-1161.
\newblock \doi{10.1080/01431161.2014.930199}.

\bibitem[Chen and Shi(2020)]{chen_spatialtemporal_2020}
H.~Chen and Z.~Shi.
\newblock A {{Spatial-Temporal Attention-Based Method}} and a {{New Dataset}} for {{Remote Sensing Image Change Detection}}.
\newblock \emph{Remote Sensing}, 12\penalty0 (10):\penalty0 1662, May 2020.
\newblock ISSN 2072-4292.
\newblock \doi{10.3390/rs12101662}.

\bibitem[Chen et~al.(2022)Chen, Qi, and Shi]{chen_remote_2022}
H.~Chen, Z.~Qi, and Z.~Shi.
\newblock Remote {{Sensing Image Change Detection}} with {{Transformers}}.
\newblock \emph{IEEE Trans. Geosci. Remote Sensing}, 60:\penalty0 1--14, 2022.
\newblock ISSN 0196-2892, 1558-0644.
\newblock \doi{10.1109/TGRS.2021.3095166}.

\bibitem[Chen et~al.(2024{\natexlab{a}})Chen, Lan, Song, {Broni-Bediako}, Xia, and Yokoya]{chen_objformer_2024}
H.~Chen, C.~Lan, J.~Song, C.~{Broni-Bediako}, J.~Xia, and N.~Yokoya.
\newblock {{ObjFormer}}: {{Learning Land-Cover Changes From Paired OSM Data}} and {{Optical High-Resolution Imagery}} via {{Object-Guided Transformer}}.
\newblock \emph{IEEE Transactions on Geoscience and Remote Sensing}, pages 1--1, 2024{\natexlab{a}}.
\newblock ISSN 1558-0644.
\newblock \doi{10.1109/TGRS.2024.3410389}.

\bibitem[Chen et~al.(2024{\natexlab{b}})Chen, Song, Han, Xia, and Yokoya]{chen_changemamba_2024}
H.~Chen, J.~Song, C.~Han, J.~Xia, and N.~Yokoya.
\newblock {{ChangeMamba}}: {{Remote Sensing Change Detection With Spatiotemporal State Space Model}}.
\newblock \emph{IEEE Transactions on Geoscience and Remote Sensing}, 62:\penalty0 1--20, 2024{\natexlab{b}}.
\newblock ISSN 1558-0644.
\newblock \doi{10.1109/TGRS.2024.3417253}.

\bibitem[Chen et~al.(2024{\natexlab{c}})Chen, Song, and Yokoya]{chen_change_2024}
H.~Chen, J.~Song, and N.~Yokoya.
\newblock Change {{Detection Between Optical Remote Sensing Imagery}} and {{Map Data}} via {{Segment Anything Model}} ({{SAM}}), Jan. 2024{\natexlab{c}}.

\bibitem[Chen et~al.(2015)Chen, Chen, Liao, Cao, Chen, Chen, He, Han, Peng, Lu, Zhang, Tong, and Mills]{chen_global_2015}
J.~Chen, J.~Chen, A.~Liao, X.~Cao, L.~Chen, X.~Chen, C.~He, G.~Han, S.~Peng, M.~Lu, W.~Zhang, X.~Tong, and J.~Mills.
\newblock Global land cover mapping at 30 ~ m resolution: {{A POK-based}} operational approach.
\newblock \emph{ISPRS Journal of Photogrammetry and Remote Sensing}, 103:\penalty0 7--27, May 2015.
\newblock ISSN 0924-2716.
\newblock \doi{10.1016/j.isprsjprs.2014.09.002}.

\bibitem[Chen et~al.(2021)Chen, Yang, and Stiefelhagen]{chen_drtanet_2021}
S.~Chen, K.~Yang, and R.~Stiefelhagen.
\newblock {{DR-TANet}}: {{Dynamic Receptive Temporal Attention Network}} for {{Street Scene Change Detection}}.
\newblock In \emph{{{IEEE Intelligent Vehicles Symposium}} 2021 ({{IV2021}}).}, May 2021.
\newblock \doi{10.48550/arXiv.2103.00879}.

\bibitem[Colin~Koeniguer and Nicolas(2020)]{colinkoeniguer_change_2020}
E.~Colin~Koeniguer and J.-M. Nicolas.
\newblock Change {{Detection Based}} on the {{Coefficient}} of {{Variation}} in {{SAR Time-Series}} of {{Urban Areas}}.
\newblock \emph{Remote Sensing}, 12\penalty0 (13):\penalty0 2089, Jan. 2020.
\newblock ISSN 2072-4292.
\newblock \doi{10.3390/rs12132089}.

\bibitem[Coppin et~al.(2004)Coppin, Jonckheere, Nackaerts, Muys, and Lambin]{coppin_digital_2004}
P.~Coppin, I.~Jonckheere, K.~Nackaerts, B.~Muys, and E.~Lambin.
\newblock Digital change detection methods in ecosystem monitoring: A review.
\newblock \emph{International Journal of Remote Sensing}, 25\penalty0 (9):\penalty0 1565--1596, May 2004.
\newblock ISSN 0143-1161.
\newblock \doi{10.1080/0143116031000101675}.

\bibitem[Corley et~al.(2024)Corley, Robinson, and Ortiz]{corley_change_2024}
I.~Corley, C.~Robinson, and A.~Ortiz.
\newblock A {{Change Detection Reality Check}}.
\newblock In \emph{{{ICLR Machine Learning}} for {{Remote Sensing}} ({{ML4RS}}) {{Workshop}}}, Feb. 2024.

\bibitem[Cracknell(2018)]{cracknell_development_2018}
A.~P. Cracknell.
\newblock The development of remote sensing in the last 40 years.
\newblock \emph{International Journal of Remote Sensing}, 39\penalty0 (23):\penalty0 8387--8427, Dec. 2018.
\newblock ISSN 0143-1161.
\newblock \doi{10.1080/01431161.2018.1550919}.

\bibitem[Crameri(2023)]{crameri_european_2023}
F.~Crameri.
\newblock European {{Ground Motion Service}}.
\newblock ESA, Oct. 2023.

\bibitem[Cui et~al.(2019)Cui, Zeng, Zhou, Xie, Wan, Hu, Xiong, Chen, Fan, and Hong]{cui_spatiotemporal_2019}
Y.~Cui, C.~Zeng, J.~Zhou, H.~Xie, W.~Wan, L.~Hu, W.~Xiong, X.~Chen, W.~Fan, and Y.~Hong.
\newblock A spatio-temporal continuous soil moisture dataset over the {{Tibet Plateau}} from 2002 to 2015.
\newblock \emph{Sci Data}, 6\penalty0 (1):\penalty0 247, Oct. 2019.
\newblock ISSN 2052-4463.
\newblock \doi{10.1038/s41597-019-0228-x}.

\bibitem[Dagobert(2022)]{dagobert_detection_2022}
T.~Dagobert.
\newblock Detection and interpretation of change in registered satellite image time series.
\newblock \emph{Image Processing On Line}, 12:\penalty0 625--651, 2022.

\bibitem[Daudt et~al.(2018{\natexlab{a}})Daudt, Le~Saux, and Boulch]{daudt_fully_2018}
R.~Daudt, B.~Le~Saux, and A.~Boulch.
\newblock Fully {{Convolutional Siamese Networks}} for {{Change Detection}}.
\newblock In \emph{2018 25th {{IEEE International Conference}} on {{Image Processing}} ({{ICIP}})}, pages 4063--4067, Oct. 2018{\natexlab{a}}.
\newblock \doi{10.1109/ICIP.2018.8451652}.

\bibitem[Daudt et~al.(2019)Daudt, Le~Saux, Boulch, and Gousseau]{daudt_multitask_2019}
R.~Daudt, B.~Le~Saux, A.~Boulch, and Y.~Gousseau.
\newblock Multitask learning for large-scale semantic change detection.
\newblock \emph{Computer Vision and Image Understanding}, 187:\penalty0 102783, Oct. 2019.
\newblock ISSN 1077-3142.
\newblock \doi{10.1016/j.cviu.2019.07.003}.

\bibitem[Daudt et~al.(2018{\natexlab{b}})Daudt, Le~Saux, Boulch, and Gousseau]{daudt_urban_2018}
R.~C. Daudt, B.~Le~Saux, A.~Boulch, and Y.~Gousseau.
\newblock Urban {{Change Detection}} for {{Multispectral Earth Observation Using Convolutional Neural Networks}}.
\newblock In \emph{{{IGARSS}} 2018 - 2018 {{IEEE International Geoscience}} and {{Remote Sensing Symposium}}}, pages 2115--2118, July 2018{\natexlab{b}}.
\newblock \doi{10.1109/IGARSS.2018.8518015}.

\bibitem[Daudt et~al.(2021)Daudt, Le~Saux, Boulch, and Gousseau]{daudt_weakly_2021}
R.~C. Daudt, B.~Le~Saux, A.~Boulch, and Y.~Gousseau.
\newblock Weakly supervised change detection using guided anisotropic diffusion.
\newblock \emph{Mach Learn}, June 2021.
\newblock ISSN 1573-0565.
\newblock \doi{10.1007/s10994-021-06008-4}.

\bibitem[De~G{\'e}lis et~al.(2021)De~G{\'e}lis, Lef{\`e}vre, Corpetti, Ristorcelli, Th{\'e}noz, and Lassalle]{degelis_benchmarking_2021}
I.~De~G{\'e}lis, S.~Lef{\`e}vre, T.~Corpetti, T.~Ristorcelli, C.~Th{\'e}noz, and P.~Lassalle.
\newblock Benchmarking {{Change Detection}} in {{Urban 3D Point Clouds}}.
\newblock In \emph{2021 {{IEEE International Geoscience}} and {{Remote Sensing Symposium IGARSS}}}, pages 3352--3355, July 2021.
\newblock \doi{10.1109/IGARSS47720.2021.9553018}.

\bibitem[De~G{\'e}lis et~al.(2024)De~G{\'e}lis, Corpetti, and Lef{\`e}vre]{degelis_change_2024}
I.~De~G{\'e}lis, T.~Corpetti, and S.~Lef{\`e}vre.
\newblock Change detection needs change information: Improving deep {{3D}} point cloud change detection.
\newblock \emph{IEEE Transactions on Geoscience and Remote Sensing}, 2024.

\bibitem[Decocq et~al.(2021)Decocq, Dupouey, and Berg{\`e}s]{decocq_dynamiques_2021}
G.~Decocq, J.-L. Dupouey, and L.~Berg{\`e}s.
\newblock {Dynamiques foresti{\`e}res {\`a} l'{\`e}re anthropoc{\`e}ne : mise au point s{\'e}mantique et proposition de d{\'e}finitions {\'e}cologiques}.
\newblock \emph{Revue foresti{\`e}re fran{\c c}aise}, 73\penalty0 (1):\penalty0 21--52, Dec. 2021.
\newblock ISSN 1951-6827.
\newblock \doi{10.20870/revforfr.2021.4993}.

\bibitem[Dee(2005)]{dee_bias_2005}
D.~P. Dee.
\newblock Bias and data assimilation.
\newblock \emph{Quarterly Journal of the Royal Meteorological Society}, 131\penalty0 (613):\penalty0 3323--3343, 2005.
\newblock ISSN 1477-870X.
\newblock \doi{10.1256/qj.05.137}.

\bibitem[Demir et~al.(2013)Demir, Bovolo, and Bruzzone]{demir_updating_2013}
B.~Demir, F.~Bovolo, and L.~Bruzzone.
\newblock Updating {{Land-Cover Maps}} by {{Classification}} of {{Image Time Series}}: {{A Novel Change-Detection-Driven Transfer Learning Approach}}.
\newblock \emph{IEEE Transactions on Geoscience and Remote Sensing}, 51\penalty0 (1):\penalty0 300--312, Jan. 2013.
\newblock ISSN 1558-0644.
\newblock \doi{10.1109/TGRS.2012.2195727}.

\bibitem[Derksen et~al.(2020)Derksen, Inglada, and Michel]{derksen_geometry_2020}
D.~Derksen, J.~Inglada, and J.~Michel.
\newblock Geometry {{Aware Evaluation}} of {{Handcrafted Superpixel-Based Features}} and {{Convolutional Neural Networks}} for {{Land Cover Mapping Using Satellite Imagery}}.
\newblock \emph{Remote Sensing}, 12:\penalty0 513, Feb. 2020.
\newblock \doi{10.3390/rs12030513}.

\bibitem[Deschamps and Sahbi(2022)]{deschamps_reinforcementbased_2022}
S.~Deschamps and H.~Sahbi.
\newblock Reinforcement-based frugal learning for satellite image change detection, Mar. 2022.

\bibitem[Dong et~al.(2024)Dong, Wang, Du, and Meng]{dong_changeclip_2024}
S.~Dong, L.~Wang, B.~Du, and X.~Meng.
\newblock {{ChangeCLIP}}: {{Remote}} sensing change detection with multimodal vision-language representation learning.
\newblock \emph{ISPRS Journal of Photogrammetry and Remote Sensing}, 208:\penalty0 53--69, Feb. 2024.
\newblock ISSN 0924-2716.
\newblock \doi{10.1016/j.isprsjprs.2024.01.004}.

\bibitem[Dong et~al.(2025)Dong, Wang, {deng}, Dai, Li, Liu, and Nong]{dong_crossdocument_2025}
Z.~Dong, M.~Wang, S.~{deng}, L.~Dai, J.~Li, X.~Liu, and R.~Nong.
\newblock Cross-{{Document Contextual Coreference Resolution}} in {{Knowledge Graphs}}, Apr. 2025.

\bibitem[Du et~al.(2002)Du, Teillet, and Cihlar]{du_radiometric_2002}
Y.~Du, P.~M. Teillet, and J.~Cihlar.
\newblock Radiometric normalization of multitemporal high-resolution satellite images with quality control for land cover change detection.
\newblock \emph{Remote Sensing of Environment}, 82\penalty0 (1):\penalty0 123--134, Sept. 2002.
\newblock ISSN 0034-4257.
\newblock \doi{10.1016/S0034-4257(02)00029-9}.

\bibitem[Dyce(2013)]{dyce_canada_2013}
M.~Dyce.
\newblock Canada between the photograph and the map: {{Aerial}} photography, geographical vision and the state.
\newblock \emph{Journal of Historical Geography}, 39:\penalty0 69--84, Jan. 2013.
\newblock ISSN 0305-7488.
\newblock \doi{10.1016/j.jhg.2012.07.002}.

\bibitem[Elgendy et~al.(2025)Elgendy, Sharshar, Aboeitta, Ashraf, and Guizani]{elgendy_geollava_2025}
H.~Elgendy, A.~Sharshar, A.~Aboeitta, Y.~Ashraf, and M.~Guizani.
\newblock {{GeoLLaVA}}: {{Efficient Fine-Tuned Vision-Language Models}} for {{Temporal Change Detection}} in {{Remote Sensing}}, May 2025.

\bibitem[{ESA}(2023)]{esa_using_2023}
{ESA}.
\newblock Using a data cube to monitor forest loss in the {{Amazon}}.
\newblock https://www.esa.int/Applications/Observing\_the\_Earth/Copernicus/Sentinel-1/Using\_a\_data\_cube\_to\_monitor\_forest\_loss\_in\_the\_Amazon, 2023.

\bibitem[Falconnier et~al.(2020)Falconnier, Corbeels, Boote, Affholder, Adam, MacCarthy, Ruane, Nendel, Whitbread, Justes, Ahuja, Akinseye, Alou, Amouzou, Anapalli, Baron, Basso, Baudron, Bertuzzi, Challinor, Chen, Deryng, Elsayed, Faye, Gaiser, Galdos, Gayler, Gerardeaux, Giner, Grant, Hoogenboom, Ibrahim, Kamali, Kersebaum, Kim, Laan, Leroux, Lizaso, Maestrini, Meier, Mequanint, Ndoli, Porter, Priesack, Ripoche, Sida, Singh, Smith, Srivastava, Sinha, Tao, Thorburn, Timlin, Traore, Twine, and Webber]{falconnier_modelling_2020}
G.~N. Falconnier, M.~Corbeels, K.~J. Boote, F.~Affholder, M.~Adam, D.~S. MacCarthy, A.~C. Ruane, C.~Nendel, A.~M. Whitbread, {\'E}.~Justes, L.~R. Ahuja, F.~M. Akinseye, I.~N. Alou, K.~A. Amouzou, S.~S. Anapalli, C.~Baron, B.~Basso, F.~Baudron, P.~Bertuzzi, A.~J. Challinor, Y.~Chen, D.~Deryng, M.~L. Elsayed, B.~Faye, T.~Gaiser, M.~Galdos, S.~Gayler, E.~Gerardeaux, M.~Giner, B.~Grant, G.~Hoogenboom, E.~S. Ibrahim, B.~Kamali, K.~C. Kersebaum, S.-H. Kim, M.~Laan, L.~Leroux, J.~I. Lizaso, B.~Maestrini, E.~A. Meier, F.~Mequanint, A.~Ndoli, C.~H. Porter, E.~Priesack, D.~Ripoche, T.~S. Sida, U.~Singh, W.~N. Smith, A.~Srivastava, S.~Sinha, F.~Tao, P.~J. Thorburn, D.~Timlin, B.~Traore, T.~Twine, and H.~Webber.
\newblock Modelling climate change impacts on maize yields under low nitrogen input conditions in sub-{{Saharan Africa}}.
\newblock \emph{Glob. Change Biol.}, 26\penalty0 (10):\penalty0 5942--5964, Oct. 2020.
\newblock ISSN 1354-1013, 1365-2486.
\newblock \doi{10.1111/gcb.15261}.

\bibitem[Fan et~al.(2016)Fan, Yang, Zipf, and Rousell]{fan_polygonbased_2016}
H.~Fan, B.~Yang, A.~Zipf, and A.~Rousell.
\newblock A polygon-based approach for matching {{OpenStreetMap}} road networks with regional transit authority data.
\newblock \emph{International Journal of Geographical Information Science}, 30\penalty0 (4):\penalty0 748--764, Apr. 2016.
\newblock ISSN 1365-8816.
\newblock \doi{10.1080/13658816.2015.1100732}.

\bibitem[Fang et~al.(2023)Fang, Li, and Li]{fang_changer_2023}
S.~Fang, K.~Li, and Z.~Li.
\newblock Changer: {{Feature Interaction}} is {{What You Need}} for {{Change Detection}}.
\newblock \emph{IEEE Trans. Geosci. Remote Sensing}, 61:\penalty0 1--11, 2023.
\newblock ISSN 0196-2892, 1558-0644.
\newblock \doi{10.1109/TGRS.2023.3277496}.

\bibitem[Farah et~al.(2024)Farah, Bachyr, Cleuziou, Halftermeyer, Gracianne, Auclair, Hafiane, and Canals]{farah_mcatnat_2024}
B.~Farah, O.~E. Bachyr, G.~Cleuziou, A.~Halftermeyer, C.~Gracianne, S.~Auclair, A.~Hafiane, and R.~Canals.
\newblock M-{{CATNAT}}: {{A Multimodal}} dataset to analyze {{French}} tweets during natural disasters.
\newblock \emph{Proceedings of the International ISCRAM Conference}, May 2024.
\newblock ISSN 2411-3387.
\newblock \doi{10.59297/yhq8bb90}.

\bibitem[Feng et~al.(2015)Feng, Tian, Zhang, Zhang, Wan, and Sun]{feng_finegrained_2015}
W.~Feng, F.-P. Tian, Q.~Zhang, N.~Zhang, L.~Wan, and J.~Sun.
\newblock Fine-{{Grained Change Detection}} of {{Misaligned Scenes With Varied Illuminations}}.
\newblock In \emph{Proceedings of the {{IEEE International Conference}} on {{Computer Vision}}}, pages 1260--1268, 2015.

\bibitem[Ferraris(2018)]{ferraris_detection_2018}
V.~Ferraris.
\newblock \emph{D{\'e}tection de Changement Par Fusion d'images de T{\'e}l{\'e}d{\'e}tection de R{\'e}solutions et Modalit{\'e}s Diff{\'e}rentes}.
\newblock PhD thesis, Institut National Polytechnique de Toulouse, Oct. 2018.

\bibitem[Ferraris et~al.(2020)Ferraris, Dobigeon, and Chabert]{ferraris_robust_2020}
V.~Ferraris, N.~Dobigeon, and M.~Chabert.
\newblock Robust fusion algorithms for unsupervised change detection between multi-band optical images --- {{A}} comprehensive case study.
\newblock \emph{Information Fusion}, 64:\penalty0 293--317, Dec. 2020.
\newblock ISSN 1566-2535.
\newblock \doi{10.1016/j.inffus.2020.08.008}.

\bibitem[Figari~Tomenotti(2021)]{figaritomenotti_heterogeneous_2021}
F.~Figari~Tomenotti.
\newblock Heterogeneous {{Change Detection}} on {{Remote Sensing Data}} with {{Self-Supervised Deep Canonically Correlated Autoencoders}}.
\newblock \emph{Septentrio Reports}, Mar. 2021.
\newblock \doi{10.7557/7.5763}.

\bibitem[Florath et~al.(2024)Florath, Chanussot, and Keller]{florath_rapid_2024}
J.~Florath, J.~Chanussot, and S.~Keller.
\newblock Rapid natural hazard extent estimation from twitter data: Investigation for hurricane impact areas.
\newblock \emph{Nat Hazards}, Mar. 2024.
\newblock ISSN 1573-0840.
\newblock \doi{10.1007/s11069-024-06488-2}.

\bibitem[{Food and FAO Agriculture Organization}(2015)]{foodandfaoagricultureorganization_status_2015}
{Food and FAO Agriculture Organization}.
\newblock Status of the {{World}}'s {{Soil Resources}}: {{Main Report}}.
\newblock Technical report, United Nations, 2015.

\bibitem[Foresta et~al.(2018)Foresta, Gourmelen, Weissgerber, Nienow, Williams, Shepherd, Drinkwater, and Plummer]{foresta_heterogeneous_2018}
L.~Foresta, N.~Gourmelen, F.~Weissgerber, P.~Nienow, J.~J. Williams, A.~Shepherd, M.~R. Drinkwater, and S.~Plummer.
\newblock Heterogeneous and rapid ice loss over the {{Patagonian Ice Fields}} revealed by {{CryoSat-2}} swath radar altimetry.
\newblock \emph{Remote Sensing of Environment}, 211:\penalty0 441--455, June 2018.
\newblock ISSN 0034-4257.
\newblock \doi{10.1016/j.rse.2018.03.041}.

\bibitem[Freudiger et~al.(2018)Freudiger, Mennekes, Seibert, and Weiler]{freudiger_historical_2018}
D.~Freudiger, D.~Mennekes, J.~Seibert, and M.~Weiler.
\newblock Historical glacier outlines from digitized topographic maps of the {{Swiss Alps}}.
\newblock \emph{Earth System Science Data}, 10\penalty0 (2):\penalty0 805--814, Apr. 2018.
\newblock ISSN 1866-3508.
\newblock \doi{10.5194/essd-10-805-2018}.

\bibitem[Gandhi et~al.(2015)Gandhi, Parthiban, Thummalu, and Christy]{gandhi_ndvi_2015}
G.~M. Gandhi, S.~Parthiban, N.~Thummalu, and A.~Christy.
\newblock Ndvi: {{Vegetation Change Detection Using Remote Sensing}} and {{Gis}} -- {{A Case Study}} of {{Vellore District}}.
\newblock \emph{Procedia Computer Science}, 57:\penalty0 1199--1210, Jan. 2015.
\newblock ISSN 1877-0509.
\newblock \doi{10.1016/j.procs.2015.07.415}.

\bibitem[Garioud et~al.(2023)Garioud, Gonthier, Landrieu, De~Wit, Valette, Poup{\'e}e, Giordano, and Wattrelos]{garioud_flair_2023a}
A.~Garioud, N.~Gonthier, L.~Landrieu, A.~De~Wit, M.~Valette, M.~Poup{\'e}e, S.~Giordano, and B.~Wattrelos.
\newblock {{FLAIR}} : A {{Country-Scale Land Cover Semantic Segmentation Dataset From Multi-Source Optical Imagery}}.
\newblock \emph{Advances in Neural Information Processing Systems}, 36:\penalty0 16456--16482, Dec. 2023.

\bibitem[Gazeaux et~al.(2013)Gazeaux, Williams, King, Bos, Dach, Deo, Moore, Ostini, Petrie, Roggero, Teferle, Olivares, and Webb]{gazeaux_detecting_2013}
J.~Gazeaux, S.~Williams, M.~King, M.~Bos, R.~Dach, M.~Deo, A.~W. Moore, L.~Ostini, E.~Petrie, M.~Roggero, F.~N. Teferle, G.~Olivares, and F.~H. Webb.
\newblock Detecting offsets in {{GPS}} time series: {{First}} results from the detection of offsets in {{GPS}} experiment.
\newblock \emph{Journal of Geophysical Research: Solid Earth}, 118\penalty0 (5):\penalty0 2397--2407, 2013.
\newblock ISSN 2169-9356.
\newblock \doi{10.1002/jgrb.50152}.

\bibitem[Gazeaux et~al.(2015)Gazeaux, Lebarbier, Collilieux, and M{\'e}tivier]{gazeaux_joint_2015}
J.~Gazeaux, E.~Lebarbier, X.~Collilieux, and L.~M{\'e}tivier.
\newblock {Joint segmentation of multiple GPS coordinate series}.
\newblock \emph{Journal de la soci{\'e}t{\'e} fran{\c c}aise de statistique}, 156\penalty0 (4):\penalty0 163--179, 2015.
\newblock ISSN 2102-6238.

\bibitem[Gesnouin et~al.(2024)Gesnouin, Tannier, Da~Silva, Tapory, Brier, Simon, Rozenberg, Woehrel, Yakaabi, Binder, Marie, Caron, Nogueira, Fontas, Puydebois, Theophile, Morandi, Petit, Creissac, Ennouchy, Valetoux, Visade, Balloux, Cortes, Devineau, Tan, Mac~Namara, and Yang]{gesnouin_llamandement_2024}
J.~Gesnouin, Y.~Tannier, C.~G. Da~Silva, H.~Tapory, C.~Brier, H.~Simon, R.~Rozenberg, H.~Woehrel, M.~E. Yakaabi, T.~Binder, G.~Marie, E.~Caron, M.~Nogueira, T.~Fontas, L.~Puydebois, M.~Theophile, S.~Morandi, M.~Petit, D.~Creissac, P.~Ennouchy, E.~Valetoux, C.~Visade, S.~Balloux, E.~Cortes, P.-E. Devineau, U.~Tan, E.~Mac~Namara, and S.~Yang.
\newblock {{LLaMandement}}: {{Large Language Models}} for {{Summarization}} of {{French Legislative Proposals}}, Jan. 2024.

\bibitem[Giordano et~al.(2018)Giordano, Le~Bris, and Mallet]{giordano_automatic_2018}
S.~Giordano, A.~Le~Bris, and C.~Mallet.
\newblock Toward {{Automatic Georeferencing}} of {{Archival Aerial Photogrammetric Surveys}}.
\newblock \emph{ISPRS Annals of the Photogrammetry, Remote Sensing and Spatial Information Sciences}, IV-2:\penalty0 105--112, May 2018.
\newblock ISSN 2194-9042.
\newblock \doi{10.5194/isprs-annals-IV-2-105-2018}.

\bibitem[{G{\'o}mez-Dans} et~al.(2022){G{\'o}mez-Dans}, Lewis, Yin, Asare, Lamptey, Aidoo, MacCarthy, Ma, Wu, Addi, {Aboagye-Ntow}, Doe, Alhassan, {Kankam-Boadu}, Huang, and Li]{gomez-dans_location_2022}
J.~L. {G{\'o}mez-Dans}, P.~E. Lewis, F.~Yin, K.~Asare, P.~Lamptey, K.~K.~Y. Aidoo, D.~S. MacCarthy, H.~Ma, Q.~Wu, M.~Addi, S.~{Aboagye-Ntow}, C.~E. Doe, R.~Alhassan, I.~{Kankam-Boadu}, J.~Huang, and X.~Li.
\newblock Location, biophysical and agronomic parameters for croplands in northern {{Ghana}}.
\newblock \emph{Earth System Science Data}, 14\penalty0 (12):\penalty0 5387--5410, Dec. 2022.
\newblock ISSN 1866-3508.
\newblock \doi{10.5194/essd-14-5387-2022}.

\bibitem[Gominski et~al.(2019)Gominski, Poreba, {Gouet-Brunet}, and Chen]{gominski_challenging_2019}
D.~Gominski, M.~Poreba, V.~{Gouet-Brunet}, and L.~Chen.
\newblock Challenging deep image descriptors for retrieval in heterogeneous iconographic collections.
\newblock In \emph{Proceedings of the 1st Workshop on {{Structuring}} and {{Understanding}} of {{Multimedia heritAge Contents}} ({{SUMAC}}'19 {{Workshop}} @ {{ACM Multimedia}} 2019)}, Nice, France, Oct. 2019.
\newblock \doi{10.1145/3347317.3357246}.

\bibitem[Gordon(1980)]{gordon_utilizing_1980}
S.~I. Gordon.
\newblock Utilizing {{LANDSAT}} imagery to monitor land-use change: {{A}} case study in {{Ohio}}.
\newblock \emph{Remote Sensing of Environment}, 9\penalty0 (3):\penalty0 189--196, May 1980.
\newblock ISSN 0034-4257.
\newblock \doi{10.1016/0034-4257(80)90028-0}.

\bibitem[Gressin et~al.(2013{\natexlab{a}})Gressin, Mallet, Vincent, and Paparoditis]{gressin_updating_2013}
A.~Gressin, C.~Mallet, N.~Vincent, and N.~Paparoditis.
\newblock Updating land cover databases using a single very high resolution satellite image.
\newblock \emph{ISPRS Ann. Photogramm. Remote Sens. Spatial Inf. Sci.}, II-3/W2:\penalty0 13--18, Oct. 2013{\natexlab{a}}.
\newblock ISSN 2194-9050.
\newblock \doi{10.5194/isprsannals-II-3-W2-13-2013}.

\bibitem[Gressin et~al.(2013{\natexlab{b}})Gressin, Vincent, Mallet, and Paparoditis]{gressin_semantic_2013}
A.~Gressin, N.~Vincent, C.~Mallet, and N.~Paparoditis.
\newblock Semantic {{Approach}} in {{Image Change Detection}}.
\newblock In J.~{Blanc-Talon}, A.~Kasinski, W.~Philips, D.~Popescu, and P.~Scheunders, editors, \emph{Advanced {{Concepts}} for {{Intelligent Vision Systems}}}, volume 8192, pages 450--459. Springer International Publishing, Cham, 2013{\natexlab{b}}.
\newblock \doi{10.1007/978-3-319-02895-8_40}.

\bibitem[Gressin et~al.(2014)Gressin, Mallet, Vincent, and Paparoditis]{gressin_updating_2014}
A.~Gressin, C.~Mallet, N.~Vincent, and N.~Paparoditis.
\newblock Updating the new {{French}} national land cover database.
\newblock In \emph{2014 {{IEEE Geoscience}} and {{Remote Sensing Symposium}}}, pages 3534--3537, July 2014.
\newblock \doi{10.1109/IGARSS.2014.6947245}.

\bibitem[Groot and Kraak(1999)]{groot_challenges_1999}
R.~Groot and M.-J. Kraak.
\newblock Challenges and opportunities for national mapping agencies development of national geospatial data infrastructure ({{NGDI}}).
\newblock In \emph{First Meeting of the Committee on {{Development Information}} ({{CODI}}) of the {{United Nations}}, {{Addis Ababa}}, {{Ethiopia}}}, June 1999.

\bibitem[Guerin et~al.(2014)Guerin, Binet, and {Pierrot-Deseilligny}]{guerin_automatic_2014}
C.~Guerin, R.~Binet, and M.~{Pierrot-Deseilligny}.
\newblock Automatic {{Detection}} of {{Elevation Changes}} by {{Differential DSM Analysis}}: {{Application}} to {{Urban Areas}}.
\newblock \emph{IEEE Journal of Selected Topics in Applied Earth Observations and Remote Sensing}, 7\penalty0 (10):\penalty0 4020--4037, Oct. 2014.
\newblock ISSN 2151-1535.
\newblock \doi{10.1109/JSTARS.2014.2300509}.

\bibitem[Guilcher et~al.(2022)Guilcher, {Olteanu-Raimond}, and Balde]{guilcher_analysis_2022}
A.~L. Guilcher, A.-M. {Olteanu-Raimond}, and M.~B. Balde.
\newblock Analysis of massive imports of open data in {{Openstreetmap}} database: {{A}} study case for {{France}}.
\newblock \emph{ISPRS Annals of the Photogrammetry, Remote Sensing and Spatial Information Sciences}, V-4-2022:\penalty0 99, 2022.
\newblock \doi{10.5194/isprs-annals-V-4-2022-99-2022}.

\bibitem[Guo(2010)]{guo_understanding_2010}
H.~Guo.
\newblock Understanding global natural disasters and the role of earth observation.
\newblock \emph{International Journal of Digital Earth}, Sept. 2010.
\newblock ISSN 1753-8947.

\bibitem[Guo et~al.(2019)Guo, Peng, Zhu, Wang, Wang, Peng, and He]{guo_modelling_2019}
Y.~Guo, C.~Peng, Q.~Zhu, M.~Wang, H.~Wang, S.~Peng, and H.~He.
\newblock Modelling the impacts of climate and land use changes on soil water erosion: {{Model}} applications, limitations and future challenges.
\newblock \emph{Journal of Environmental Management}, 250:\penalty0 109403, Nov. 2019.
\newblock ISSN 0301-4797.
\newblock \doi{10.1016/j.jenvman.2019.109403}.

\bibitem[Gupta et~al.(2019)Gupta, Goodman, Patel, Hosfelt, Sajeev, Heim, Doshi, Lucas, Choset, and Gaston]{gupta_creating_2019}
R.~Gupta, B.~Goodman, N.~Patel, R.~Hosfelt, S.~Sajeev, E.~Heim, J.~Doshi, K.~Lucas, H.~Choset, and M.~Gaston.
\newblock Creating {{xBD}}: {{A Dataset}} for {{Assessing Building Damage}} from {{Satellite Imagery}}.
\newblock In \emph{Proceedings of the {{IEEE}}/{{CVF Conference}} on {{Computer Vision}} and {{Pattern Recognition Workshops}}}, pages 10--17, 2019.

\bibitem[Guyet and Nicolas(2016)]{guyet_long_2016}
T.~Guyet and H.~Nicolas.
\newblock Long term analysis of time series of satellite images.
\newblock \emph{Pattern Recognition Letters}, 70:\penalty0 17--23, Jan. 2016.
\newblock ISSN 01678655.
\newblock \doi{10.1016/j.patrec.2015.11.005}.

\bibitem[Hallegatte et~al.(2016)Hallegatte, Rogelj, Allen, Clarke, Edenhofer, Field, Friedlingstein, {van Kesteren}, Knutti, Mach, Mastrandrea, Michel, Minx, Oppenheimer, Plattner, Riahi, Schaeffer, Stocker, and {van Vuuren}]{hallegatte_mapping_2016}
S.~Hallegatte, J.~Rogelj, M.~Allen, L.~Clarke, O.~Edenhofer, C.~B. Field, P.~Friedlingstein, L.~{van Kesteren}, R.~Knutti, K.~J. Mach, M.~Mastrandrea, A.~Michel, J.~Minx, M.~Oppenheimer, G.-K. Plattner, K.~Riahi, M.~Schaeffer, T.~F. Stocker, and D.~P. {van Vuuren}.
\newblock Mapping the climate change challenge.
\newblock \emph{Nature Clim Change}, 6\penalty0 (7):\penalty0 663--668, July 2016.
\newblock ISSN 1758-6798.
\newblock \doi{10.1038/nclimate3057}.

\bibitem[Hapuarachchi et~al.(2011)Hapuarachchi, Wang, and Pagano]{hapuarachchi_review_2011}
H.~a.~P. Hapuarachchi, Q.~J. Wang, and T.~C. Pagano.
\newblock A review of advances in flash flood forecasting.
\newblock \emph{Hydrological Processes}, 25\penalty0 (18):\penalty0 2771--2784, 2011.
\newblock ISSN 1099-1085.
\newblock \doi{10.1002/hyp.8040}.

\bibitem[Hauck et~al.(2016)Hauck, Schmidt, and Werner]{hauck_using_2016}
J.~Hauck, J.~Schmidt, and A.~Werner.
\newblock Using social network analysis to identify key stakeholders in agricultural biodiversity governance and related land-use decisions at regional and local level.
\newblock \emph{Ecology and Society}, 21\penalty0 (2), 2016.
\newblock ISSN 1708-3087.

\bibitem[He et~al.(2024)He, Zhang, Li, Zhang, Luo, Li, and Zhang]{he_changeguided_2024}
L.~He, M.~Zhang, Y.~Li, J.~Zhang, S.~Luo, S.~Li, and X.~Zhang.
\newblock Change-{{Guided Similarity Pyramid Network}} for {{Semantic Change Detection}}.
\newblock \emph{IEEE Transactions on Geoscience and Remote Sensing}, pages 1--1, 2024.
\newblock ISSN 1558-0644.
\newblock \doi{10.1109/TGRS.2024.3434451}.

\bibitem[Hechtlinger et~al.(2018)Hechtlinger, P{\'o}czos, and Wasserman]{hechtlinger_cautious_2018}
Y.~Hechtlinger, B.~P{\'o}czos, and L.~Wasserman.
\newblock Cautious {{Deep Learning}}.
\newblock \emph{arXiv:1805.09460 [cs, stat]}, May 2018.

\bibitem[Heidemann et~al.(2012)Heidemann, Klier, and Probst]{heidemann_online_2012}
J.~Heidemann, M.~Klier, and F.~Probst.
\newblock Online social networks: {{A}} survey of a global phenomenon.
\newblock \emph{Computer Networks}, 56\penalty0 (18):\penalty0 3866--3878, Dec. 2012.
\newblock ISSN 1389-1286.
\newblock \doi{10.1016/j.comnet.2012.08.009}.

\bibitem[Hojte et~al.(2023)Hojte, Johansen, Fraas, and Flatz]{concito_impact_2023}
S.~Hojte, A.~B. Johansen, E.~Fraas, and J.~Flatz.
\newblock Impact and {{Opportunities}} of the 2023-27 {{CAP Reform}} in {{Denmark}}.
\newblock Technical report, Concito, 2023.

\bibitem[Horn(1975)]{horn_forest_1975}
H.~S. Horn.
\newblock Forest {{Succession}}.
\newblock \emph{Scientific American}, 232\penalty0 (5):\penalty0 90--101, 1975.
\newblock ISSN 0036-8733.

\bibitem[Hornacek et~al.(2012)Hornacek, Wagner, Sabel, Truong, Snoeij, Hahmann, Diedrich, and Doubkova]{hornacek_potential_2012}
M.~Hornacek, W.~Wagner, D.~Sabel, H.-L. Truong, P.~Snoeij, T.~Hahmann, E.~Diedrich, and M.~Doubkova.
\newblock Potential for {{High Resolution Systematic Global Surface Soil Moisture Retrieval}} via {{Change Detection Using Sentinel-1}}.
\newblock \emph{IEEE Journal of Selected Topics in Applied Earth Observations and Remote Sensing}, 5\penalty0 (4):\penalty0 1303--1311, Aug. 2012.
\newblock ISSN 2151-1535.
\newblock \doi{10.1109/JSTARS.2012.2190136}.

\bibitem[Hoshiba and Aoki(2015)]{hoshiba_numerical_2015}
M.~Hoshiba and S.~Aoki.
\newblock Numerical {{Shake Prediction}} for {{Earthquake Early Warning}}: {{Data Assimilation}}, {{Real}}-{{Time Shake Mapping}}, and {{Simulation}} of {{Wave Propagation}}.
\newblock \emph{Bulletin of the Seismological Society of America}, 105\penalty0 (3):\penalty0 1324--1338, May 2015.
\newblock ISSN 0037-1106.
\newblock \doi{10.1785/0120140280}.

\bibitem[Howarth and Wickware(1981)]{howarth_procedures_1981}
P.~J. Howarth and G.~M. Wickware.
\newblock Procedures for change detection using {{Landsat}} digital data.
\newblock \emph{International Journal of Remote Sensing}, 2\penalty0 (3):\penalty0 277--291, July 1981.
\newblock ISSN 0143-1161.
\newblock \doi{10.1080/01431168108948362}.

\bibitem[Hu et~al.(2014)Hu, Ge, and Hou]{hu_using_2014}
H.~Hu, Y.~Ge, and D.~Hou.
\newblock Using {{Web Crawler Technology}} for {{Geo-Events Analysis}}: {{A Case Study}} of the {{Huangyan Island Incident}}.
\newblock \emph{Sustainability}, 6\penalty0 (4):\penalty0 1896--1912, Apr. 2014.
\newblock ISSN 2071-1050.
\newblock \doi{10.3390/su6041896}.

\bibitem[Huang et~al.(2025)Huang, Yu, Chen, Yokoya, Li, and Plaza]{huang_sar_2025}
H.~Huang, Y.~Yu, T.~Chen, N.~Yokoya, J.~Li, and A.~Plaza.
\newblock {{SAR}} and {{Social Media-Based Change Detection}} with {{Dual Threshold Fusion}} for {{Flood Inundation Mapping}}.
\newblock \emph{IEEE Journal of Selected Topics in Applied Earth Observations and Remote Sensing}, pages 1--13, 2025.
\newblock ISSN 2151-1535.
\newblock \doi{10.1109/JSTARS.2025.3579062}.

\bibitem[Irvin et~al.(2025)Irvin, Liu, Chen, Dormoy, Kim, Khanna, Zheng, and Ermon]{irvin_teochat_2025}
J.~A. Irvin, E.~R. Liu, J.~C. Chen, I.~Dormoy, J.~Kim, S.~Khanna, Z.~Zheng, and S.~Ermon.
\newblock {{TEOChat}}: {{A Large Vision-Language Assistant}} for {{Temporal Earth Observation Data}}.
\newblock In \emph{{{ICLR}}}. arXiv, Jan. 2025.
\newblock \doi{10.48550/arXiv.2410.06234}.

\bibitem[Ivanovic et~al.(2019)Ivanovic, {Olteanu-Raimond}, Musti{\`e}re, and Devogele]{ivanovic_filteringbased_2019}
S.~S. Ivanovic, A.-M. {Olteanu-Raimond}, S.~Musti{\`e}re, and T.~Devogele.
\newblock A {{Filtering-Based Approach}} for {{Improving Crowdsourced GNSS Traces}} in a {{Data Update Context}}.
\newblock \emph{ISPRS International Journal of Geo-Information}, 8\penalty0 (9):\penalty0 380, Sept. 2019.
\newblock ISSN 2220-9964.
\newblock \doi{10.3390/ijgi8090380}.

\bibitem[Ivanovic et~al.(2020)Ivanovic, {Olteanu-Raimond}, Musti{\`e}re, and Devogele]{ivanovic_potential_2020}
S.~S. Ivanovic, A.-M. {Olteanu-Raimond}, S.~Musti{\`e}re, and T.~Devogele.
\newblock Potential of {{Crowdsourced Traces}} for {{Detecting Updates}} in {{Authoritative Geographic Data}}.
\newblock In P.~Kyriakidis, D.~Hadjimitsis, D.~Skarlatos, and A.~Mansourian, editors, \emph{Geospatial {{Technologies}} for {{Local}} and {{Regional Development}}}, Lecture {{Notes}} in {{Geoinformation}} and {{Cartography}}, pages 205--221, Cham, 2020. Springer International Publishing.
\newblock ISBN 978-3-030-14745-7.
\newblock \doi{10.1007/978-3-030-14745-7_12}.

\bibitem[Jakubik et~al.(2023)Jakubik, Roy, Phillips, Fraccaro, Godwin, Zadrozny, Szwarcman, Gomes, Nyirjesy, Edwards, Kimura, Simumba, Chu, Mukkavilli, Lambhate, Das, Bangalore, Oliveira, Muszynski, Ankur, Ramasubramanian, Gurung, Khallaghi, Hanxi, Li, Cecil, Ahmadi, Kordi, Alemohammad, Maskey, Ganti, Weldemariam, and Ramachandran]{jakubik_foundation_2023}
J.~Jakubik, S.~Roy, C.~E. Phillips, P.~Fraccaro, D.~Godwin, B.~Zadrozny, D.~Szwarcman, C.~Gomes, G.~Nyirjesy, B.~Edwards, D.~Kimura, N.~Simumba, L.~Chu, S.~K. Mukkavilli, D.~Lambhate, K.~Das, R.~Bangalore, D.~Oliveira, M.~Muszynski, K.~Ankur, M.~Ramasubramanian, I.~Gurung, S.~Khallaghi, Hanxi, Li, M.~Cecil, M.~Ahmadi, F.~Kordi, H.~Alemohammad, M.~Maskey, R.~Ganti, K.~Weldemariam, and R.~Ramachandran.
\newblock Foundation {{Models}} for {{Generalist Geospatial Artificial Intelligence}}, Nov. 2023.

\bibitem[James et~al.(2012)James, Hodgson, Ghoshal, and Latiolais]{james_geomorphic_2012}
L.~A. James, M.~E. Hodgson, S.~Ghoshal, and M.~M. Latiolais.
\newblock Geomorphic change detection using historic maps and {{DEM}} differencing: {{The}} temporal dimension of geospatial analysis.
\newblock \emph{Geomorphology}, 137\penalty0 (1):\penalty0 181--198, Jan. 2012.
\newblock ISSN 0169-555X.
\newblock \doi{10.1016/j.geomorph.2010.10.039}.

\bibitem[Jayawardene et~al.(2021)Jayawardene, Huggins, Prasanna, and Fakhruddin]{jayawardene_role_2021}
V.~Jayawardene, T.~J. Huggins, R.~Prasanna, and B.~Fakhruddin.
\newblock The role of data and information quality during disaster response decision-making.
\newblock \emph{Progress in Disaster Science}, 12:\penalty0 100202, Dec. 2021.
\newblock ISSN 2590-0617.
\newblock \doi{10.1016/j.pdisas.2021.100202}.

\bibitem[Ji et~al.(2019)Ji, Wei, and Lu]{ji_fully_2019}
S.~Ji, S.~Wei, and M.~Lu.
\newblock Fully {{Convolutional Networks}} for {{Multisource Building Extraction From}} an {{Open Aerial}} and {{Satellite Imagery Data Set}}.
\newblock \emph{IEEE Transactions on Geoscience and Remote Sensing}, 57\penalty0 (1):\penalty0 574--586, Jan. 2019.
\newblock ISSN 1558-0644.
\newblock \doi{10.1109/TGRS.2018.2858817}.

\bibitem[Jin et~al.(2013)Jin, Yang, Danielson, Homer, Fry, and Xian]{jin_comprehensive_2013}
S.~Jin, L.~Yang, P.~Danielson, C.~Homer, J.~Fry, and G.~Xian.
\newblock A comprehensive change detection method for updating the {{National Land Cover Database}} to circa 2011.
\newblock \emph{Remote Sensing of Environment}, 132:\penalty0 159--175, May 2013.
\newblock ISSN 0034-4257.
\newblock \doi{10.1016/j.rse.2013.01.012}.

\bibitem[{Kadri-Dahmani}(2001)]{kadri-dahmani_updating_2001}
H.~{Kadri-Dahmani}.
\newblock Updating {{Data}} in {{GIS}} : {{Towards}} a more generic approach.
\newblock \emph{ICC Proceedings}, 3:\penalty0 6--10, 2001.

\bibitem[{Kadri-Dahmani} et~al.(2006){Kadri-Dahmani}, Bertelle, Duchamp, and Osmani]{kadri-dahmani_consistent_2008}
H.~{Kadri-Dahmani}, C.~Bertelle, G.~H. Duchamp, and A.~Osmani.
\newblock Consistent {{Updating}} of {{Geographical DataBase}} as {{Emergent Property}} over {{Influence System}}.
\newblock \emph{International Journal of Modelling, Identification and Control}, 3\penalty0 (1):\penalty0 58--68, 2006.

\bibitem[Kang and Lu(2011)]{kang_change_2011}
Z.~Kang and Z.~Lu.
\newblock The {{Change Detection}} of {{Building Models Using Epochs}} of {{Terrestrial Point Clouds}}.
\newblock In \emph{2011 {{International Workshop}} on {{Multi-Platform}}/{{Multi-Sensor Remote Sensing}} and {{Mapping}}}, pages 1--6, Xiamen, China, Jan. 2011. IEEE.
\newblock ISBN 978-1-4244-9402-6.
\newblock \doi{10.1109/M2RSM.2011.5697381}.

\bibitem[Kauppinen and Hyv{\"o}nen(2007)]{kauppinen_modeling_2007}
T.~Kauppinen and E.~Hyv{\"o}nen.
\newblock Modeling and {{Reasoning About Changes}} in {{Ontology Time Series}}.
\newblock volume~14, pages 319--338. Jan. 2007.
\newblock ISBN 978-0-387-37019-4.
\newblock \doi{10.1007/978-0-387-37022-4_11}.

\bibitem[Kawamura(1971)]{kawamura_automatic_1971}
J.~G. Kawamura.
\newblock Automatic recognition of changes in urban development from aerial photographs.
\newblock \emph{IEEE Transactions on Systems, Man, \& Cybernetics}, SMC-1\penalty0 (3):\penalty0 230--239, 1971.
\newblock ISSN 0018-9472.
\newblock \doi{10.1109/TSMC.1971.4308290}.

\bibitem[Kennedy et~al.(2007)Kennedy, Cohen, and Schroeder]{kennedy_trajectorybased_2007}
R.~E. Kennedy, W.~B. Cohen, and T.~A. Schroeder.
\newblock Trajectory-based change detection for automated characterization of forest disturbance dynamics.
\newblock \emph{Remote Sensing of Environment}, 110\penalty0 (3):\penalty0 370--386, Oct. 2007.
\newblock ISSN 0034-4257.
\newblock \doi{10.1016/j.rse.2007.03.010}.

\bibitem[Kennedy et~al.(2010)Kennedy, Yang, and Cohen]{kennedy_detecting_2010}
R.~E. Kennedy, Z.~Yang, and W.~B. Cohen.
\newblock Detecting trends in forest disturbance and recovery using yearly {{Landsat}} time series: 1. {{LandTrendr}} --- {{Temporal}} segmentation algorithms.
\newblock \emph{Remote Sensing of Environment}, 114\penalty0 (12):\penalty0 2897--2910, Dec. 2010.
\newblock ISSN 0034-4257.
\newblock \doi{10.1016/j.rse.2010.07.008}.

\bibitem[Kharroubi et~al.(2022)Kharroubi, Poux, Ballouch, Hajji, and Billen]{kharroubi_three_2022}
A.~Kharroubi, F.~Poux, Z.~Ballouch, R.~Hajji, and R.~Billen.
\newblock Three {{Dimensional Change Detection Using Point Clouds}}: {{A Review}}.
\newblock \emph{Geomatics}, 2\penalty0 (4):\penalty0 457--485, Dec. 2022.
\newblock ISSN 2673-7418.
\newblock \doi{10.3390/geomatics2040025}.

\bibitem[Khelifi and Mignotte(2020)]{khelifi_deep_2020}
L.~Khelifi and M.~Mignotte.
\newblock Deep {{Learning}} for {{Change Detection}} in {{Remote Sensing Images}}: {{Comprehensive Review}} and {{Meta-Analysis}}.
\newblock \emph{IEEE Access}, 8:\penalty0 126385--126400, 2020.
\newblock ISSN 2169-3536.
\newblock \doi{10.1109/ACCESS.2020.3008036}.

\bibitem[Kim and Kim(2025)]{kim_generalizable_2025}
J.~Kim and U.~Kim.
\newblock Towards {{Generalizable Scene Change Detection}}, Mar. 2025.

\bibitem[Kirillov et~al.(2023)Kirillov, Mintun, Ravi, Mao, Rolland, Gustafson, Xiao, Whitehead, Berg, Lo, Doll{\'a}r, and Girshick]{kirillov_segment_2023}
A.~Kirillov, E.~Mintun, N.~Ravi, H.~Mao, C.~Rolland, L.~Gustafson, T.~Xiao, S.~Whitehead, A.~C. Berg, W.-Y. Lo, P.~Doll{\'a}r, and R.~Girshick.
\newblock Segment {{Anything}}.
\newblock In \emph{Proceedings of the {{IEEE}}/{{CVF International Conference}} on {{Computer Vision}}}, pages 4015--4026, Apr. 2023.
\newblock \doi{10.48550/arXiv.2304.02643}.

\bibitem[Knoefel et~al.(2021)Knoefel, Herrmann, Sindram, and Hovenbitzer]{knoefel_germanys_2021}
P.~Knoefel, D.~Herrmann, M.~Sindram, and M.~Hovenbitzer.
\newblock Germanys {{First Cloud-based Web Service For Land Monitoring Using Copernicus Sentinel-2 Data}}.
\newblock \emph{ISPRS Ann. Photogramm. Remote Sens. Spatial Inf. Sci.}, V-3-2021:\penalty0 133--140, June 2021.
\newblock ISSN 2194-9050.
\newblock \doi{10.5194/isprs-annals-V-3-2021-133-2021}.

\bibitem[Knudsen and Olsen(2003)]{knudsen_automated_2003}
T.~Knudsen and B.~P. Olsen.
\newblock Automated {{Change Detection}} for {{Updates}} of {{Digital Map Databases}}.
\newblock \emph{photogramm eng remote sensing}, 69\penalty0 (11):\penalty0 1289--1296, Nov. 2003.
\newblock ISSN 00991112.
\newblock \doi{10.14358/PERS.69.11.1289}.

\bibitem[Kulkarni and Michels(2012)]{kulkarni_temporal_2012}
K.~Kulkarni and J.-E. Michels.
\newblock Temporal features in {{SQL}}:2011.
\newblock \emph{SIGMOD Rec.}, 41\penalty0 (3):\penalty0 34--43, Oct. 2012.
\newblock ISSN 0163-5808.
\newblock \doi{10.1145/2380776.2380786}.

\bibitem[Kuriyal et~al.(2025)Kuriyal, Vincent, Aubry, and Landrieu]{kuriyal_codex_2025}
A.~Kuriyal, E.~Vincent, M.~Aubry, and L.~Landrieu.
\newblock {{CoDEx}}: {{Combining Domain Expertise}} for {{Spatial Generalization}} in {{Satellite Image Analysis}}.
\newblock In \emph{Proceedings of the {{Computer Vision}} and {{Pattern Recognition Conference}} ({{CVPR}}) {{Workshops}}}, pages 2194--2203, Nashville (Tennessee), United States, Apr. 2025.

\bibitem[Le~Bris and Giordano(2019)]{lebris_archival_2019}
A.~Le~Bris and S.~Giordano.
\newblock Archival aerial photogrammetric surveys: A data source to study land use/cover evolution over the last century -- opportunities and issues.
\newblock In \emph{International {{Land Use Symposium}} ({{ILUS}} 2019)}, Paris, France, Dec. 2019.

\bibitem[Lei et~al.(2021)Lei, Peng, Zhang, Ke, and Li]{lei_hierarchical_2021}
Y.~Lei, D.~Peng, P.~Zhang, Q.~Ke, and H.~Li.
\newblock Hierarchical {{Paired Channel Fusion Network}} for {{Street Scene Change Detection}}.
\newblock \emph{IEEE Trans Image Process}, 30:\penalty0 55--67, 2021.
\newblock ISSN 1941-0042.
\newblock \doi{10.1109/TIP.2020.3031173}.

\bibitem[Lel{\'e}gard et~al.(2020)Lel{\'e}gard, Le~Bris, and Giordano]{lelegard_correction_2020}
L.~Lel{\'e}gard, A.~Le~Bris, and S.~Giordano.
\newblock Correction of {{Systematic Radiometric Inhomogeneity}} in {{Scanned Aerial Campaigns Using Principal Component Analysis}}.
\newblock \emph{ISPRS Annals of the Photogrammetry, Remote Sensing and Spatial Information Sciences}, V-2-2020:\penalty0 871--876, Aug. 2020.
\newblock ISSN 2194-9042.
\newblock \doi{10.5194/isprs-annals-V-2-2020-871-2020}.

\bibitem[Li et~al.(2025)Li, He, Li, Guo, and Zhang]{li_overcoming_2025}
J.~Li, W.~He, Z.~Li, Y.~Guo, and H.~Zhang.
\newblock Overcoming the uncertainty challenges in detecting building changes from remote sensing images.
\newblock \emph{ISPRS Journal of Photogrammetry and Remote Sensing}, 220:\penalty0 1--17, Feb. 2025.
\newblock ISSN 0924-2716.
\newblock \doi{10.1016/j.isprsjprs.2024.11.017}.

\bibitem[Li et~al.(2019{\natexlab{a}})Li, Peng, Chen, Jiao, Zhou, and Shang]{li_deep_2019}
Y.~Li, C.~Peng, Y.~Chen, L.~Jiao, L.~Zhou, and R.~Shang.
\newblock A {{Deep Learning Method}} for {{Change Detection}} in {{Synthetic Aperture Radar Images}}.
\newblock \emph{IEEE Transactions on Geoscience and Remote Sensing}, 57\penalty0 (8):\penalty0 5751--5763, Aug. 2019{\natexlab{a}}.
\newblock ISSN 1558-0644.
\newblock \doi{10.1109/TGRS.2019.2901945}.

\bibitem[Li et~al.(2019{\natexlab{b}})Li, Cao, Wei, Duan, Wu, Hou, and Zhu]{li_timeseries_2019}
Z.~Li, Y.~Cao, J.~Wei, M.~Duan, L.~Wu, J.~Hou, and J.~Zhu.
\newblock Time-series {{InSAR}} ground deformation monitoring: {{Atmospheric}} delay modeling and estimating.
\newblock \emph{Earth-Science Reviews}, 192:\penalty0 258--284, May 2019{\natexlab{b}}.
\newblock ISSN 0012-8252.
\newblock \doi{10.1016/j.earscirev.2019.03.008}.

\bibitem[Liu et~al.(2024)Liu, Chen, Zhang, Qi, Zou, and Shi]{liu_changeagent_2024}
C.~Liu, K.~Chen, H.~Zhang, Z.~Qi, Z.~Zou, and Z.~Shi.
\newblock Change-{{Agent}}: {{Toward Interactive Comprehensive Remote Sensing Change Interpretation}} and {{Analysis}}.
\newblock \emph{IEEE Transactions on Geoscience and Remote Sensing}, 62:\penalty0 1--16, 2024.
\newblock ISSN 1558-0644.
\newblock \doi{10.1109/TGRS.2024.3425815}.

\bibitem[Liu et~al.(2018)Liu, Gong, Qin, and Zhang]{liu_deep_2018a}
J.~Liu, M.~Gong, K.~Qin, and P.~Zhang.
\newblock A {{Deep Convolutional Coupling Network}} for {{Change Detection Based}} on {{Heterogeneous Optical}} and {{Radar Images}}.
\newblock \emph{IEEE Transactions on Neural Networks and Learning Systems}, 29\penalty0 (3):\penalty0 545--559, Mar. 2018.
\newblock ISSN 2162-2388.
\newblock \doi{10.1109/TNNLS.2016.2636227}.

\bibitem[Longley et~al.(2005)Longley, Rhind, Maguire, and Goodchild]{longley_introduction_2005}
P.~A. Longley, D.~W. Rhind, D.~J. Maguire, and M.~F. Goodchild.
\newblock Introduction.
\newblock In \emph{Geographical {{Information Systems}}: {{Principles}}, {{Techniques}}, {{Management}} and {{Applications}}, 2nd {{Edition}}, {{Abridged}}}, {{GIS}}, {{Remote Sensing}} \& {{Cartography}}. Wiley, 2005.

\bibitem[Lowman et~al.(2024)Lowman, Zheng, Fraser, The, and Valipour]{lowman_seefar_2024}
J.~Lowman, K.~L. Zheng, R.~Fraser, J.~V.~G. The, and M.~Valipour.
\newblock {{SeeFar}}: {{Satellite Agnostic Multi-Resolution Dataset}} for {{Geospatial Foundation Models}}, June 2024.

\bibitem[Luangluewut(2024)]{luangluewut_problem_2024}
W.~Luangluewut.
\newblock The problem of using segmentation data for change detection in surveyed areas in {{Thailand}}.
\newblock In \emph{2024 28th {{International Computer Science}} and {{Engineering Conference}} ({{ICSEC}})}, pages 1--5, Nov. 2024.
\newblock \doi{10.1109/ICSEC62781.2024.10770744}.

\bibitem[Lv et~al.(2022)Lv, Huang, Li, Zhao, Benediktsson, Sun, and Falco]{lv_land_2022}
Z.~Lv, H.~Huang, X.~Li, M.~Zhao, J.~A. Benediktsson, W.~Sun, and N.~Falco.
\newblock Land {{Cover Change Detection With Heterogeneous Remote Sensing Images}}: {{Review}}, {{Progress}}, and {{Perspective}}.
\newblock \emph{Proceedings of the IEEE}, 110\penalty0 (12):\penalty0 1976--1991, Dec. 2022.
\newblock ISSN 1558-2256.
\newblock \doi{10.1109/JPROC.2022.3219376}.

\bibitem[Majeed et~al.(2022)Majeed, Khan, and Hwang]{majeed_comprehensive_2022}
A.~Majeed, S.~Khan, and S.~O. Hwang.
\newblock A {{Comprehensive Analysis}} of {{Privacy-Preserving Solutions Developed}} for {{Online Social Networks}}.
\newblock \emph{Electronics}, 11\penalty0 (13):\penalty0 1931, Jan. 2022.
\newblock ISSN 2079-9292.
\newblock \doi{10.3390/electronics11131931}.

\bibitem[Malila(1980)]{malila_change_1980}
W.~Malila.
\newblock Change {{Vector Analysis}}: {{An Approach}} for {{Detecting Forest Changes}} with {{Landsat}}.
\newblock \emph{LARS Symposia}, Jan. 1980.

\bibitem[Mall et~al.(2022)Mall, Hariharan, and Bala]{mall_change_2022}
U.~Mall, B.~Hariharan, and K.~Bala.
\newblock Change {{Event Dataset}} for {{Discovery}} from {{Spatio-temporal Remote Sensing Imagery}}.
\newblock In \emph{Thirty-Sixth {{Conference}} on {{Neural Information Processing Systems Datasets}} and {{Benchmarks Track}}}, Oct. 2022.

\bibitem[Mall et~al.(2023)Mall, Hariharan, and Bala]{mall_changeaware_2023}
U.~Mall, B.~Hariharan, and K.~Bala.
\newblock Change-{{Aware Sampling}} and {{Contrastive Learning}} for {{Satellite Images}}.
\newblock In \emph{{{CVPR}}}, 2023.

\bibitem[Mallet and Le~Bris(2020)]{mallet_current_2020}
C.~Mallet and A.~Le~Bris.
\newblock Current {{Challenges}} in {{Operational Very High Resolution Land-cover Mapping}}.
\newblock \emph{The International Archives of the Photogrammetry, Remote Sensing and Spatial Information Sciences}, XLIII-B2-2020:\penalty0 703--710, Aug. 2020.
\newblock ISSN 1682-1750.
\newblock \doi{10.5194/isprs-archives-XLIII-B2-2020-703-2020}.

\bibitem[Mattson and Haack(1987)]{mattson_role_1987}
W.~J. Mattson and R.~A. Haack.
\newblock The {{Role}} of {{Drought}} in {{Outbreaks}} of {{Plant-Eating Insects}}.
\newblock \emph{BioScience}, 37\penalty0 (2):\penalty0 110--118, 1987.
\newblock ISSN 0006-3568.
\newblock \doi{10.2307/1310365}.

\bibitem[{MAXAR}(2024)]{maxar_persistent_2024}
{MAXAR}.
\newblock Persistent {{Change Monitoring}}.
\newblock https://www.maxar.com/products/persistent-change-monitoring, 2024.

\bibitem[Mendieta et~al.(2023)Mendieta, Han, Shi, Zhu, and Chen]{mendieta_geospatial_2023}
M.~Mendieta, B.~Han, X.~Shi, Y.~Zhu, and C.~Chen.
\newblock Towards {{Geospatial Foundation Models}} via {{Continual Pretraining}}.
\newblock In \emph{2023 {{IEEE}}/{{CVF International Conference}} on {{Computer Vision}} ({{ICCV}})}, pages 16760--16770, Paris, France, Oct. 2023. IEEE.
\newblock \doi{10.1109/ICCV51070.2023.01541}.

\bibitem[Metzger et~al.(2023)Metzger, T{\"u}rkoglu, Daudt, Wegner, and Schindler]{metzger_urban_2023}
N.~Metzger, M.~{\"O}. T{\"u}rkoglu, R.~C. Daudt, J.~D. Wegner, and K.~Schindler.
\newblock Urban {{Change Forecasting}} from {{Satellite Images}}.
\newblock \emph{PFG}, Oct. 2023.
\newblock ISSN 2512-2819.
\newblock \doi{10.1007/s41064-023-00258-8}.

\bibitem[Meyer and Turner(1994)]{meyer_changes_1994}
W.~B. Meyer and B.~L. Turner, editors.
\newblock \emph{Changes in {{Land Use}} and {{Land Cover}}: {{A Global Perspective}}}.
\newblock Cambridge {{University Press}}. Cambridge University Press, university corporation for atmospheric research office for interdisciplinary earth studies edition, 1994.

\bibitem[Mouat et~al.(1993)Mouat, Mahin, and Lancaster]{mouat_remote_1993}
D.~A. Mouat, G.~G. Mahin, and J.~Lancaster.
\newblock Remote sensing techniques in the analysis of change detection.
\newblock \emph{Geocarto International}, 8\penalty0 (2):\penalty0 39--50, June 1993.
\newblock ISSN 1010-6049.
\newblock \doi{10.1080/10106049309354407}.

\bibitem[Naumann et~al.(2024)Naumann, Bonerath, and Haunert]{naumann_manymany_2024}
A.~Naumann, A.~Bonerath, and J.-H. Haunert.
\newblock Many-{{To-Many Polygon Matching}} {\`a} {{La Jaccard}}.
\newblock \emph{LIPIcs, Volume 308, ESA 2024}, 308:\penalty0 90:1--90:15, 2024.
\newblock ISSN 1868-8969.
\newblock \doi{10.4230/LIPICS.ESA.2024.90}.

\bibitem[Nebiker et~al.(2014)Nebiker, Lack, and Deuber]{nebiker_building_2014}
S.~Nebiker, N.~Lack, and M.~Deuber.
\newblock Building {{Change Detection}} from {{Historical Aerial Photographs Using Dense Image Matching}} and {{Object-Based Image Analysis}}.
\newblock \emph{Remote Sensing}, 6\penalty0 (9):\penalty0 8310--8336, Sept. 2014.
\newblock ISSN 2072-4292.
\newblock \doi{10.3390/rs6098310}.

\bibitem[Nemmour and Chibani(2006)]{nemmour_multiple_2006}
H.~Nemmour and Y.~Chibani.
\newblock Multiple support vector machines for land cover change detection: {{An}} application for mapping urban extensions.
\newblock \emph{ISPRS Journal of Photogrammetry and Remote Sensing}, 61\penalty0 (2):\penalty0 125--133, Nov. 2006.
\newblock ISSN 0924-2716.
\newblock \doi{10.1016/j.isprsjprs.2006.09.004}.

\bibitem[Nielsen et~al.(1998)Nielsen, Conradsen, and Simpson]{nielsen_multivariate_1998}
A.~A. Nielsen, K.~Conradsen, and J.~J. Simpson.
\newblock Multivariate {{Alteration Detection}} ({{MAD}}) and {{MAF Postprocessing}} in {{Multispectral}}, {{Bitemporal Image Data}}: {{New Approaches}} to {{Change Detection Studies}}.
\newblock \emph{Remote Sensing of Environment}, 64\penalty0 (1):\penalty0 1--19, 1998.
\newblock ISSN 0034-4257.

\bibitem[Niroshan and Carswell(2024)]{niroshan_ml_2024}
L.~Niroshan and J.~D. Carswell.
\newblock {{ML Updates}} for {{OpenStreetMap}}: {{Analysis}} of {{Research Gaps}} and {{Future Directions}}, June 2024.

\bibitem[Niu et~al.(2019)Niu, Gong, Zhan, and Yang]{niu_conditional_2019}
X.~Niu, M.~Gong, T.~Zhan, and Y.~Yang.
\newblock A {{Conditional Adversarial Network}} for {{Change Detection}} in {{Heterogeneous Images}}.
\newblock \emph{IEEE Geoscience and Remote Sensing Letters}, 16\penalty0 (1):\penalty0 45--49, Jan. 2019.
\newblock ISSN 1558-0571.
\newblock \doi{10.1109/LGRS.2018.2868704}.

\bibitem[{Olteanu-Raimond} et~al.(2017){Olteanu-Raimond}, Hart, Foody, Touya, Kellenberger, and Demetriou]{olteanu-raimond_scale_2017}
A.-M. {Olteanu-Raimond}, G.~Hart, G.~M. Foody, G.~Touya, T.~Kellenberger, and D.~Demetriou.
\newblock The {{Scale}} of {{VGI}} in {{Map Production}}: {{A Perspective}} on {{European National Mapping Agencies}}.
\newblock \emph{Transactions in GIS}, 21\penalty0 (1):\penalty0 74--90, 2017.
\newblock ISSN 1467-9671.
\newblock \doi{10.1111/tgis.12189}.

\bibitem[{Olteanu-Raimond} et~al.(2020){Olteanu-Raimond}, See, Schultz, Foody, Riffler, Gasber, Jolivet, Le~Bris, Meneroux, Liu, Poup{\'e}e, and Gombert]{olteanu-raimond_use_2020}
A.-M. {Olteanu-Raimond}, L.~See, M.~Schultz, G.~Foody, M.~Riffler, T.~Gasber, L.~Jolivet, A.~Le~Bris, Y.~Meneroux, L.~Liu, M.~Poup{\'e}e, and M.~Gombert.
\newblock Use of {{Automated Change Detection}} and {{VGI Sources}} for {{Identifying}} and {{Validating Urban Land Use Change}}.
\newblock \emph{Remote Sensing}, 12\penalty0 (7):\penalty0 1186, Apr. 2020.
\newblock \doi{10.3390/rs12071186}.

\bibitem[OpenAI et~al.(2024)OpenAI, Achiam, Adler, Agarwal, Ahmad, Akkaya, Aleman, Almeida, Altenschmidt, Altman, Anadkat, Avila, Babuschkin, Balaji, Balcom, Baltescu, Bao, Bavarian, Belgum, Bello, Berdine, {Bernadett-Shapiro}, Berner, Bogdonoff, Boiko, Boyd, Brakman, Brockman, Brooks, Brundage, Button, Cai, Campbell, Cann, Carey, Carlson, Carmichael, Chan, Chang, Chantzis, Chen, Chen, Chen, Chen, Chen, Chess, Cho, Chu, Chung, Cummings, Currier, Dai, Decareaux, Degry, Deutsch, Deville, Dhar, Dohan, Dowling, Dunning, Ecoffet, Eleti, Eloundou, Farhi, Fedus, Felix, Fishman, Forte, Fulford, Gao, Georges, Gibson, Goel, Gogineni, Goh, {Gontijo-Lopes}, Gordon, Grafstein, Gray, Greene, Gross, Gu, Guo, Hallacy, Han, Harris, He, Heaton, Heidecke, Hesse, Hickey, Hickey, Hoeschele, Houghton, Hsu, Hu, Hu, Huizinga, Jain, Jain, Jang, Jiang, Jiang, Jin, Jin, Jomoto, Jonn, Jun, Kaftan, Kaiser, Kamali, Kanitscheider, Keskar, Khan, Kilpatrick, Kim, Kim, Kim, Kirchner, Kiros, Knight, Kokotajlo, Kondraciuk, Kondrich,
  Konstantinidis, Kosic, Krueger, Kuo, Lampe, Lan, Lee, Leike, Leung, Levy, Li, Lim, Lin, Lin, Litwin, Lopez, Lowe, Lue, Makanju, Malfacini, Manning, Markov, Markovski, Martin, Mayer, Mayne, McGrew, McKinney, McLeavey, McMillan, McNeil, Medina, Mehta, Menick, Metz, Mishchenko, Mishkin, Monaco, Morikawa, Mossing, Mu, Murati, Murk, M{\'e}ly, Nair, Nakano, Nayak, Neelakantan, Ngo, Noh, Ouyang, O'Keefe, Pachocki, Paino, Palermo, Pantuliano, Parascandolo, Parish, Parparita, Passos, Pavlov, Peng, Perelman, Peres, Petrov, Pinto, Michael, Pokorny, Pokrass, Pong, Powell, Power, Power, Proehl, Puri, Radford, Rae, Ramesh, Raymond, Real, Rimbach, Ross, Rotsted, Roussez, Ryder, Saltarelli, Sanders, Santurkar, Sastry, Schmidt, Schnurr, Schulman, Selsam, Sheppard, Sherbakov, Shieh, Shoker, Shyam, Sidor, Sigler, Simens, Sitkin, Slama, Sohl, Sokolowsky, Song, Staudacher, Such, Summers, Sutskever, Tang, Tezak, Thompson, Tillet, Tootoonchian, Tseng, Tuggle, Turley, Tworek, Uribe, Vallone, Vijayvergiya, Voss, Wainwright, Wang,
  Wang, Wang, Ward, Wei, Weinmann, Welihinda, Welinder, Weng, Weng, Wiethoff, Willner, Winter, Wolrich, Wong, Workman, Wu, Wu, Wu, Xiao, Xu, Yoo, Yu, Yuan, Zaremba, Zellers, Zhang, Zhang, Zhao, Zheng, Zhuang, Zhuk, and Zoph]{openai_gpt4_2024}
OpenAI, J.~Achiam, S.~Adler, S.~Agarwal, L.~Ahmad, I.~Akkaya, F.~L. Aleman, D.~Almeida, J.~Altenschmidt, S.~Altman, S.~Anadkat, R.~Avila, I.~Babuschkin, S.~Balaji, V.~Balcom, P.~Baltescu, H.~Bao, M.~Bavarian, J.~Belgum, I.~Bello, J.~Berdine, G.~{Bernadett-Shapiro}, C.~Berner, L.~Bogdonoff, O.~Boiko, M.~Boyd, A.-L. Brakman, G.~Brockman, T.~Brooks, M.~Brundage, K.~Button, T.~Cai, R.~Campbell, A.~Cann, B.~Carey, C.~Carlson, R.~Carmichael, B.~Chan, C.~Chang, F.~Chantzis, D.~Chen, S.~Chen, R.~Chen, J.~Chen, M.~Chen, B.~Chess, C.~Cho, C.~Chu, H.~W. Chung, D.~Cummings, J.~Currier, Y.~Dai, C.~Decareaux, T.~Degry, N.~Deutsch, D.~Deville, A.~Dhar, D.~Dohan, S.~Dowling, S.~Dunning, A.~Ecoffet, A.~Eleti, T.~Eloundou, D.~Farhi, L.~Fedus, N.~Felix, S.~P. Fishman, J.~Forte, I.~Fulford, L.~Gao, E.~Georges, C.~Gibson, V.~Goel, T.~Gogineni, G.~Goh, R.~{Gontijo-Lopes}, J.~Gordon, M.~Grafstein, S.~Gray, R.~Greene, J.~Gross, S.~S. Gu, Y.~Guo, C.~Hallacy, J.~Han, J.~Harris, Y.~He, M.~Heaton, J.~Heidecke, C.~Hesse, A.~Hickey,
  W.~Hickey, P.~Hoeschele, B.~Houghton, K.~Hsu, S.~Hu, X.~Hu, J.~Huizinga, S.~Jain, S.~Jain, J.~Jang, A.~Jiang, R.~Jiang, H.~Jin, D.~Jin, S.~Jomoto, B.~Jonn, H.~Jun, T.~Kaftan, {\L}.~Kaiser, A.~Kamali, I.~Kanitscheider, N.~S. Keskar, T.~Khan, L.~Kilpatrick, J.~W. Kim, C.~Kim, Y.~Kim, J.~H. Kirchner, J.~Kiros, M.~Knight, D.~Kokotajlo, {\L}.~Kondraciuk, A.~Kondrich, A.~Konstantinidis, K.~Kosic, G.~Krueger, V.~Kuo, M.~Lampe, I.~Lan, T.~Lee, J.~Leike, J.~Leung, D.~Levy, C.~M. Li, R.~Lim, M.~Lin, S.~Lin, M.~Litwin, T.~Lopez, R.~Lowe, P.~Lue, A.~Makanju, K.~Malfacini, S.~Manning, T.~Markov, Y.~Markovski, B.~Martin, K.~Mayer, A.~Mayne, B.~McGrew, S.~M. McKinney, C.~McLeavey, P.~McMillan, J.~McNeil, D.~Medina, A.~Mehta, J.~Menick, L.~Metz, A.~Mishchenko, P.~Mishkin, V.~Monaco, E.~Morikawa, D.~Mossing, T.~Mu, M.~Murati, O.~Murk, D.~M{\'e}ly, A.~Nair, R.~Nakano, R.~Nayak, A.~Neelakantan, R.~Ngo, H.~Noh, L.~Ouyang, C.~O'Keefe, J.~Pachocki, A.~Paino, J.~Palermo, A.~Pantuliano, G.~Parascandolo, J.~Parish, E.~Parparita,
  A.~Passos, M.~Pavlov, A.~Peng, A.~Perelman, F.~d. A.~B. Peres, M.~Petrov, H.~P. d.~O. Pinto, Michael, Pokorny, M.~Pokrass, V.~H. Pong, T.~Powell, A.~Power, B.~Power, E.~Proehl, R.~Puri, A.~Radford, J.~Rae, A.~Ramesh, C.~Raymond, F.~Real, K.~Rimbach, C.~Ross, B.~Rotsted, H.~Roussez, N.~Ryder, M.~Saltarelli, T.~Sanders, S.~Santurkar, G.~Sastry, H.~Schmidt, D.~Schnurr, J.~Schulman, D.~Selsam, K.~Sheppard, T.~Sherbakov, J.~Shieh, S.~Shoker, P.~Shyam, S.~Sidor, E.~Sigler, M.~Simens, J.~Sitkin, K.~Slama, I.~Sohl, B.~Sokolowsky, Y.~Song, N.~Staudacher, F.~P. Such, N.~Summers, I.~Sutskever, J.~Tang, N.~Tezak, M.~B. Thompson, P.~Tillet, A.~Tootoonchian, E.~Tseng, P.~Tuggle, N.~Turley, J.~Tworek, J.~F.~C. Uribe, A.~Vallone, A.~Vijayvergiya, C.~Voss, C.~Wainwright, J.~J. Wang, A.~Wang, B.~Wang, J.~Ward, J.~Wei, C.~J. Weinmann, A.~Welihinda, P.~Welinder, J.~Weng, L.~Weng, M.~Wiethoff, D.~Willner, C.~Winter, S.~Wolrich, H.~Wong, L.~Workman, S.~Wu, J.~Wu, M.~Wu, K.~Xiao, T.~Xu, S.~Yoo, K.~Yu, Q.~Yuan, W.~Zaremba,
  R.~Zellers, C.~Zhang, M.~Zhang, S.~Zhao, T.~Zheng, J.~Zhuang, W.~Zhuk, and B.~Zoph.
\newblock {{GPT-4 Technical Report}}, Mar. 2024.

\bibitem[{Osi{\'n}ska-Skotak} et~al.(2019){Osi{\'n}ska-Skotak}, Radecka, Pi{\'o}rkowski, {Michalska-Hejduk}, Kope{\'c}, {Tokarska-Guzik}, Ostrowski, Kania, and Niedzielko]{osinska-skotak_mapping_2019}
K.~{Osi{\'n}ska-Skotak}, A.~Radecka, H.~Pi{\'o}rkowski, D.~{Michalska-Hejduk}, D.~Kope{\'c}, B.~{Tokarska-Guzik}, W.~Ostrowski, A.~Kania, and J.~Niedzielko.
\newblock Mapping {{Succession}} in {{Non-Forest Habitats}} by {{Means}} of {{Remote Sensing}}: {{Is}} the {{Data Acquisition Time Critical}} for {{Species Discrimination}}?
\newblock \emph{Remote Sensing}, 11\penalty0 (22):\penalty0 2629, Jan. 2019.
\newblock ISSN 2072-4292.
\newblock \doi{10.3390/rs11222629}.

\bibitem[Pan et~al.(2019)Pan, Zhou, Liu, and Wang]{pan_global_2019}
L.~Pan, H.~Zhou, Y.~Liu, and M.~Wang.
\newblock Global event influence model: Integrating crowd motion and social psychology for global anomaly detection in dense crowds.
\newblock \emph{JEI}, 28\penalty0 (2):\penalty0 023033, Apr. 2019.
\newblock ISSN 1017-9909, 1560-229X.
\newblock \doi{10.1117/1.JEI.28.2.023033}.

\bibitem[Pang et~al.(2023)Pang, Wu, Ding, Song, and Xia]{pang_detecting_2023}
C.~Pang, J.~Wu, J.~Ding, C.~Song, and G.-S. Xia.
\newblock Detecting {{Building Changes}} with {{Off-Nadir Aerial Images}}, Jan. 2023.

\bibitem[Paranjape et~al.(2025)Paranjape, {de Melo}, and Patel]{paranjape_mambabased_2024}
J.~N. Paranjape, C.~{de Melo}, and V.~M. Patel.
\newblock A {{Mamba-based Siamese Network}} for {{Remote Sensing Change Detection}}.
\newblock In \emph{Proceedings of the {{IEEE}}/{{CVF Winter Conference}} on {{Applications}} of {{Computer Vision}}}, Tuscon, 2025.

\bibitem[Parker et~al.(2003)Parker, Manson, Janssen, Hoffmann, and Deadman]{parker_multiagent_2003}
D.~C. Parker, S.~M. Manson, M.~A. Janssen, M.~J. Hoffmann, and P.~Deadman.
\newblock Multi-{{Agent Systems}} for the {{Simulation}} of {{Land-Use}} and {{Land-Cover Change}}: {{A Review}}.
\newblock \emph{Annals of the Association of American Geographers}, 93\penalty0 (2):\penalty0 314--337, June 2003.
\newblock ISSN 0004-5608.
\newblock \doi{10.1111/1467-8306.9302004}.

\bibitem[Pasquarella et~al.(2022)Pasquarella, Ar{\'e}valo, Bratley, Bullock, Gorelick, Yang, and Kennedy]{pasquarella_demystifying_2022}
V.~J. Pasquarella, P.~Ar{\'e}valo, K.~H. Bratley, E.~L. Bullock, N.~Gorelick, Z.~Yang, and R.~E. Kennedy.
\newblock Demystifying {{LandTrendr}} and {{CCDC}} temporal segmentation.
\newblock \emph{International Journal of Applied Earth Observation and Geoinformation}, 110:\penalty0 102806, June 2022.
\newblock ISSN 15698432.
\newblock \doi{10.1016/j.jag.2022.102806}.

\bibitem[Peng et~al.(2025)Peng, Liu, Zhang, Guan, Li, and Bruzzone]{peng_deep_2025}
D.~Peng, X.~Liu, Y.~Zhang, H.~Guan, Y.~Li, and L.~Bruzzone.
\newblock Deep learning change detection techniques for optical remote sensing imagery: {{Status}}, perspectives and challenges.
\newblock \emph{International Journal of Applied Earth Observation and Geoinformation}, 136:\penalty0 104282, Feb. 2025.
\newblock ISSN 15698432.
\newblock \doi{10.1016/j.jag.2024.104282}.

\bibitem[Petit et~al.(2001)Petit, Scudder, and Lambin]{petit_quantifying_2001}
C.~Petit, T.~Scudder, and E.~Lambin.
\newblock Quantifying processes of land-cover change by remote sensing: {{Resettlement}} and rapid land-cover changes in south-eastern {{Zambia}}.
\newblock \emph{International Journal of Remote Sensing}, 22\penalty0 (17):\penalty0 3435--3456, Jan. 2001.
\newblock ISSN 0143-1161.
\newblock \doi{10.1080/01431160010006881}.

\bibitem[Peuquet(2005)]{peuquet_time_2005}
D.~J. Peuquet.
\newblock Time in {{GIS}} and geographical databases.
\newblock In \emph{Geographical {{Information Systems}}: {{Principles}}, {{Techniques}}, {{Management}} and {{Applications}}, 2nd {{Edition}}, {{Abridged}}}, {{GIS}}, {{Remote Sensing}} \& {{Cartography}}, pages 91--103. Wiley, 2005.

\bibitem[Pickles(1995)]{pickles_ground_1995}
J.~Pickles.
\newblock \emph{Ground {{Truth}}: {{The Social Implications}} of {{Geographic Information Systems}}}.
\newblock Guilford Press, Jan. 1995.
\newblock ISBN 978-0-89862-295-9.

\bibitem[Pohl and Van~Genderen(1998)]{pohl_multisensor_1998}
C.~Pohl and J.~L. Van~Genderen.
\newblock Multisensor image fusion in remote sensing: {{Concepts}}, methods and applications.
\newblock \emph{International Journal of Remote Sensing}, 19\penalty0 (5):\penalty0 823--854, Jan. 1998.
\newblock ISSN 0143-1161.
\newblock \doi{10.1080/014311698215748}.

\bibitem[Preiss and Stacy(2006)]{preiss_coherent_2006}
M.~Preiss and N.~J. Stacy.
\newblock Coherent {{Change Detection}}: {{Theoretical Description}} and {{Experimental Results}}.
\newblock Technical report, {Defence Science and Technology Organisation Australia}, 2006.

\bibitem[Price(1992)]{price_shrub_1992}
K.~P. Price.
\newblock Shrub {{Dieback}} in a {{Semiarid Ecosystem}}: {{The Integration}} of {{Remote Sensing}} and {{Geographic Information Systems}} for {{Detecting Vegetation Change}}.
\newblock \emph{Photogrammetric Engineering \& Remote Sensing}, 58\penalty0 (4):\penalty0 455--463, 1992.

\bibitem[{P{\u a}tru-Stupariu} et~al.(2013){P{\u a}tru-Stupariu}, Angelstam, Elbakidze, Huzui, and Andersson]{patru-stupariu_using_2013}
I.~{P{\u a}tru-Stupariu}, P.~Angelstam, M.~Elbakidze, A.~Huzui, and K.~Andersson.
\newblock Using forest history and spatial patterns to identify potential high conservation value forests in {{Romania}}.
\newblock \emph{Biodivers Conserv}, 22\penalty0 (9):\penalty0 2023--2039, Aug. 2013.
\newblock ISSN 1572-9710.
\newblock \doi{10.1007/s10531-013-0523-3}.

\bibitem[Qin(2021)]{qin_change_2021}
R.~Qin.
\newblock Chapter 7. {{Change Detection}} for {{Geodatabase Updating}}.
\newblock In \emph{{{3D}}/{{4D City Modelling}}: From Sensors to Applications}. arXiv, June 2021.

\bibitem[Qin et~al.(2016)Qin, Tian, and Reinartz]{qin_3d_2016}
R.~Qin, J.~Tian, and P.~Reinartz.
\newblock {{3D}} change detection -- {{Approaches}} and applications.
\newblock \emph{ISPRS Journal of Photogrammetry and Remote Sensing}, 122:\penalty0 41--56, Dec. 2016.
\newblock ISSN 0924-2716.
\newblock \doi{10.1016/j.isprsjprs.2016.09.013}.

\bibitem[Quarello et~al.(2022)Quarello, Bock, and Lebarbier]{quarello_gnssseg_2022}
A.~Quarello, O.~Bock, and E.~Lebarbier.
\newblock {{GNSSseg}}, a {{Statistical Method}} for the {{Segmentation}} of {{Daily GNSS IWV Time Series}}.
\newblock \emph{Remote Sensing}, 14\penalty0 (14):\penalty0 3379, Jan. 2022.
\newblock ISSN 2072-4292.
\newblock \doi{10.3390/rs14143379}.

\bibitem[Rainu~Nandal(2013)]{rainunandal_spatiotemporal_2013}
R.~N. Rainu~Nandal.
\newblock Spatio-{{Temporal Database}} and {{Its Models}}: {{A Review}}.
\newblock \emph{IOSR-JCE}, 11\penalty0 (2):\penalty0 91--100, 2013.
\newblock ISSN 22788727, 22780661.
\newblock \doi{10.9790/0661-11291100}.

\bibitem[Raspini et~al.(2018)Raspini, Bianchini, Ciampalini, Del~Soldato, Solari, Novali, Del~Conte, Rucci, Ferretti, and Casagli]{raspini_continuous_2018}
F.~Raspini, S.~Bianchini, A.~Ciampalini, M.~Del~Soldato, L.~Solari, F.~Novali, S.~Del~Conte, A.~Rucci, A.~Ferretti, and N.~Casagli.
\newblock Continuous, semi-automatic monitoring of ground deformation using {{Sentinel-1}} satellites.
\newblock \emph{Sci Rep}, 8\penalty0 (1):\penalty0 7253, May 2018.
\newblock ISSN 2045-2322.
\newblock \doi{10.1038/s41598-018-25369-w}.

\bibitem[R{\"a}th et~al.(2023)R{\"a}th, {Gr{\^e}t-Regamey}, Jiao, Wu, and {van Strien}]{rath_settlement_2023}
Y.~M. R{\"a}th, A.~{Gr{\^e}t-Regamey}, C.~Jiao, S.~Wu, and M.~J. {van Strien}.
\newblock Settlement relationships and their morphological homogeneity across time and scale.
\newblock \emph{Sci Rep}, 13\penalty0 (1):\penalty0 11248, July 2023.
\newblock ISSN 2045-2322.
\newblock \doi{10.1038/s41598-023-38338-9}.

\bibitem[Robin et~al.(2010)Robin, Moisan, and {Le Hegarat-Mascle}]{robin_acontrario_2010}
A.~Robin, L.~Moisan, and S.~{Le Hegarat-Mascle}.
\newblock An {{A-Contrario Approach}} for {{Subpixel Change Detection}} in {{Satellite Imagery}}.
\newblock \emph{IEEE Transactions on Pattern Analysis and Machine Intelligence}, 32\penalty0 (11):\penalty0 1977--1993, Nov. 2010.
\newblock ISSN 1939-3539.
\newblock \doi{10.1109/TPAMI.2010.37}.

\bibitem[Roscher et~al.(2024)Roscher, Ru{\ss}wurm, Gevaert, Kampffmeyer, dos Santos, Vakalopoulou, H{\"a}nsch, Hansen, Nogueira, Prexl, and Tuia]{roscher_better_2024}
R.~Roscher, M.~Ru{\ss}wurm, C.~Gevaert, M.~Kampffmeyer, J.~A. dos Santos, M.~Vakalopoulou, R.~H{\"a}nsch, S.~Hansen, K.~Nogueira, J.~Prexl, and D.~Tuia.
\newblock Better, {{Not Just More}}: {{Data-Centric Machine Learning}} for {{Earth Observation}}, June 2024.

\bibitem[Rupnik et~al.(2017)Rupnik, Daakir, and Pierrot~Deseilligny]{rupnik_micmac_2017}
E.~Rupnik, M.~Daakir, and M.~Pierrot~Deseilligny.
\newblock {{MicMac}} -- a free, open-source solution for photogrammetry.
\newblock \emph{Open Geospatial Data, Software and Standards}, 2\penalty0 (1):\penalty0 14, June 2017.
\newblock ISSN 2363-7501.
\newblock \doi{10.1186/s40965-017-0027-2}.

\bibitem[R{\r u}{\v z}i{\v c}ka et~al.(2020)R{\r u}{\v z}i{\v c}ka, D'Aronco, Wegner, and Schindler]{ruzicka_deep_2020}
V.~R{\r u}{\v z}i{\v c}ka, S.~D'Aronco, J.~D. Wegner, and K.~Schindler.
\newblock Deep {{Active Learning}} in {{Remote Sensing}} for data efficient {{Change Detection}}, Aug. 2020.

\bibitem[Sachdeva and Zisserman(2023)]{sachdeva_change_2023}
R.~Sachdeva and A.~Zisserman.
\newblock The {{Change You Want}} to {{See}} ({{Now}} in {{3D}}).
\newblock In \emph{Proceedings of the {{IEEE}}/{{CVF International Conference}} on {{Computer Vision}}}, pages 2060--2069, Aug. 2023.

\bibitem[Saha(2024)]{saha_confidence_2024}
S.~Saha.
\newblock Confidence {{Estimation}} in {{Unsupervised Deep Change Vector Analysis}}.
\newblock \emph{IEEE Transactions on Geoscience and Remote Sensing}, 62:\penalty0 1--9, 2024.
\newblock ISSN 1558-0644.
\newblock \doi{10.1109/TGRS.2024.3504742}.

\bibitem[Saha et~al.(2019)Saha, Bovolo, and Bruzzone]{saha_unsupervised_2019}
S.~Saha, F.~Bovolo, and L.~Bruzzone.
\newblock Unsupervised {{Deep Change Vector Analysis}} for {{Multiple-Change Detection}} in {{VHR Images}}.
\newblock \emph{IEEE Transactions on Geoscience and Remote Sensing}, 57\penalty0 (6):\penalty0 3677--3693, June 2019.
\newblock ISSN 1558-0644.
\newblock \doi{10.1109/TGRS.2018.2886643}.

\bibitem[Saidi et~al.(2024)Saidi, Idbraim, Karmoude, Masse, and Arbelo]{saidi_deeplearning_2024}
S.~Saidi, S.~Idbraim, Y.~Karmoude, A.~Masse, and M.~Arbelo.
\newblock Deep-{{Learning}} for {{Change Detection Using Multi-Modal Fusion}} of {{Remote Sensing Images}}: {{A Review}}.
\newblock \emph{Remote Sensing}, 16\penalty0 (20):\penalty0 3852, Jan. 2024.
\newblock ISSN 2072-4292.
\newblock \doi{10.3390/rs16203852}.

\bibitem[{Saldana-Perez} et~al.(2019){Saldana-Perez}, Cavali{\`e}re, {Torres-Ruiz}, and Moreno]{saldana-perez_when_2019}
M.~{Saldana-Perez}, C.~Cavali{\`e}re, M.~{Torres-Ruiz}, and M.~Moreno.
\newblock When {{Twitter Becomes}} a {{Data Source}} for {{Geospatial Analysis}}.
\newblock \emph{Research in Computing Science}, 148:\penalty0 357--374, Dec. 2019.
\newblock \doi{10.13053/rcs-148-10-30}.

\bibitem[Sasagawa et~al.(2013)Sasagawa, Baltsavias, Kocaman~Aksakal, and Wegner]{sasagawa_investigation_2013}
A.~Sasagawa, E.~Baltsavias, S.~Kocaman~Aksakal, and J.~D. Wegner.
\newblock Investigation on automatic change detection using pixel-changes and {{DSM-changes}} with {{ALOS-PRISM}} triplet images.
\newblock \emph{The International Archives of the Photogrammetry, Remote Sensing and Spatial Information Sciences}, XL-7-W2:\penalty0 213--217, Oct. 2013.
\newblock ISSN 1682-1750.
\newblock \doi{10.5194/isprsarchives-XL-7-W2-213-2013}.

\bibitem[Settles(2009)]{settles_active_2009}
B.~Settles.
\newblock Active {{Learning Literature Survey}}.
\newblock Technical {{Report}}, University of Wisconsin-Madison Department of Computer Sciences, 2009.

\bibitem[Shafiee and Zechman(2013)]{shafiee_agentbased_2013}
M.~E. Shafiee and E.~M. Zechman.
\newblock An agent-based modeling framework for sociotechnical simulation of water distribution contamination events.
\newblock \emph{Journal of Hydroinformatics}, 15\penalty0 (3):\penalty0 862--880, Feb. 2013.
\newblock ISSN 1464-7141.
\newblock \doi{10.2166/hydro.2013.158}.

\bibitem[Shafique et~al.(2022)Shafique, Cao, Khan, Asad, and Aslam]{shafique_deep_2022}
A.~Shafique, G.~Cao, Z.~Khan, M.~Asad, and M.~Aslam.
\newblock Deep {{Learning-Based Change Detection}} in {{Remote Sensing Images}}: {{A Review}}.
\newblock \emph{Remote Sensing}, 14\penalty0 (4):\penalty0 871, Jan. 2022.
\newblock ISSN 2072-4292.
\newblock \doi{10.3390/rs14040871}.

\bibitem[Shen et~al.(2021)Shen, Lu, Chen, Wei, Xie, Yue, Chen, Lv, and Jiang]{shen_s2looking_2021}
L.~Shen, Y.~Lu, H.~Chen, H.~Wei, D.~Xie, J.~Yue, R.~Chen, S.~Lv, and B.~Jiang.
\newblock {{S2Looking}}: {{A Satellite Side-Looking Dataset}} for {{Building Change Detection}}.
\newblock \emph{Remote Sensing}, 13\penalty0 (24):\penalty0 5094, Dec. 2021.
\newblock ISSN 2072-4292.
\newblock \doi{10.3390/rs13245094}.

\bibitem[Shi et~al.(2020)Shi, Zhang, Zhang, Chen, and Zhan]{shi_change_2020}
W.~Shi, M.~Zhang, R.~Zhang, S.~Chen, and Z.~Zhan.
\newblock Change {{Detection Based}} on {{Artificial Intelligence}}: {{State-of-the-Art}} and {{Challenges}}.
\newblock \emph{Remote Sensing}, 12\penalty0 (10):\penalty0 1688, Jan. 2020.
\newblock ISSN 2072-4292.
\newblock \doi{10.3390/rs12101688}.

\bibitem[Singh(1989)]{singh_review_1989}
A.~Singh.
\newblock Review {{Article Digital}} change detection techniques using remotely-sensed data.
\newblock \emph{International Journal of Remote Sensing}, 10\penalty0 (6):\penalty0 989--1003, June 1989.
\newblock ISSN 0143-1161.
\newblock \doi{10.1080/01431168908903939}.

\bibitem[Song et~al.(2023)Song, Chen, and Yokoya]{song_syntheworld_2023}
J.~Song, H.~Chen, and N.~Yokoya.
\newblock {{SyntheWorld}}: {{A Large-Scale Synthetic Dataset}} for {{Land Cover Mapping}} and {{Building Change Detection}}, Sept. 2023.

\bibitem[{Stevens-Rumann} et~al.(2022){Stevens-Rumann}, Prichard, Whitman, Parisien, and Meddens]{stevens-rumann_considering_2022}
C.~S. {Stevens-Rumann}, S.~J. Prichard, E.~Whitman, M.-A. Parisien, and A.~J. Meddens.
\newblock Considering regeneration failure in the context of changing climate and disturbance regimes in western {{North America}}.
\newblock \emph{Can. J. For. Res.}, 52\penalty0 (10):\penalty0 1281--1302, Oct. 2022.
\newblock ISSN 0045-5067.
\newblock \doi{10.1139/cjfr-2022-0054}.

\bibitem[Stilla and Xu(2023)]{stilla_change_2023}
U.~Stilla and Y.~Xu.
\newblock Change detection of urban objects using {{3D}} point clouds: {{A}} review.
\newblock \emph{ISPRS Journal of Photogrammetry and Remote Sensing}, 197:\penalty0 228--255, Mar. 2023.
\newblock ISSN 0924-2716.
\newblock \doi{10.1016/j.isprsjprs.2023.01.010}.

\bibitem[Sui et~al.(2012)Sui, Elwood, and Goodchild]{sui_crowdsourcing_2012}
D.~Sui, S.~Elwood, and M.~Goodchild.
\newblock \emph{Crowdsourcing {{Geographic Knowledge}}: {{Volunteered Geographic Information}} ({{VGI}}) in {{Theory}} and {{Practice}}}.
\newblock Springer Science \& Business Media, Aug. 2012.
\newblock ISBN 978-94-007-4587-2.

\bibitem[Sumbul et~al.(2021)Sumbul, {de Wall}, Kreuziger, Marcelino, Costa, Benevides, Caetano, Demir, and Markl]{sumbul_bigearthnetmm_2021}
G.~Sumbul, A.~{de Wall}, T.~Kreuziger, F.~Marcelino, H.~Costa, P.~Benevides, M.~Caetano, B.~Demir, and V.~Markl.
\newblock {{BigEarthNet-MM}}: {{A Large Scale Multi-Modal Multi-Label Benchmark Archive}} for {{Remote Sensing Image Classification}} and {{Retrieval}}.
\newblock \emph{IEEE Geosci. Remote Sens. Mag.}, 9\penalty0 (3):\penalty0 174--180, Sept. 2021.
\newblock ISSN 2168-6831, 2473-2397, 2373-7468.
\newblock \doi{10.1109/MGRS.2021.3089174}.

\bibitem[Sun et~al.(2021)Sun, Lei, Li, Sun, and Kuang]{sun_nonlocal_2021}
Y.~Sun, L.~Lei, X.~Li, H.~Sun, and G.~Kuang.
\newblock Nonlocal patch similarity based heterogeneous remote sensing change detection.
\newblock \emph{Pattern Recognition}, 109:\penalty0 107598, Jan. 2021.
\newblock ISSN 0031-3203.
\newblock \doi{10.1016/j.patcog.2020.107598}.

\bibitem[Tailanian et~al.(2023)Tailanian, Mus{\'e}, and Pardo]{tailanian_contrario_2023}
M.~Tailanian, P.~Mus{\'e}, and {\'A}.~Pardo.
\newblock A {{Contrario Multi-scale Anomaly Detection Method}} for~{{Industrial Quality Inspection}}.
\newblock In M.~A. Wani and V.~Palade, editors, \emph{Deep {{Learning Applications}}, {{Volume}} 4}, Advances in {{Intelligent Systems}} and {{Computing}}, pages 193--216. Springer Nature, Singapore, 2023.
\newblock ISBN 978-981-19-6153-3.
\newblock \doi{10.1007/978-981-19-6153-3_8}.

\bibitem[Taneja et~al.(2011)Taneja, Ballan, and Pollefeys]{taneja_image_2011}
A.~Taneja, L.~Ballan, and M.~Pollefeys.
\newblock Image based detection of geometric changes in urban environments.
\newblock In \emph{2011 {{International Conference}} on {{Computer Vision}}}, pages 2336--2343, Barcelona, Spain, Nov. 2011. IEEE.
\newblock \doi{10.1109/ICCV.2011.6126515}.

\bibitem[Tian et~al.(2020)Tian, Ma, Zheng, and Zhong]{tian_hiucd_2020}
S.~Tian, A.~Ma, Z.~Zheng, and Y.~Zhong.
\newblock Hi-{{UCD}}: {{A Large-scale Dataset}} for {{Urban Semantic Change Detection}} in {{Remote Sensing Imagery}}, Dec. 2020.

\bibitem[Tian et~al.(2022)Tian, Zhong, Zheng, Ma, Tan, and Zhang]{tian_largescale_2022}
S.~Tian, Y.~Zhong, Z.~Zheng, A.~Ma, X.~Tan, and L.~Zhang.
\newblock Large-scale deep learning based binary and semantic change detection in ultra high resolution remote sensing imagery: {{From}} benchmark datasets to urban application.
\newblock \emph{ISPRS Journal of Photogrammetry and Remote Sensing}, 193:\penalty0 164--186, Nov. 2022.
\newblock ISSN 0924-2716.
\newblock \doi{10.1016/j.isprsjprs.2022.08.012}.

\bibitem[Tian et~al.(2023)Tian, Tan, Ma, Zheng, Zhang, and Zhong]{tian_temporalagnostic_2023}
S.~Tian, X.~Tan, A.~Ma, Z.~Zheng, L.~Zhang, and Y.~Zhong.
\newblock Temporal-agnostic change region proposal for semantic change detection.
\newblock \emph{ISPRS Journal of Photogrammetry and Remote Sensing}, 204:\penalty0 306--320, Oct. 2023.
\newblock ISSN 0924-2716.
\newblock \doi{10.1016/j.isprsjprs.2023.06.017}.

\bibitem[Toker et~al.(2022)Toker, Kondmann, Weber, Eisenberger, Camero, Hu, Hoderlein, {\c S}enaras, Davis, Cremers, Marchisio, Zhu, and {Leal-Taix{\'e}}]{toker_dynamicearthnet_2022}
A.~Toker, L.~Kondmann, M.~Weber, M.~Eisenberger, A.~Camero, J.~Hu, A.~P. Hoderlein, {\c C}.~{\c S}enaras, T.~Davis, D.~Cremers, G.~Marchisio, X.~X. Zhu, and L.~{Leal-Taix{\'e}}.
\newblock {{DynamicEarthNet}}: {{Daily Multi-Spectral Satellite Dataset}} for {{Semantic Change Segmentation}}.
\newblock \emph{Proceedings of the IEEE/CVF Conference on Computer Vision and Pattern Recognition}, pages 21158--21167, Mar. 2022.

\bibitem[Touvron et~al.(2023)Touvron, Lavril, Izacard, Martinet, Lachaux, Lacroix, Rozi{\`e}re, Goyal, Hambro, Azhar, Rodriguez, Joulin, Grave, and Lample]{touvron_llama_2023}
H.~Touvron, T.~Lavril, G.~Izacard, X.~Martinet, M.-A. Lachaux, T.~Lacroix, B.~Rozi{\`e}re, N.~Goyal, E.~Hambro, F.~Azhar, A.~Rodriguez, A.~Joulin, E.~Grave, and G.~Lample.
\newblock {{LLaMA}}: {{Open}} and {{Efficient Foundation Language Models}}.
\newblock https://arxiv.org/abs/2302.13971v1, Feb. 2023.

\bibitem[Truong et~al.(2020)Truong, Oudre, and Vayatis]{truong_selective_2020}
C.~Truong, L.~Oudre, and N.~Vayatis.
\newblock Selective review of offline change point detection methods.
\newblock \emph{Signal Processing}, 167:\penalty0 107299, Feb. 2020.
\newblock ISSN 01651684.
\newblock \doi{10.1016/j.sigpro.2019.107299}.

\bibitem[Truong et~al.(2022)Truong, Touya, and {de Runz}]{truong_role_2022}
Q.~T. Truong, G.~Touya, and C.~{de Runz}.
\newblock The role of citizens and geoinformation in providing alerts and crisis information to the public.
\newblock \emph{Security and Defence Quarterly}, 40\penalty0 (4):\penalty0 58--74, 2022.
\newblock ISSN 2300-8741.

\bibitem[Tsujimoto et~al.(2024)Tsujimoto, Ouchi, Kamigaito, and Watanabe]{tsujimoto_temporal_2024}
R.~Tsujimoto, H.~Ouchi, H.~Kamigaito, and T.~Watanabe.
\newblock Towards {{Temporal Change Explanations}} from {{Bi-Temporal Satellite Images}}, June 2024.

\bibitem[Vallet(2013)]{vallet_homological_2013}
B.~Vallet.
\newblock Homological persistence for shape based change detection between {{Digital Elevation Models}}.
\newblock \emph{ISPRS Ann. Photogramm. Remote Sens. Spatial Inf. Sci.}, II-3/W2:\penalty0 49--54, Oct. 2013.
\newblock ISSN 2194-9050.
\newblock \doi{10.5194/isprsannals-II-3-W2-49-2013}.

\bibitem[Van~Beek and Van~Asch(2004)]{vanbeek_regional_2004}
L.~Van~Beek and T.~Van~Asch.
\newblock Regional {{Assessment}} of the {{Effects}} of {{Land-Use Change}} on {{Landslide Hazard By Means}} of {{Physically Based Modelling}}.
\newblock \emph{Natural Hazards}, 31\penalty0 (1):\penalty0 289--304, Jan. 2004.
\newblock ISSN 1573-0840.
\newblock \doi{10.1023/B:NHAZ.0000020267.39691.39}.

\bibitem[{van der Velden} et~al.(2025){van der Velden}, Klerkx, Dessein, and Debruyne]{vandervelden_governance_2025}
D.~{van der Velden}, L.~Klerkx, J.~Dessein, and L.~Debruyne.
\newblock Governance by satellite: {{Remote}} sensing, bureaucrats and agency in the {{Common Agricultural Policy}} of the {{European Union}}.
\newblock \emph{Journal of Rural Studies}, 114:\penalty0 103558, Feb. 2025.
\newblock ISSN 0743-0167.
\newblock \doi{10.1016/j.jrurstud.2024.103558}.

\bibitem[Varghese et~al.(2025)Varghese, Gao, and Hoskere]{varghese_viewdelta_2025}
S.~Varghese, J.~Gao, and V.~Hoskere.
\newblock {{ViewDelta}}: {{Text-Prompted Change Detection}} in {{Unaligned Images}}, Mar. 2025.

\bibitem[Verbesselt et~al.(2012)Verbesselt, Zeileis, and Herold]{verbesselt_realtime_2012}
J.~Verbesselt, A.~Zeileis, and M.~Herold.
\newblock Near real-time disturbance detection using satellite image time series.
\newblock \emph{Remote Sensing of Environment}, 123:\penalty0 98--108, Aug. 2012.
\newblock ISSN 00344257.
\newblock \doi{10.1016/j.rse.2012.02.022}.

\bibitem[Vincent et~al.(2025)Vincent, Ponce, and Aubry]{vincent_satellite_2024}
E.~Vincent, J.~Ponce, and M.~Aubry.
\newblock Satellite {{Image Time Series Semantic Change Detection}}: {{Novel Architecture}} and {{Analysis}} of {{Domain Shift}}.
\newblock In \emph{Proceedings of the {{Computer Vision}} and {{Pattern Recognition Conference}} ({{CVPR}}) {{Workshops}}}, Nashville (Tennessee), United States, 2025.

\bibitem[{Voiron-Canicio}(2012)]{voiron-canicio_forecasting_2012}
C.~{Voiron-Canicio}.
\newblock {Forecasting Change in Prospective and Spatial Change in Geoprospective}.
\newblock \emph{L'Espace g{\'e}ographique}, 41\penalty0 (2):\penalty0 99--110, 2012.
\newblock ISSN 0046-2497.

\bibitem[Wang et~al.(2023)Wang, Zhang, and Shi]{wang_stcrnet_2023}
L.~Wang, M.~Zhang, and W.~Shi.
\newblock {{STCRNet}}: {{A Semi-Supervised Network Based}} on {{Self-Training}} and {{Consistency Regularization}} for {{Change Detection}} in {{VHR Remote Sensing Images}}.
\newblock \emph{IEEE J. Sel. Top. Appl. Earth Observations Remote Sensing}, pages 1--12, 2023.
\newblock ISSN 1939-1404, 2151-1535.
\newblock \doi{10.1109/JSTARS.2023.3345017}.

\bibitem[Wang et~al.(2024)Wang, Zhang, Bai, Chang, Chen, Zhang, and Tao]{wang_deep_2024}
Z.~Wang, J.~Zhang, L.~Bai, H.~Chang, Y.~Chen, Y.~Zhang, and J.~Tao.
\newblock A {{Deep Learning Based Platform}} for {{Remote Sensing Images Change Detection Integrating Crowdsourcing}} and {{Active Learning}}.
\newblock \emph{Sensors}, 24\penalty0 (5):\penalty0 1509, Jan. 2024.
\newblock ISSN 1424-8220.
\newblock \doi{10.3390/s24051509}.

\bibitem[Wu et~al.(2017)Wu, Du, Cui, and Zhang]{wu_postclassification_2017}
C.~Wu, B.~Du, X.~Cui, and L.~Zhang.
\newblock A post-classification change detection method based on iterative slow feature analysis and {{Bayesian}} soft fusion.
\newblock \emph{Remote Sensing of Environment}, 199:\penalty0 241--255, Sept. 2017.
\newblock ISSN 0034-4257.
\newblock \doi{10.1016/j.rse.2017.07.009}.

\bibitem[Wu et~al.(2022)Wu, Li, Yuan, Qin, Miao, and Gong]{wu_commonality_2022}
Y.~Wu, J.~Li, Y.~Yuan, A.~K. Qin, Q.-G. Miao, and M.-G. Gong.
\newblock Commonality {{Autoencoder}}: {{Learning Common Features}} for {{Change Detection From Heterogeneous Images}}.
\newblock \emph{IEEE Transactions on Neural Networks and Learning Systems}, 33\penalty0 (9):\penalty0 4257--4270, Sept. 2022.
\newblock ISSN 2162-2388.
\newblock \doi{10.1109/TNNLS.2021.3056238}.

\bibitem[Xavier et~al.(2016)Xavier, {Ariza-L{\'o}pez}, and {Ure{\~n}a-C{\'a}mara}]{xavier_survey_2016}
E.~M.~A. Xavier, F.~J. {Ariza-L{\'o}pez}, and M.~A. {Ure{\~n}a-C{\'a}mara}.
\newblock A {{Survey}} of {{Measures}} and {{Methods}} for {{Matching Geospatial Vector Datasets}}.
\newblock \emph{ACM Comput. Surv.}, 49\penalty0 (2):\penalty0 39:1--39:34, Aug. 2016.
\newblock ISSN 0360-0300.
\newblock \doi{10.1145/2963147}.

\bibitem[Xiao et~al.(2013)Xiao, Vallet, and Paparoditis]{xiao_change_2013}
W.~Xiao, B.~Vallet, and N.~Paparoditis.
\newblock Change {{Detection}} in {{3D Point Clouds Acquired}} by a {{Mobile Mapping System}}.
\newblock \emph{ISPRS Ann. Photogramm. Remote Sens. Spatial Inf. Sci.}, II-5/W2:\penalty0 331--336, Oct. 2013.
\newblock ISSN 2194-9050.
\newblock \doi{10.5194/isprsannals-II-5-W2-331-2013}.

\bibitem[Xiao et~al.(2023)Xiao, Cao, Tang, Zhang, and Chen]{xiao_3d_2023}
W.~Xiao, H.~Cao, M.~Tang, Z.~Zhang, and N.~Chen.
\newblock {{3D}} urban object change detection from aerial and terrestrial point clouds: {{A}} review.
\newblock \emph{International Journal of Applied Earth Observation and Geoinformation}, 118:\penalty0 103258, Apr. 2023.
\newblock ISSN 1569-8432.
\newblock \doi{10.1016/j.jag.2023.103258}.

\bibitem[Xuan et~al.(2025)Xuan, Wang, Qi, Chen, Zheng, Zhong, Xia, and Yokoya]{xuan_dynamicvl_2025}
W.~Xuan, J.~Wang, H.~Qi, Z.~Chen, Z.~Zheng, Y.~Zhong, J.~Xia, and N.~Yokoya.
\newblock {{DynamicVL}}: {{Benchmarking Multimodal Large Language Models}} for {{Dynamic City Understanding}}, May 2025.

\bibitem[Yang et~al.(2022)Yang, Xia, Liu, Du, Yang, Pelillo, and Zhang]{yang_semantic_2021}
K.~Yang, G.-S. Xia, Z.~Liu, B.~Du, W.~Yang, M.~Pelillo, and L.~Zhang.
\newblock Semantic {{Change Detection}} with {{Asymmetric Siamese Networks}}.
\newblock \emph{IEEE Transactions on Geoscience and Remote Sensing}, \penalty0 (60):\penalty0 1--18, 2022.
\newblock \doi{10.48550/arXiv.2010.05687}.

\bibitem[Yang et~al.(2018)Yang, Tang, Stewart, Dong, Zhang, and Li]{yang_automatic_2018}
X.~Yang, L.~Tang, K.~Stewart, Z.~Dong, X.~Zhang, and Q.~Li.
\newblock Automatic change detection in lane-level road networks using {{GPS}} trajectories.
\newblock \emph{International Journal of Geographical Information Science}, 32\penalty0 (3):\penalty0 601--621, Mar. 2018.
\newblock ISSN 1365-8816.
\newblock \doi{10.1080/13658816.2017.1402913}.

\bibitem[Young(2014)]{young_improving_2014}
S.~W.~H. Young.
\newblock Improving {{Library User Experience}} with {{A}}/{{B Testing}}: {{Principles}} and {{Process}}.
\newblock \emph{Weave: Journal of Library User Experience}, 1\penalty0 (1), 2014.
\newblock ISSN 2333-3316.
\newblock \doi{10.3998/weave.12535642.0001.101}.

\bibitem[Yu and Yang(2017)]{yu_3d_2017}
M.~Yu and C.~Yang.
\newblock A {{3D}} multi-threshold, region-growing algorithm for identifying dust storm features from model simulations.
\newblock \emph{International Journal of Geographical Information Science}, 31\penalty0 (5):\penalty0 939--961, May 2017.
\newblock ISSN 1365-8816.
\newblock \doi{10.1080/13658816.2016.1250900}.

\bibitem[Yu et~al.(2020)Yu, Bambacus, Cervone, Clarke, Duffy, Huang, Li, Li, Li, Liu, Resch, Yang, and Yang]{yu_spatiotemporal_2020}
M.~Yu, M.~Bambacus, G.~Cervone, K.~Clarke, D.~Duffy, Q.~Huang, J.~Li, W.~Li, Z.~Li, Q.~Liu, B.~Resch, J.~Yang, and C.~Yang.
\newblock Spatiotemporal event detection: A review.
\newblock \emph{International Journal of Digital Earth}, 13\penalty0 (12):\penalty0 1339--1365, Dec. 2020.
\newblock ISSN 1753-8947.
\newblock \doi{10.1080/17538947.2020.1738569}.

\bibitem[Yuan et~al.(2015)Yuan, Meng, Lin, Sahli, Yue, Chen, Zhao, Kong, and He]{yuan_continuous_2015}
Y.~Yuan, Y.~Meng, L.~Lin, H.~Sahli, A.~Yue, J.~Chen, Z.~Zhao, Y.~Kong, and D.~He.
\newblock Continuous {{Change Detection}} and {{Classification Using Hidden Markov Model}}: {{A Case Study}} for {{Monitoring Urban Encroachment}} onto {{Farmland}} in {{Beijing}}.
\newblock \emph{Remote Sensing}, 7\penalty0 (11):\penalty0 15318--15339, Nov. 2015.
\newblock ISSN 2072-4292.
\newblock \doi{10.3390/rs71115318}.

\bibitem[Zhan et~al.(2024{\natexlab{a}})Zhan, Zhu, Lan, and Dang]{zhan_crossdomain_2024}
T.~Zhan, Y.~Zhu, J.~Lan, and Q.~Dang.
\newblock Cross-{{Domain Separable Translation Network}} for {{Multimodal Image Change Detection}}, July 2024{\natexlab{a}}.

\bibitem[Zhan et~al.(2017)Zhan, Fu, Yan, Sun, Wang, and Qiu]{zhan_change_2017}
Y.~Zhan, K.~Fu, M.~Yan, X.~Sun, H.~Wang, and X.~Qiu.
\newblock Change {{Detection Based}} on {{Deep Siamese Convolutional Network}} for {{Optical Aerial Images}}.
\newblock \emph{IEEE Geoscience and Remote Sensing Letters}, 14\penalty0 (10):\penalty0 1845--1849, Oct. 2017.
\newblock ISSN 1558-0571.
\newblock \doi{10.1109/LGRS.2017.2738149}.

\bibitem[Zhan et~al.(2024{\natexlab{b}})Zhan, Xiong, and Yuan]{zhan_skyeyegpt_2024}
Y.~Zhan, Z.~Xiong, and Y.~Yuan.
\newblock {{SkyEyeGPT}}: {{Unifying Remote Sensing Vision-Language Tasks}} via {{Instruction Tuning}} with {{Large Language Model}}, Jan. 2024{\natexlab{b}}.

\bibitem[Zhang et~al.(2022)Zhang, Feng, Hu, Tapete, Pan, Liang, Cigna, and Yue]{zhang_domain_2022}
C.~Zhang, Y.~Feng, L.~Hu, D.~Tapete, L.~Pan, Z.~Liang, F.~Cigna, and P.~Yue.
\newblock A domain adaptation neural network for change detection with heterogeneous optical and {{SAR}} remote sensing images.
\newblock \emph{International Journal of Applied Earth Observation and Geoinformation}, 109:\penalty0 102769, May 2022.
\newblock ISSN 1569-8432.
\newblock \doi{10.1016/j.jag.2022.102769}.

\bibitem[Zhang et~al.(2023)Zhang, Sun, Herrera, and Zou]{zhang_datadriven_2023}
Q.~Zhang, Z.~Sun, L.~C. Herrera, and S.~Zou.
\newblock Data-{{Driven Quickest Change Detection}} in ({{Hidden}}) {{Markov Models}}.
\newblock In \emph{{{IEEE International Conference}} on {{Acoustics}}, {{Speech}}, and {{Signal Processing}}}. arXiv, Nov. 2023.

\bibitem[Zhang et~al.(2024)Zhang, Cai, Zhang, Zhuang, and Mao]{zhang_earthgpt_2024}
W.~Zhang, M.~Cai, T.~Zhang, Y.~Zhuang, and X.~Mao.
\newblock {{EarthGPT}}: {{A Universal Multi-modal Large Language Model}} for {{Multi-sensor Image Comprehension}} in {{Remote Sensing Domain}}, Jan. 2024.

\bibitem[Zhang et~al.(2017)Zhang, Wu, Wang, and Su]{zhang_geoeventbased_2017}
Y.~Zhang, W.~Wu, Q.~Wang, and F.~Su.
\newblock A {{Geo-Event-Based Geospatial Information Service}}: {{A Case Study}} of {{Typhoon Hazard}}.
\newblock \emph{Sustainability}, 9\penalty0 (4):\penalty0 534, Apr. 2017.
\newblock ISSN 2071-1050.
\newblock \doi{10.3390/su9040534}.

\bibitem[Zhao et~al.(2023)Zhao, Zhang, Dong, and Du]{zhao_adapting_2023}
Y.~Zhao, Y.~Zhang, Y.~Dong, and B.~Du.
\newblock Adapting {{Vision Transformer}} for {{Efficient Change Detection}}, Dec. 2023.

\bibitem[Zheng et~al.(2022)Zheng, Zhong, Tian, Ma, and Zhang]{zheng_changemask_2022}
Z.~Zheng, Y.~Zhong, S.~Tian, A.~Ma, and L.~Zhang.
\newblock {{ChangeMask}}: {{Deep}} multi-task encoder-transformer-decoder architecture for semantic change detection.
\newblock \emph{ISPRS Journal of Photogrammetry and Remote Sensing}, 183:\penalty0 228--239, Jan. 2022.
\newblock ISSN 0924-2716.
\newblock \doi{10.1016/j.isprsjprs.2021.10.015}.

\bibitem[Zhou et~al.(2014)Zhou, Yu, and Qin]{zhou_multilevel_2014}
J.~Zhou, B.~Yu, and J.~Qin.
\newblock Multi-{{Level Spatial Analysis}} for {{Change Detection}} of {{Urban Vegetation}} at {{Individual Tree Scale}}.
\newblock \emph{Remote Sensing}, 6\penalty0 (9):\penalty0 9086--9103, Sept. 2014.
\newblock ISSN 2072-4292.
\newblock \doi{10.3390/rs6099086}.

\bibitem[Zhu et~al.(2024)Zhu, Huang, Huang, Shao, and Cheng]{zhu_changevit_2024}
D.~Zhu, X.~Huang, H.~Huang, Z.~Shao, and Q.~Cheng.
\newblock {{ChangeViT}}: {{Unleashing Plain Vision Transformers}} for {{Change Detection}}, June 2024.

\bibitem[Zhu et~al.(2018)Zhu, Suomalainen, Liu, Hyypp{\"a}, Kaartinen, and Haggren]{zhu_review_2018}
L.~Zhu, J.~Suomalainen, J.~Liu, J.~Hyypp{\"a}, H.~Kaartinen, and H.~Haggren.
\newblock A {{Review}}: {{Remote Sensing Sensors}}.
\newblock In R.~B. Rustamov, S.~Hasanova, and M.~H. Zeynalova, editors, \emph{Multi-Purposeful {{Application}} of {{Geospatial Data}}}. InTech, May 2018.
\newblock \doi{10.5772/intechopen.71049}.

\bibitem[Zhu et~al.(2022)Zhu, Qiu, and Ye]{zhu_remote_2022}
Z.~Zhu, S.~Qiu, and S.~Ye.
\newblock Remote sensing of land change: {{A}} multifaceted perspective.
\newblock \emph{Remote Sensing of Environment}, 282:\penalty0 113266, Dec. 2022.
\newblock ISSN 0034-4257.
\newblock \doi{10.1016/j.rse.2022.113266}.

\end{thebibliography}
\endgroup
}

\newpage
\section*{Acknowledgements}

We would like to thank Anna Cristofol for her help writing an internal report, which provided the framework for writing this overview about change detection. We would like to thank Floryne Roche for the first version of the \ref{fig:graphical_abstract} and Clément Mallet for his review and valuable feedback. 
We thank our colleagues for IGN in the innovation, research, partenarial and production teams for inspiring discussions in the working group about change detection from March 2023 to November 2024.

\appendix

\section{Facets of change}
\label{annex:facets}

\begin{figure*}[ht!]
    \centering
        \includegraphics[width=1\linewidth]{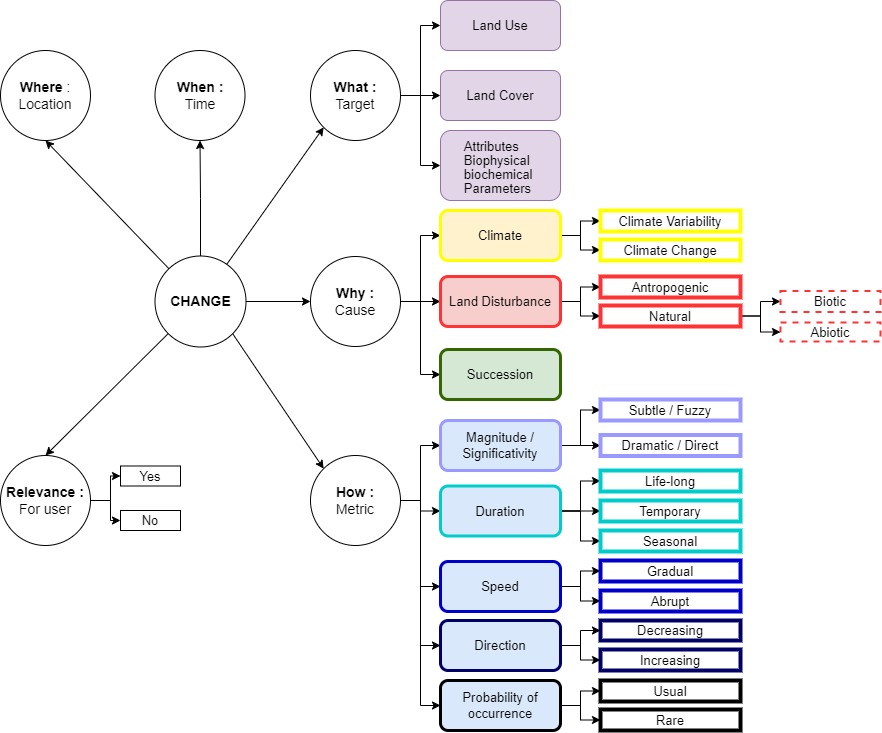}
    \caption{{\bf Taxonomy of Change.} Hierarchical classification system for the six facets of change, inspired by \cite{zhu_remote_2022}.}
    \label{fig:taxonomy}
\end{figure*}

Change definition varies based on the intended application. 
This concept of change is deeply intertwined with the adopted representation of the world, or \emph{ground truth}, as defined by database specifications in a geographic information system (GIS) \cite{pickles_ground_1995}. The interpretation of what constitutes a change depends on the objects of 
interest \cite{kadri-dahmani_updating_2001}. %
Thus, there is no such thing as a generic change and only spatiotemporal events \cite{yu_spatiotemporal_2020}. A change can be characterized by six aspects, summarized in \cref{fig:taxonomy}.

The first five facets of change are defined by \cite{zhu_remote_2022} in the case of remote sensing data but can be generalized to any kind of input data.
The first two aspects involve identifying when and where a significant difference occurs between two observations of the same element.
The third aspect of change is the answer to the "what" question, i.e. the semantization of this change, involving land cover or land use changes \cite{meyer_changes_1994}, or alterations in continuous biophysical parameters \cite{gomez-dans_location_2022,falconnier_modelling_2020,cui_spatiotemporal_2019}. 
The granularity of these changes can vary significantly, from hectare-scale objects \cite{knoefel_germanys_2021} to decimeter \cite{tian_hiucd_2020}.
The fourth aspect is answering the "how" question using various measures of change, such as its magnitude, duration or direction \cite{petit_quantifying_2001,kennedy_trajectorybased_2007}. This is the characterization of change. Changes can be subtle (few trees \cite{zhou_multilevel_2014}) or dramatic (large tree falls due to a storm \cite{chehata_objectbased_2014}). 
They may occur gradually (e.g. dieback) or abruptly (e.g. fire) and can be perennial or temporary. 
They may occur gradually (e.g. dieback \cite{price_shrub_1992}) or abruptly (e.g. fire \cite{balsak_evaluation_2023}) and can be perennial or temporary. 
Observing the speed of change depends on the frequency of data collection, and changes may be seasonal, reflecting natural or anthropogenic phenomena. 
The "why" aspect scopes the causes of change, which can be complex and multifaceted. Causes may include climatic variability or ongoing climate change but also natural, biotic, or anthropogenic disturbances \cite{zhu_remote_2022,mattson_role_1987}. 
A change can be linked to several causes. 
For example, drought can favor certain insect attacks \cite{mattson_role_1987}.
Over longer time scales, the concept of succession \cite{horn_forest_1975,decocq_dynamiques_2021} becomes relevant to define the natural evolution of an ecosystem from one state of equilibrium to another, with biotic replacement \cite{osinska-skotak_mapping_2019}. 
Finally, the relevance of observed changes for mapmakers or policymakers is crucial \cite{groot_challenges_1999,kadri-dahmani_updating_2001}. Not all detected changes necessitate updates to geodatabases, as their significance can vary depending on user perspectives \cite{pickles_ground_1995}. Filters need to be applied as appropriate. 
This complexity often hinders the industrialization of change detection methods, as updating a database does not always correlate with observed changes, reciprocally.

\section{Review of automated change detection techniques categorized by input data}
\label{annex:Review_methods}

In this section, we will discuss several methods categorized by the type of input data utilized. Each category encompasses techniques drawn from the principal families previously outlined in section \ref{sec:FourFamilies_Overview}. Generally, it is observed that as the spatial resolution of remote sensing data increases, the efficacy of change detection methods tends to decline. This is due to the amplified significance of non-change differences, such as variations in acquisition conditions (illumination, slope, etc.), and natural fluctuations in vegetation or soil texture. 

\subsection{2D Image}

2D imaging is the most traditional source for automatic change detection. It is studied for many decades since the beginning of the field of remote sensing \cite{singh_review_1989}. % newkirk_common_1990,mouat_remote_1993
 
Image change detection methods can be divided into 3 possible configurations
by type of comparison :
\begin{compactenum}
\item Comparing two photographs from the same sensor ;
\item Comparing two photographs from different sensors;
\item The comparison between a photograph and a geographic representation.
\end{compactenum}

\begin{figure*}
    \centering
    \includegraphics[width=1\linewidth]{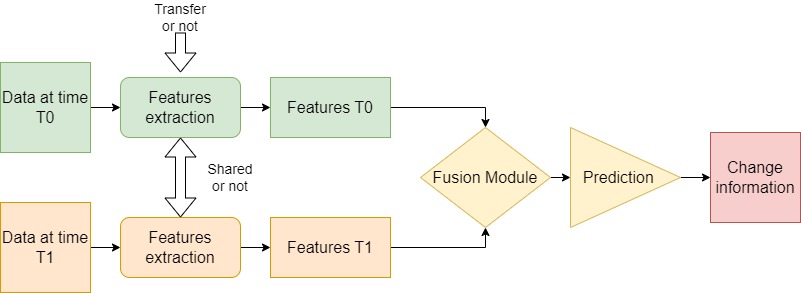}
    \caption{The generic bi-date change detection pipeline, based on machine learning models. Figure inspired by \cite{shi_change_2020}.}
    \label{fig:generic_pipeline_AI}
\end{figure*}

In the context of 2D image analysis, we find the four main families of methods mentioned above.
Rule-based methods include those based on algebraic operations on images, followed by thresholding. We can mention the difference between images, the ratio between them \cite{howarth_procedures_1981} or the change vector analysis (CVA) \cite{malila_change_1980}, which consists in considering the multispectral difference vector in polar or hyperspherical coordinates and attempts to characterize the changes on the basis of the vectors associated with each pixel. For this type of method, a great deal of attention is often paid to image pre-processing, with significant radiometric and geometric corrections, as well as homogenization between the two images under consideration \cite{du_radiometric_2002}. These methods are fast, but very sensitive to signal noise. As a result, they are more likely to be used for images in the ten-meter range than in the decimeter range.
The family of statistical methods includes data reduction methods such as principal component analysis \cite{mouat_remote_1993} (which searches for the main directions of signal variation), and multivariate alteration detection methods \cite{nielsen_multivariate_1998}, which seek to find a linear combination of input data according to various criteria to maximize the spatial coherence of correlations. These methods are more robust to changes in radiometry or atmospheric conditions. These methods can also be complex to calibrate and are also sensitive to changes in vegetation or object edges, leading to many false alarms.
In the case of machine learning methods, deep neural network-based methods provide the best results after the reemergence of these methods in the 2012s for natural image classification \cite{zhan_change_2017,daudt_fully_2018,khelifi_deep_2020,shafique_deep_2022} compared to traditional ML methods \cite{nemmour_multiple_2006}. The arrival of Transformer methods around 2020 for image recognition has not brought any methodological breakthrough to the change detection task \cite{bandara_transformerbased_2022,zhu_changevit_2024,chen_changemamba_2024}. The simplest approach is the "post-classification" approach \cite{wu_postclassification_2017}, which consists of performing a semantic segmentation of the input images individually and then differentiating between these predictions (according to the visual classes of interest) to obtain a change map but it can be bad because of the propagation of errors resulting from semantic segmentation. 
Changes are rare and can be drowned out by prediction noise. For this reason, many methods have addressed the issue of predicting the change map by leveraging information at both dates in a joint manner, as shown in the figure \ref{fig:generic_pipeline_AI}. 
Most of the works focus on transferring a new backbone from cutting edge computer vision \cite{zhu_changevit_2024}, proposing new fusion modules \cite{he_changeguided_2024,tian_temporalagnostic_2023,fang_changer_2023} or new training losses sometimes with negligible improvement \cite{corley_change_2024}. 
Finally, simulation methods involve simulating observed changes in a physically plausible way: changes in shading, illumination, sensor behavior, or landscape evolution. These are mainly used for model pretraining or evaluation of future sensor characteristics \cite{champion_detection_2011}.

\subsubsection{Identical Sensors configuration}

This configuration is the most widely studied due to its simplicity and maturity, particularly for human-induced changes. Deep learning currently outperforms classical machine learning methods \cite{nemmour_multiple_2006}, although overall performance remains lower than in generic computer vision tasks such as semantic segmentation or object detection, partly due to the strong class imbalance between changed and unchanged areas. Most approaches address binary change detection, e.g., building appearance or disappearance \cite{shen_s2looking_2021,ji_fully_2019,chen_spatialtemporal_2020}, while semantic change detection \cite{tian_hiucd_2020,toker_dynamicearthnet_2022,yang_semantic_2021}—tracking transitions between land-cover classes—is less explored due to its combinatorial complexity and is often tackled using semantic segmentation backbones \cite{daudt_weakly_2021}. Reported results are frequently limited to small geographic areas (e.g., urban regions \cite{tian_hiucd_2020}) and narrow class sets (e.g., buildings, roads \cite{ji_fully_2019,shen_s2looking_2021,chen_spatialtemporal_2020}), although mixture-of-experts strategies show promise for broader generalization \cite{kuriyal_codex_2025}. Extensive surveys are available for deep learning-based approaches \cite{shi_change_2020,shafique_deep_2022}, while Vision-Language Models currently underperform compared to supervised CNN/Transformer baselines \cite{dong_changeclip_2024,elgendy_geollava_2025,xuan_dynamicvl_2025}.

\subsubsection{Configuration with different sensors}
\label{sec:diff_sensors}

The second configuration compares two acquisitions from different sensors, also called heterogeneous change detection. It consists of comparing two different kinds of acquisitions \cite{barthelet_building_2011,figaritomenotti_heterogeneous_2021,wu_commonality_2022} and not fusing heterogeneous acquisitions of the same scene \cite{kawamura_automatic_1971,zhu_review_2018,zhan_crossdomain_2024}.
The boundary between the different configurations is not strict. However, as soon as there is a significant spectral or spatial difference between the two sensors, one might consider itself in this configuration. 
Due to its significant challenges and lack of labeled datasets, this subject is not often researched. For example, it would involve comparing SAR and optical images \cite{barthelet_building_2011}. As the spectral and spatial differences between these two modalities are very significant, the problem is to model the difference between these two sensors and their acquisition specificities before being able to estimate what the real change is \cite{sun_nonlocal_2021,ferraris_robust_2020}. The solutions provided can be based on domain adaptation \cite{zhang_domain_2022} or style transfer \cite{niu_conditional_2019}.
However, this method can exploit the complementary characteristics of one modality over another (such as repeatability or higher spatial resolution). 
For deep learning methods, it remains to be seen what the best solution would be between an agnostic model \cite{wu_commonality_2022} and a model that treats each of the modalities differently \cite{figaritomenotti_heterogeneous_2021}.
Two recent reviews on this topic for land cover change can be found in \cite{lv_land_2022,saidi_deeplearning_2024}.

\subsubsection{Configuration for raster and vector comparison}

Finally, the third configuration is "comparison between a shot and a geographic database". This involves comparing a raster source with the current state of the vector database on the given date. The simplest way to do this is to compare the prediction of an image recognition model applied to the raster source and to take the difference with a raster version of the database to get the differences.

This topic has been rather studied in the years 2010-2015 with classical machine learning methods \cite{gressin_updating_2014,gressin_updating_2013} before being put aside by academic research with the emergence of deep learning around 2012 before being studied with this tool \cite{bernhard_mapformer_2023,chen_objformer_2024,chen_change_2024}. This task is very difficult because the geographic database is often generalized, but very promising for operational cases \cite{niroshan_ml_2024}. There are therefore a number of differences between automatic predictions and database specifications. The results often have a very high uncertainty on the specification limits \cite{li_overcoming_2025}, which makes them difficult to use operationally and requires a great deal of post-processing. However, these methods are robust to sensor changes, they can learn from the state of the base at state N-1 \cite{gressin_semantic_2013} and therefore require fewer annotations, and they can automatically adapt to any new base or zone.
These families are the most directly useful for updating geographic databases, as they enable both updating and upgrading. They can also be used to complement updates from other information sources.

\subsubsection{Optical Images from Remote Sensing}

Airborne, spaceborne or UHV optical images are the main source used in automatic change detection methods, as they offer significant spectral and spatial information. In addition, the fact that they have historically been the main source for human photo-interpretation for the creation of modern maps and geographic databases \cite{bagley_concerning_1922,baldwin_use_1947,dyce_canada_2013} has strongly influenced past technical and technological choices, as well as the lines of research studied in the past. Optical images can be divided into hyperspectral, multispectral and panchromatic images, depending on the number of bands. 
Hyperspectral images are composed of hundreds of spectral bands, narrow over a large portion of the electromagnetic spectrum; the band range is generally less than 10 nm. Multispectral images generally contain several bands, but fewer than 15. The spectral resolution of multispectral images is in the range of 0.1 times the wavelength. Panchromatic images have a single band formed by using the total light energy available in the visible spectrum (rather than partitioning it into different bands).
Multi-temporal hyperspectral images offer promising performance in detecting low-resolution changes in vegetation or in the geological composition of soils. However, due to the generally low spatial resolution of hyperspectral images, the pixels mix different objects and image interpretation is rather difficult. 
Change detection methods for hyperspectral images have to solve the problems of high dimensionality, mixed pixels, and a limited training dataset \cite{amieva_deeplearningbased_2023}. In addition, these methods are costly and require precise calibration.
The use of multi-spectral imagery has grown considerably with the arrival of satellites such as LandSat \cite{gordon_utilizing_1980} and Sentinel-2 \cite{daudt_urban_2018}. 
These images can offer a wide variety of colors, textures and other spectral properties.
As mentioned above, most change detection methods have been studied and developed for this type of input data \cite{shen_s2looking_2021,zhan_change_2017,ji_fully_2019,chen_spatialtemporal_2020,daudt_fully_2018,daudt_multitask_2019,singh_review_1989,mouat_remote_1993}.

\subsubsection{SAR images}

SAR (Synthetic Aperture Radar) images are the result of a technique that uses signal processing to improve resolution beyond the limits of the antenna's physical aperture. SAR imaging has the advantage of being independent of atmospheric conditions and solar illumination and of providing interferometric information. 
It is a type of imaging that is highly sensitive to the dielectric properties of the observed surface. This technology can be useful for detecting changes in certain types of object, such as an antenna or a building, as some of these will have a strong RADAR signature despite their small size. 
Depending on the type of band used for acquisition, one will not obtain an image with the same resolution. C-band sensors like Sentinel-1 have a good ratio between spatial resolution (10 m) and returned signal. L-band sensors can penetrate some vegetation, but with less spatial resolution, while X-band sensors have metric resolution but will not penetrate vegetation.
This type of instrument can also be used to detect very subtle changes, such as soil deformation \cite{raspini_continuous_2018}, soil moisture \cite{hornacek_potential_2012} or glacier movements \cite{foresta_heterogeneous_2018}. It may be required to instrument the site. 
It is therefore a valuable source of information for change detection. SAR images can be multi-band and multi-polarization. 
Polarization information can be used to better characterize what is on the ground (e.g. vegetation). In addition, interferometric measurements of the SAR signal can be a proxy for detecting change when a loss of coherence is observed \cite{preiss_coherent_2006}, especially when a time series is available; see \ref{sec:time_series}.
However, it is a technology that is less widely available than optical imaging, with a higher cost of entry and a greater need for expertise. This makes it a less studied and less available technology. In addition, SAR images still suffer from the effects of speckle noise, which can make the process of detecting changes more difficult than with optical images. This sometimes necessitates averaging the signal from a temporal stack of SAR images to obtain a less noisy acquisition. Another limitation of RADAR data is that the signal can potentially penetrate the ground over an unknown distance and, therefore, measure both the surface and the subsurface.
Statistical methods are still very efficient for SAR input data \cite{colinkoeniguer_change_2020}.
Change detection methods using deep learning techniques are also state-of-the-art methods for bi-date change detection \cite{li_deep_2019}. Note that currently change detection between SAR images from two sensors on ascendant and descendant trajectories leads to very poor performances, unlike in the optical case.

\subsubsection{Street View Image}

Unlike optical and SAR images, Street View images are captured at ground level rather than from the air and most of the time there are panoramic views (180$^{\circ}$ or 360$^{\circ}$). They provide more detailed information in relatively small areas and from different viewing angles, which can be useful for detecting dynamic changes in almost real time. Change detection methods based on Street View images focus on modifications to the urban visual landscape \cite{varghese_viewdelta_2025,kim_generalizable_2025}, such as the addition or removal of specific landmarks, crosswalks, vehicles, and other roadside buildings \cite{lei_hierarchical_2021,chen_drtanet_2021}. A critical challenge lies in identifying noisy changes caused by various lighting conditions, camera viewpoints, occlusions, and shadows \cite{sachdeva_change_2023}.
The variability in shooting conditions complicates the identification and quantification of intended semantic changes in Street View images. Therefore, deep learning algorithms are the most efficient techniques for detecting changes in Street View images, which most closely resemble the natural images used in current computer vision methodologies, heavily influenced by autonomous vehicle research. Smartphone user images are also pertinent. Data fusion with 3D data acquired simultaneously may improve performance as seen in \cref{sec:fusion}.

\subsection{3D Data}
\label{sec:3Ddata}

3D data in all its forms (point clouds, meshes, voxels) is also an important source of information to detect changes in the landscape \cite{qin_3d_2016,stilla_change_2023,kharroubi_three_2022}. The advantages of 3D data include the ease with which data from different resolutions can be compared, the fact that height is a component that is robust to variations in illumination, that it is free from slope effects even at high resolution, and that the data can be acquired from a wide variety of angles. However, these data are more expensive to acquire than 2D data, and may contain significant artifacts: problems with image correlation in photogrammetry \cite{rupnik_micmac_2017}, for example.
It should be noted that in 3D, there are not just two possible states for an object, but three: change, no change, or unknown because it is visible on one acquisition but not on the other, due to occlusions \cite{xiao_change_2013}. The main limitation of rule-based methods in the 3D case is related to data quality \cite{vallet_homological_2013}. For most of these methods, the change alerts arise from sensor-related variations in acquisitions or from elements within the specification limit, such as the limits of the building roof \cite{champion_2d_2010}.
The first major family of methods (which can be included in rule-based methods) consists in comparing geometries with each other. Depending on the shooting scenario (oblique, from above) and the data format (DSM, point clouds, stereo images, etc.), the geometric comparison can be quite different as mentioned in \cite{qin_change_2021}. It can be a 2.5D comparison, such as height or depth difference \cite{sasagawa_investigation_2013,guerin_automatic_2014,champion_2d_2010}, which is very simple to implement and scalable on a large scale, but very sensitive to registration errors. 
Alternatively, it can be a fully 3D comparison using distance measurement \cite{kang_change_2011,stilla_change_2023}. This method is less sensitive to small registration errors, but can be time-consuming when searching for correspondences between the two surfaces. 
Finally, a set of methods is based on the projection of 2D image pairs onto a DSM or reference point cloud to compare reprojected data \cite{taneja_image_2011,kang_change_2011}. This reduces stereo image matching errors and can be applied to stereo images with large intersection angles. This method is particularly effective when high-quality DSMs or dot clouds are available. Adding spectral information to the 3D data can remove ambiguities and improve semantic segmentation performance \cite{xiao_3d_2023}.
Similarly, in the context of 3D data, recent years have seen an evolution in the methods developed, with a shift from rule-based methods (such as MNS difference thresholding) to machine learning methods \cite{degelis_change_2024,kharroubi_three_2022}.
 
Deep learning methods for detecting changes between 3D representations is an active and promising area of research, even though these data are much larger than 2D data. We can thus distinguish methods based on the idea of comparison after classification of the point cloud or on the comparison of the two 3D models within the same model. 
As with 2D data, the first limitation is the absence of massive and diverse training sets, but the synthetic data set seems to transfer efficiently \cite{degelis_benchmarking_2021}.
In addition, methods for merging 3D meshes from different acquisitions are mature but not yet industrialized. This involves merging several meshes of different spatial resolutions, in order to remove non-perennial objects from the mesh \cite{xiao_change_2013}. It can also be used to update a high-resolution mesh with a new acquisition of lower spatial resolution, while retaining as much information as possible from the initial mesh.
It may be noted that the current difficulties of change detection on 3D data are uncertainty management (to compensate for the absence of a quality index), the management of heterogeneous sensors and the semantic classification of observed changes as mentioned in \cite{qin_3d_2016}.

\subsection{Time Series}
\label{sec:time_series}

\begin{figure}
    \centering
    \includegraphics[width=1\linewidth]{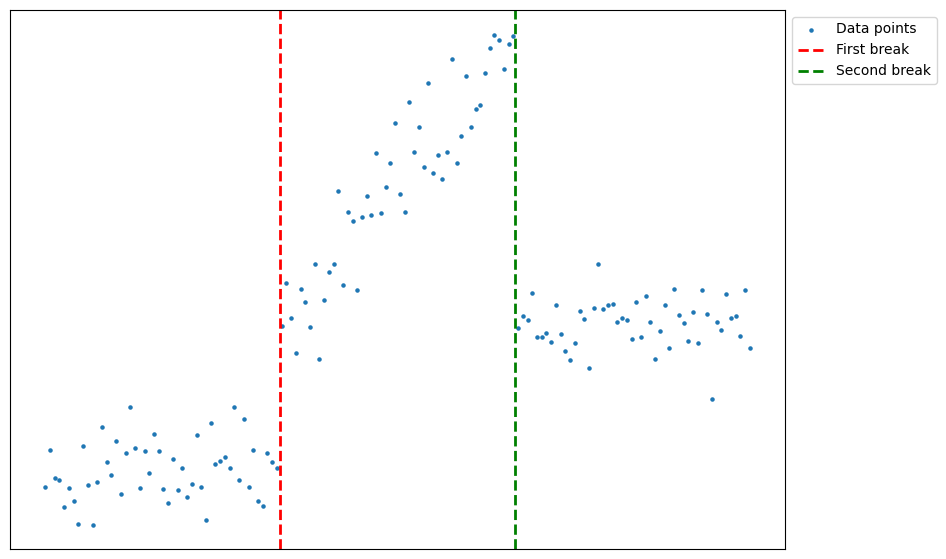}
    \caption{Illustration of two breaks in a time series.}
    \label{fig:time_series}
\end{figure}

Time series are crucial for change detection, as they inherently allow tracking phenomena over both short and extended intervals. Applications include time series from image sensors, position sensors, or statistical surveys.
Note that time series should not be treated in the same way as previous data, as more than two observations of the same point in the territory are available (usually many more). This large number of observations makes it possible to determine an average trajectory and deviations from it \cite{pasquarella_demystifying_2022}. This is known as break detection. 
Breakpoint detection \cite{truong_selective_2020,amiri-simkooei_offset_2018,gazeaux_detecting_2013,quarello_gnssseg_2022} is used to identify time intervals when a model transforms into a new model (such as a change in mean value or variance), as shown in Figure \ref{fig:time_series}. This differs from anomaly or outlier detection, which identifies time intervals that significantly deviate from a specified model. The former presumes a persistent change, whereas the latter considers a brief anomaly.
The limitations of these methods are that they require considerable expertise and adaptation for each new application.
High repeatability can enable to distinguish changes earlier and characterize changes that would not be seen with a single observation. For example, when one can determine the phenology or health of a plant with a series of images.

In the context of satellite imagery, several robust indicators have been developed to track specific elements of the Earth's surface based on their spectral signature. These include NDVI for vegetation monitoring \cite{gandhi_ndvi_2015}, and DDWI for coastline monitoring \cite{abdelhady_simple_2022} in optical images (Sentinel-2 type). There are also methods to detect changes in the time series of SAR images \cite{colinkoeniguer_change_2020}. However, it requires a certain number of acquisitions to parameterize the model or correct for noise before a suitable result can be obtained, at least five images and ideally around twenty. These methods can be used to monitor natural phenomena such as deforestation or forest health \cite{kennedy_detecting_2010}. Those methods are mature and can provide operational tools \cite{esa_using_2023}.
 It should be noted that there are sensors with good repeatability and better resolution such as Planet satellites \cite{toker_dynamicearthnet_2022}, but this is often to the detriment of geometric quality and image localization, which can hinder their use for change detection. The supervised machine learning method tends to be competitive in the detection of time-series changes without modeling the behavior of the signal \cite{toker_dynamicearthnet_2022,vincent_satellite_2024} but lack of spatial generalization capacity.

Statistical methods are the most commonly used methods for detecting changes in time series.  
Simple examples of statistical methods include estimating the mean and standard deviation of a signal \cite{knoefel_germanys_2021}, or more complex statistics \cite{derksen_geometry_2020} (such as median, mean shift, moments of order N). Most of these methods involve defining a good statistical estimator, and then modeling the probability distribution of the signal under study as closely as possible.
Sliding window methods \cite{truong_selective_2020} involve estimating an average behavior over a window (e.g. a linear trend using least squares) and comparing (using a T-student statistical test) that on either side of the same point to estimate a significant difference. However, these methods are highly sensitive to noise in the data.
Parametric methods (such as BFAST \cite{verbesselt_realtime_2012}) can be broken down into two phases. The first phase involves the calibration of a parametric function on the available splines or piecewise linear samples. This enables the normal trajectory of the observed phenomenon to be represented, and seasonal variations, background drifts or trends and residual noise to be estimated. The second phase, known as "tracking", consists of estimating a deviation from this estimated trajectory to detect a change.
The maximum likelihood estimation method \cite{bhattacharya_maximum_1987} consists in inferring the parameters of the probability distribution of a given sample by searching for the parameter values that maximize the likelihood function. Various studies have sought to add various penalties to limit the number of breaks detected, as well as the type of behavior models used (linear with various period terms, piecewise linear, stochastic).
Hidden Markov models \cite{zhang_datadriven_2023,yuan_continuous_2015} are also a solution to represent a temporal signal. Here, the underlying system is considered a Markov process of unknown parameters, of which only a few are known. Estimation of this type of model can be done efficiently using expectation-maximization algorithms (similar to maximum likelihood estimation), but this is sometimes computationally time-consuming.
The "a contrario" method \cite{robin_acontrario_2010,tailanian_contrario_2023,dagobert_detection_2022}, slightly different from other statistical methods, consists in estimating whether the observed phenomenon is due to chance or not, i.e. their probability. Significant changes are defined as events with a low probability of occurrence based on the number of false alarms. It enables interaction with the false alarm rate that one want to obtain by adjusting the hypothesis test threshold. The difficulty lies in defining the right statistics for each use case.

Methods for detecting breaks in GNSS signal time series have not yet reached operational status \cite{gazeaux_detecting_2013}. These methods are based on maximum likelihood estimation (with different penalization \cite{bertin_semiparametric_2017,gazeaux_joint_2015}), hypothesis testing \cite{amiri-simkooei_offset_2018} or anomaly detection by calculating distances over sliding windows \cite{truong_selective_2020}. In this particular case, manual methods tend to be better at detecting breaks and estimating velocities on simulated series, and to meet the requirement for millimetric accuracy.
In addition, data on statistical developments in a given area can indicate current trends \cite{canada_depth_2024}, but statistical studies often take a long time to consolidate and publish.
Finally, in the case of geolocated trajectories, the fact of having a large number of observations is a way of obtaining a less noisy signal and then comparing this signal with a reference using matching methods. For example, by comparing the paths of a large group with their expected path \cite{ivanovic_filteringbased_2019}. For this type of data, the primary challenge is in efficiently identifying anomalies in the geolocated signal (outliers) and displacements connected to ground conditions unrelated to the path (e.g., detours to reach a picnic area).

\subsection{Text Data}

In the field of natural language processing, it is the notion of event rather than change that is relevant \cite{yu_spatiotemporal_2020,zhang_geoeventbased_2017}. The recent advancements in automatic language processing techniques has made it possible to mine data and analyze messages posted online by citizens.
Academic and private research has mainly focused on detecting events on social networks, and Twitter in particular, by opportunism: messages on Twitter being until recently easily recoverable. Events that are intended to identify usually involve singular disasters or crises \cite{zhang_geoeventbased_2017,hu_using_2014,florath_rapid_2024} sometimes coupled with remote sensing data \cite{huang_sar_2025}, but can be used to identify stakeholders in land use decisions at the local level \cite{hauck_using_2016}.
In the last decade, works on the exploitation of online traces have concluded that geolocated digital traces, such as tweets, initially have an overestimated potential for the precise study of socio-spatial issues \cite{saldana-perez_when_2019}.
%There is no evidence of universal virtual behavioral patterns. 
One limitation is that the structure of virtual activity is inconsistent and unreliable as a spatial sample, both in normal circumstances and during crises. As a result, some locations are over-represented, while others are invisible. The proposed algorithms are unable to model explanatory factors for virtual activity locations that can be generalized from one territory to another, or from one context to another.
Nevertheless, recent works tried to tackle those problems for operational use \cite{adrot_using_2022,farah_mcatnat_2024}.

Exploiting identified sources, such as the local press, seems to have a greater
potential to update the geodatabase, but this has not been studied.
It would therefore be advisable to rely on fixed, or even institutional data streams, rather than free web crawling, to perform change detection for NMA.
For textual data, information on territorial changes is not necessarily linked to an exact geographic location, but rather to a textual description, when available. These descriptions, varying in precision, can help localize the change on a certain scale. The question of privacy arises when online personal messages are used \cite{majeed_comprehensive_2022}.

\subsection{Historical Data}

Several research projects have focused on detecting changes in historical data using old photographs or maps. 
They aim to monitor the evolution of the territory over a long period of time \cite{lebris_archival_2019} sometimes several hundred years, for example, with the evolution of the landscape around a remarkable building \cite{gominski_challenging_2019}, or the evolution of certain bio-physical characteristics: forests \cite{bowman_human_2011}, hydrology \cite{james_geomorphic_2012}, roads \cite{rath_settlement_2023}, etc. The difficulty of this task is further increased by the increasing differences between the various "acquisitions" \cite{gominski_challenging_2019}: photography, maps, drawings, paintings, etc.
In the case of aerial photography alone, differences in radiometry and normalization are a real lock in the detection of changes between 1950 and today \cite{lelegard_correction_2020}. In addition, there is a real lack of annotation of historical data. 
It should be noted that techniques for coregistration of historical images are fairly mature \cite{giordano_automatic_2018} and are the first necessary step before change detection. Then automatic exploitation of historical maps is needed to scale the analysis \cite{patru-stupariu_using_2013,freudiger_historical_2018}.
A long-term analysis of the territory can distinguish between significant changes and normal or natural evolutions. However, this requires the ability to model the fact that what is considered a change by users may evolve over time. This has an impact on the cartographic representation of the territory.

\subsection{Vector Data}

Matching vector data is quite common for updating the geodatabase, for instance, for road networks \cite{fan_polygonbased_2016}, relief or building footprint matching \cite{naumann_manymany_2024}.
It is the process of gathering different data sources with pronounced similarity, in a geospatial way, it can also be named linking, alignment or reconciliation as mentioned in \cite{xavier_survey_2016}. 
Matching methods are based on similarity measures that can be geometric, topological, attribute, context, or semantic. The methods can be schema matching (finding the semantic correspondence), feature matching (finding the correspondence between objets) or internal matching (matching parts of a geometry). 
The methods can be one-to-one (1:1), one-to-many (1:n), and many-to-many (m:n) matching. 

The main challenges for this task are adding geographical knowledge, using multiscale information \cite{abbaspour_multiscale_2021}, computational efficiency and scalability \cite{naumann_manymany_2024}, complex shapes, differences in data acquisition \cite{fan_polygonbased_2016}, handling uncertainty, and dealing with more than two datasets at the same time. 
The matching research community should provide benchmark test data that covers various real-world scenarios \cite{xavier_survey_2016}. Schema matching is an active research topic in computer science, but we believe that it does not receive enough attention for updating geospatial data.

\section{Output level}
\label{annex:output_level}

\begin{figure*}[h]
    \centering
    \includegraphics[width=1\linewidth]{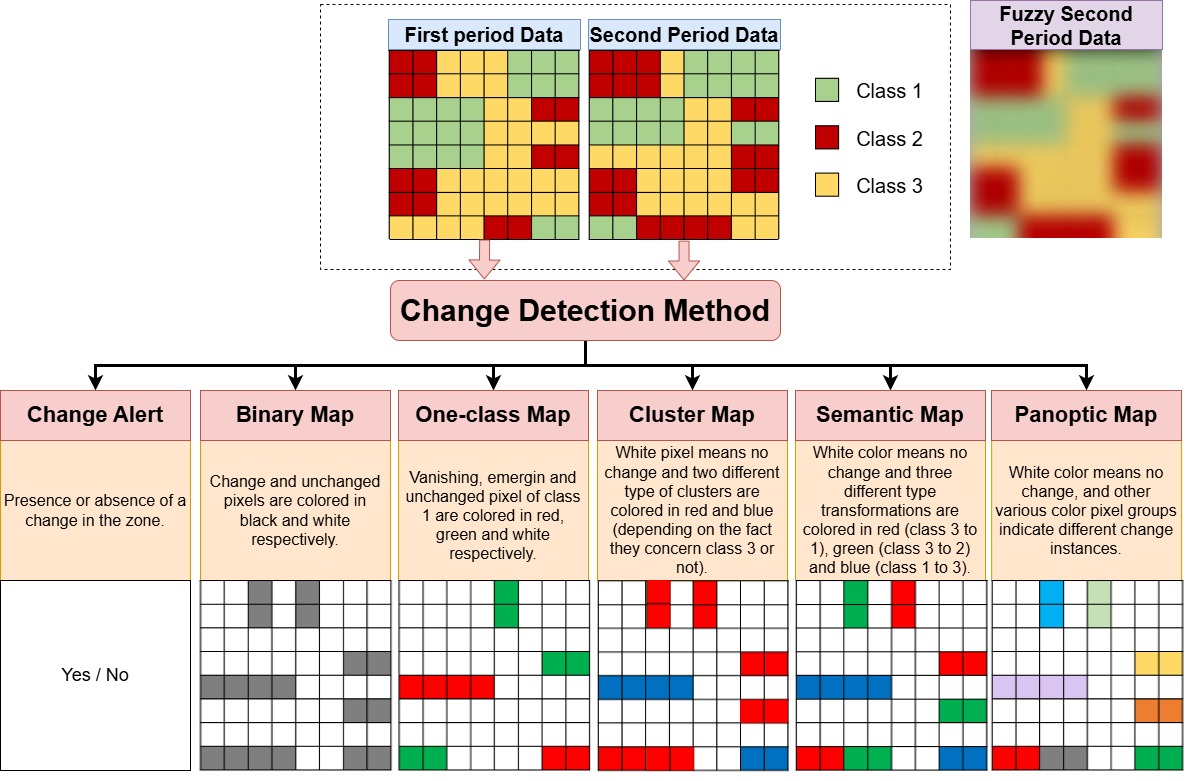}
    \caption{Different types of change output can be obtained. Figure inspired by \cite{shi_change_2020}.}
    \label{fig:output}
\end{figure*}

\Cref{fig:output} presents a depiction of the various output tiers possible with raster-based methods  between two input data at T1 and T2, but it is straightforward to extend it to time series or other kinds of data. The different output levels are :
\begin{compactenum}
    \item A binary alert: informing about a change in the zone or not.
    \item A binary change map: a map indicating each element of the input data that has changed between the two dates.
    \item A mono-class (mono-theme) change map: the focus is on a single category of change.
    \item A clustered change map: similar changes are grouped together (the type of similarity considered depends on the methods).
    \item A semantic change map: each segment is accompanied by information indicating its starting class (at T1) and its ending class (at T2).
    \item A panoptic change map: each object that has changed is segmented individually (for example, each building in a new district).
\end{compactenum}

It should be noted that there are two ways of approaching the question of change. It can be \textit{object-specific}, by developing methods that are specific to a given object (for instance, building detection \cite{ji_fully_2019}), or \textit{holistic}, by developing methods that detect all changes \cite{dagobert_detection_2022,bruzzone_automatic_2000}, with often posterior qualification.

\end{document}